\documentclass[twoside,11pt]{article}

\usepackage{jmlr2e} 
\usepackage{amsfonts,amsmath,stmaryrd}
\usepackage{booktabs}
\usepackage{dsfont}
\usepackage{mathrsfs} 
\usepackage{multirow}
\usepackage{tikz}
\usepackage{upgreek}
\usepackage{enumitem}
\usepackage{graphicx}
\usepackage{multicol}
\usepackage{ntheorem} 
\usepackage[nameinlink,capitalize]{cleveref}
\usepackage{lscape} 
\usepackage{xspace}
\usepackage{etoc} 

\makeatletter
\def\cleartheorem#1{\expandafter\let\csname#1\endcsname\relax
    \expandafter\let\csname c@#1\endcsname\relax
}
\makeatother
\cleartheorem{example}
\cleartheorem{theorem}
\cleartheorem{lemma}
\cleartheorem{proposition}
\cleartheorem{definition}
\cleartheorem{corollary}

\newtheorem{theorem}{Theorem}
\newtheorem{lemma}[theorem]{Lemma} 
\newtheorem{proposition}[theorem]{Proposition} 
\newtheorem{corollary}[theorem]{Corollary}
\newtheorem{definition}[theorem]{Definition}

\usepackage{subfig}
\DeclareGraphicsExtensions{.pdf,.png,.eps,.jpg,.jpeg,.tif,.tiff,.ps}
\graphicspath{{figs//}}

\allowdisplaybreaks

\definecolor{navy}{rgb}{0.1, 0.1, 0.8}
\definecolor{gray}{rgb}{0.4, 0.4, 0.4}
\definecolor{olive}{rgb}{0.1, 0.5, 0.1}
\definecolor{ruby}{rgb}{0.8, 0.1, 0.3}
\definecolor{ao}{rgb}{0.0, 0.5, 0.0}

\newcommand{\paperTitle}{Interval-censored Hawkes processes}

\usepackage{lastpage}
\jmlrheading{23}{2022}{1-\pageref{LastPage}}{8/21; Revised
6/22}{11/22}{21-0917}{Marian-Andrei Rizoiu, Alexander Soen, Shidi Li, Pio Calderon, Leanne Dong, Aditya Krishna Menon and Lexing Xie}

\ShortHeadings{\paperTitle}{Rizoiu, Soen, Li, Calderon, Dong, Menon, and Xie}
\firstpageno{1}

\newcommand{\ie}{i.e.\xspace}

\newcommand{\bheader}[1]{\vspace{1.5mm}\noindent\textbf{#1}}

\newcommand{\defEq}{\stackrel{.}{=}}

\newcommand{\tick}{$\checkmark$}
\newcommand{\cross}{$\times$}

\newcommand\independent{\protect\mathpalette{\protect\independenT}{\perp}}
\def\independenT#1#2{\mathrel{\rlap{$#1#2$}\mkern2mu{#1#2}}}

\newcommand{\argmax}[2]{\underset{#1}{\operatorname{argmax }}\, #2}

\renewcommand{\Pr}{\mathbb{P}}
\newcommand{\E}[2]{{\mathbb{E}_{#1}}\left[ #2 \right]}

\newcommand{\rectangle}{\tikz \fill [black] (0.1, 0.1) rectangle (0.2,0.2);}

\newcommand{\HSf}{\mathcal{H}}

\newcommand{\N}{\mathsf{N}}

\newcommand{\LCal}{\mathscr{L}}

\newcommand{\lambdaSf}{\uplambda}

\renewcommand{\d}{\mathrm{d}}
\newcommand{\dx}{\d x}
\newcommand{\du}{\d u}

\newcommand{\HPintens}{\lambda}
\newcommand{\MBPintens}{\xi}
\newcommand{\immiIntens}{s}
\newcommand{\HPcomp}{\Lambda}
\newcommand{\MBPcomp}{\Xi}
\newcommand{\immiComp}{S}

\newcommand{\arbintens}{\lambda}
\newcommand{\exointens}{\lambda^{\text{exo}}}
\newcommand{\edointens}{\lambda^{\text{endo}}}

\newcommand{\immiCounts}{S}
\newcommand{\HPCounts}{N}
\newcommand{\MBPCounts}{M}

\newcommand{\immiObserVolume}{\mathsf{S}}
\newcommand{\HPObserVolume}{\mathsf{C}}

\newcommand{\OffspringVolume}{\mathsf{F}}

\newcommand{\immiMDelta}{s_{m\delta}}
\newcommand{\immiIntensStep}{s_{\rectangle}}

\newcommand{\exoIntensity}{\xi^{\text{exo}}}
\newcommand{\edoIntensity}{\xi^{\text{endo}}}
\newcommand{\edoMBPcomp}{\Xi^{\text{endo}}}

\newcommand{\eventOccurTime}{t}
\newcommand{\immiOccurTime}{\mathfrak{s}}
\newcommand{\obsTime}{o}
\newcommand{\immiTime}{q}
\newcommand{\eventObservTime}{\obsTime}

\newcommand{\eventOccurTimeSet}{\Omega}
\newcommand{\immiOccurTimeSet}{\Psi}

\newcommand{\obsTimeSet}{O}
\newcommand{\immiObsTimeSet}{Q}

\newcommand{\offspringOccurTime}{f}
\newcommand{\offspringOccurTimeSet}{\Upsilon}

\newcommand{\kernel}{\phi} \newcommand{\bregdiv}{B}

\newcommand{\edoLL}{\mathscr{L}^{\text{endo}}}
\newcommand{\param}{\theta}
\newcommand{\edoparam}{\param^{\text{endo}}}
\newcommand{\exoparam}{\param^{\text{exo}}}
\newcommand{\npll}{\LCal_{\rm{IC-LL}}}

\newcommand{\approxpoint}{d}
\newcommand{\approxupper}{D}
\newcommand{\approxset}{\mathcal{D}}

\newcommand{\ignore}[1]{}

\newcommand{\actived}{\texttt{ACTIVE}\xspace}
\newcommand{\hprect}{\texttt{HP-pc}\xspace}
\newcommand{\hpsin}{\texttt{HP-sin}\xspace} 
\begin{document}
\etocdepthtag.toc{mtchapter} 

\title{\paperTitle}

\author{\name Marian-Andrei Rizoiu \email Marian-Andrei.Rizoiu@uts.edu.au \\
       \addr University of Technology Sydney \\
       Ultimo NSW 2007, Australia \\
       \name Alexander Soen \email alexander.soen@anu.edu.au \\
       \addr The Australian National University \\
       Canberra ACT 2601, Australia \\
       \name Shidi Li \email shidi.li@anu.edu.au \\
       \addr The Australian National University \\
       Canberra ACT 2601, Australia \\
       \name Pio Calderon \email piogabrielle.b.calderon@student.uts.edu.au \\
       \addr University of Technology Sydney \\
       Ultimo NSW 2007, Australia \\
       \name Leanne J. Dong \email leanne.dong@concordia.ca \\
       \addr Concordia University \\
       Montr\'eal, Canada \\       
       \name Aditya Krishna Menon \email adityakmenon@google.com \\
       \addr Google Research \\
       \name Lexing Xie \email lexing.xie@anu.edu.au \\
       \addr The Australian National University \\
       Canberra ACT 2601, Australia
       }

\editor{Mohammad Emtiyaz Khan}

\maketitle

\begin{abstract}

Interval-censored data solely records the aggregated counts of events during specific time intervals -- such as the number of patients admitted to the hospital or the volume of vehicles passing traffic loop detectors -- and not the exact occurrence time of the events.
It is currently not understood how to fit the Hawkes point processes to this kind of data.
Its typical loss function (the point process log-likelihood) cannot be computed without exact event times.
Furthermore, it does not have the independent increments property to use the Poisson likelihood.
This work builds a novel point process, a set of tools, and approximations for fitting Hawkes
processes within interval-censored data scenarios.
First, we define the Mean Behavior Poisson process (MBPP), a novel Poisson process with a direct parameter correspondence to the popular self-exciting Hawkes process. 
We fit MBPP in the interval-censored setting using an interval-censored Poisson log-likelihood (IC-LL). 
We use the parameter equivalence to uncover the parameters of the associated Hawkes process.
Second, we introduce two novel exogenous functions to distinguish the exogenous from the endogenous events.
We propose the multi-impulse exogenous function -- for when the exogenous events are observed as event time -- and the latent homogeneous Poisson process exogenous function -- for when the exogenous events are presented as interval-censored volumes.
Third, we provide several approximation methods to estimate the intensity and compensator function of MBPP when no analytical solution exists. 
Fourth and finally, we connect the interval-censored loss of MBPP to a broader class of Bregman divergence-based functions.
Using the connection, we show that the popularity estimation algorithm Hawkes Intensity Process (HIP)~\citep{Rizoiu:2017} is a particular case of the MBPP. 
We verify our models through empirical testing on synthetic data and real-world data.
We find that our MBPP outperforms HIP on real-world datasets for the task of popularity prediction.
This work makes it possible to efficiently fit the Hawkes process to interval-censored data. \end{abstract}


\section{Introduction}

Point processes are a class of well-understood mathematical instruments to model the occurrence of events in time (and optionally space).
The \emph{Hawkes process}~\citep{Hawkes:1971} is a particular type of point process that can model \emph{self-excitation} -- i.e., the occurrence of one event increases the likelihood of future events.
The Hawkes process was successfully deployed for applications where the events are individually observable -- such as earthquakes~\citep{Ogata1988,Cox:1980,Daley:2003},
financial transactions~\citep{Bowsher:2007,Hardiman:2013}, or
social media postings~\citep{Zhao2015,Kobayashi2016,Mishra2016}.
However, in many scenarios, one does not know the exact historical event times, but rather only the counts of events in specific time intervals.
Such \emph{interval-censored}~\citep{bernoulli1760essai} scenarios could be due to data availability -- say, for epidemics when the exact infection time for each individual is unknown and the number of hospital admissions per day is known --
or due to data privacy -- 
e.g., e-commerce websites do not expose who bought a particular item, but they will report rankings of most bought items.
Due to its lack of independent increments property (see \cref{ssect:the poisson and HP}), the Hawkes process cannot be directly fitted in the interval-censored scenario.
This work circumvents this shortcoming by building a novel point process, a set of tools, and approximations for fitting Hawkes processes within interval-censored data scenario.

There is prior work that deals with interval-censored data and Hawkes processes.
Here we highlight three families of approaches, and we present an in-depth review in \cref{subsec:lit-rev-hawkes-estimation}.
The first family is the integer-valued auto-regressive (INAR) models~\citep{Kirchner2016,Kirchner2017,Manolakis2019}.
These have been shown to yield approximations of the Hawkes model parameters after appropriate scaling. 
However, these works do not expose the connection between the Hawkes process's event-time and interval-censored formulations.
The second family of approaches refers to interval-censored data as panel count data~\citep{Sun2013, Zhu2014, Ding2018a, Moreno2020}.
These works typically concentrate on the non-parametric estimation of the intensity of Poisson processes from interval-censored data.
As far as we are aware, they do not specifically consider the Hawkes process and are not directly link to the current work apart from the presentation of the data.
The third family is the Hawkes Intensity Processes (HIP)~\citep{Rizoiu:2017}.
The HIP parameters are fitted by minimizing the squared error of the observed counts to the expected intensity of a Hawkes process.
However, HIP has several shortcomings (detailed in \cref{subsec:hip}), e.g., its fitting objective does not relate to the likelihood of a particular process, and it does not establish a link to the parameters of the underlying Hawkes process.
To our knowledge, no prior work has proposed a generalized treatment of Hawkes processes across event-time and interval-censored data.

\begin{figure}[tbp]
	\centering
	\includegraphics[width=.99\textwidth]{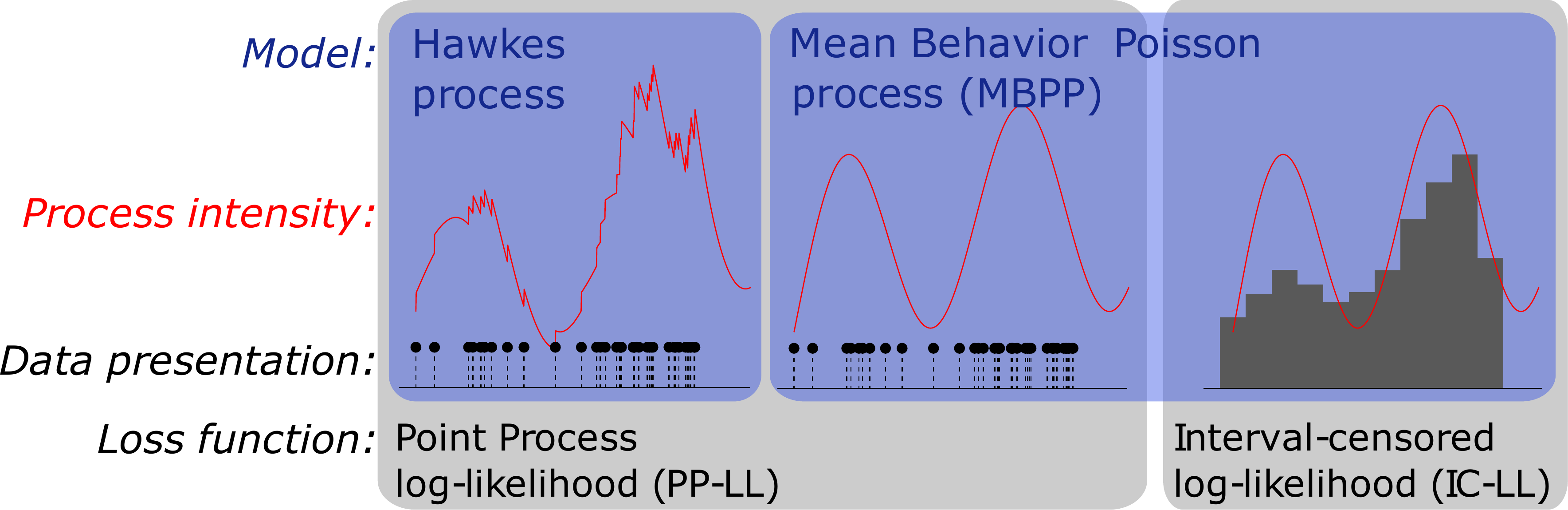}
	\caption{
		Illustration of the applicability of the Hawkes process and MBPP in event time (left gray bubble) and interval censored settings (right gray bubble). The Hawkes process is usable only when the data is in event time format, which can then be fitted using the standard point process log-likelihood (PP-LL). On the other hand, the MBPP can be used with both event time and interval-censored data by switching between PP-LL and the interval-censored log-likelihood (IC-LL).
	}
	\label{fig:interval-censored setting}
\end{figure}

In this paper, we build a new point process and a set of tools to operate with Hawkes processes on both event time and interval-censored data.
First, we introduce the Mean Behavior Poisson (MBPP) -- a non-homogeneous Poisson process, with a deterministic intensity function equal to the expectation of a Hawkes process intensity, taken over all possible realizations.
There is a one-to-one parameter equivalence between MBPP and the Hawkes process.
Next, we propose two approximations for the event intensity and compensator (the integral of the intensity) of MBPP, and we use them in an \emph{interval-censored log-likelihood} (IC-LL) which allows fitting the process on interval-censored data.
We show that IC-LL can be interpreted as a Bregman divergence, and we use it to draw a connection between the MBPP and the HIP~\citep{Rizoiu:2017} models.
Last, we empirically show in \cref{sec:synthetic_experiments} that MBPP allows one to retrieve the Hawkes model parameters for both interval-censored and event time data.
\cref{fig:interval-censored setting} illustrates the relation between Hawkes and MBPP.
When a realization is observed as event times, both Hawkes and MBPP can successfully fit the model using a \emph{point process likelihood} (PP-LL).
When the same realization is observed interval-censored, only MBPP can be used to retrieve the original parameters using the \emph{interval-censored likelihood} (IC-LL) (see \cref{sec:interval-censored point processes}).

We also study scenarios in which the exogenous events in the Hawkes process are separable from the endogenously generated events.
This is often the case in real-world applications --- such as our popularity prediction exercise in \cref{sec:real_world_experiments} where the shares and tweets driving the views of YouTube videos are observed separately from the views process.
In this setup, the exogenous events enter the process according to an unknown intensity function, and they can be observed either as event times or interval-censored.
We construct two novel exogenous functions -- the multi-impulse function for when the exogenous events are event times and the latent homogeneous Poisson process function for when the exogenous events are interval-censored -- and a new fitting procedure that only accounts for endogenous events.
Finally, we show that on real-world data our approaches outperform HIP~\citep{Rizoiu:2017}, the current state of the art in popularity prediction.

\textbf{The main contributions of this work are as follows}:
\begin{enumerate}[itemsep=0pt,topsep=0pt]
	\item[\textbf{C1}:] we show how to construct a non-homogeneous Mean Behavior Poisson process (MBPP) that approximates the mean behavior of the underlying Hawkes process (\cref{sec:mbp}).
	The likelihood of this process allows the use of a broader class of Bregman divergence based loss functions (\cref{sec:bregman-gen}).
	
    \item[\textbf{C2}:] we prove that the MBPP defines a causal, linear, continuous-time, time-invariant system with the impulse response computed as an infinite sum of convolutions (\cref{theorem:MBP-LTI}).
	This allows finding both a closed-form solution for the MBPP intensity (where it exists) and a numerical approximation (where it does not exist).

	\item[\textbf{C3}:] we propose a second approximation for the MBPP intensity (\cref{subsec:npllcalculate}), which is numerically more efficient, can exploit information about interval lengths and be used for predicting future event volumes (i.e., for forecasting).
	
	\item[\textbf{C4}:] we introduce a set of tools (exogenous functions, loss functions) 
	that allow fitting the parameters of a Hawkes process using MBPP,
when either the exogenous or the endogenous events (or both) are interval-censored (\cref{sec:processes_with_observed_exogenous_stimuli}).

	\item[\textbf{C5}:] on a real-world dataset containing information about tweets and YouTube views associated with YouTube videos, we show our approach outperforms HIP~\citep{Rizoiu:2017}, the current state-of-the art in popularity prediction (\cref{sec:real_world_experiments}).

\end{enumerate}
 
\section{Background and notations}
\label{sec:prereq}

In this section, we provide an overview of the different mathematical objects used in this paper.
First, in \cref{subsec:point-proces-def}, we introduce point processes and Poisson processes using random measures, and we discuss processes with independent increments.
Second, in \cref{ssect:the poisson and HP}, we introduce the Hawkes self-exciting point process, and we show that it does not have the independent increments property.
Third, in \cref{subsec:parameter estimation}, we discuss the parameter estimation of point processes.
Last, in \cref{subsec:hip}, we present the Hawkes intensity process --- a Hawkes-based interval-censored process and the previous work closest to the present paper.

In this paper, we use the following elementary notations --- for more specialized notations concerning point processes, please refer to \cref{tab:specnotation}. 
We denote \( \llbracket p \rrbracket = 1 \) if \( p \) is true and \( \llbracket p \rrbracket = 0 \) if \( p \) is false, following notation in \citet{Knuth1992}. 
The Laplace transform and inverse Laplace transform are defined as $\mathcal{L}\{\cdot\}$ and $\mathcal{L}^{-1}\{\cdot\}$ respectively. The Dirac delta function \( \delta(x) \) is defined by \( \int_{a}^{b} \delta(x) f(x) \, \dx = f(0) \) if \( f \) is continuous and \( a < 0 < b \). The convolution of functions \( f \) and \( g \) is defined as \( (f * g)(t) = \int_{-\infty}^{\infty} f(\tau)g(t-\tau) \, \mathrm{d}\tau \).

\begin{table}[tbp]
    \centering
    \caption{Notation used in this paper. Throughout this paper, we denote observed quantities using sans-serif font face.}
    \label{tab:specnotation}
    \begin{tabular}{lll}
        \toprule
        Symbol & Meaning & Defined/Introduced \\
        \midrule
        \( \HPintens(t) \) & Hawkes Process Intensity & \cref{eq:hawkes-intensity} \\
        \( \MBPintens(t) \) & Mean Behavior Poisson Intensity & \cref{eqn:intensity-renewal,eq:xi-definition} \\
        \( \HPcomp(t) \) & Hawkes Process Compensator & \cref{eq:compensator-def} \\
        \( \MBPcomp(t) \) & Mean Behavior Poisson Compensator & \cref{eq:hawkes compensator process} \\
        \( \HPCounts(t) \) & Hawkes Counting Process & \cref{eq:events-in-interval-prob} \\
        \( \MBPCounts(t) \) & Mean Behavior Poisson Counting Process & \cref{eq:MBP-interval-censored-likelihood} \\
        \( \HPObserVolume(x, y] \) & Observed counts in interval \( (x, y ] \) & \cref{eqn:approx-hip-objective,eqn:sec4_prob_example} \\
        \( \immiObserVolume(x, y] \) & Observed immigrant counts in interval \( (x, y ] \) & \cref{eq:latentbestparam} \\
        \( \OffspringVolume(x, y] \) & Observed offspring counts in interval \( (x, y ] \) & \cref{eq:bregman_end\obsTime_example} \\
        \( \bregdiv_{\varphi}(x) \) & Bregman Divergence & \cref{eq:bregman_def} \\
        \( \immiMDelta(t) \) & Multi-impulse exogenous function & \cref{eq:multi-impulse exogenous function} \\
        \( \immiIntensStep(t) \) & Latent Homogeneous Poisson process exog. func. & \cref{eq:latentexogfunc} \\
        \( \immiOccurTimeSet_{T} \) & Set of immigrant events up to time \( T \) & \cref{eq:edopointlikelihood} \\
        \( \offspringOccurTimeSet_{T} \) & Set of offspring events up to time \( T \) & \cref{eq:edopointlikelihood} \\
        \bottomrule
    \end{tabular}
\end{table} 
\subsection{Random measures, Point processes and Poisson processes}
\label{subsec:point-proces-def}

In this section, we briefly introduce point processes and Poisson processes using the notion of random measures.
For a more complete (yet accessible) introduction of point processes using random measures, we refer to \citep{grandell1977point}.
We opted for this rigorous definition of point processes for completeness reasons.
For a more intuitive and streamlined introduction using counting processes and intensity functions, we refer the reader to \citep{Rizoiu:2017c, Laub:2015}.

\textbf{Point processes as random measures.}
Let $(X,\mathcal{B})$ be a measurable state space and let $(\Omega,\mathcal{F},\Pr)$ be a probability space.
Let $M=M(X)$ be the space of all $\sigma$-finite measures on the sigma algebra $\mathcal{B}$.
Let $\mathcal{M}$ denote the smallest $\sigma$-algebra on $M$ with respect to which the function $\mu\mapsto \mu(B)$ is measurable for all $B\in\mathcal{B}$.
A \emph{random measure}~\citep{benes2007stochastic} on $X$ is a measurable map \( \eta : (\Omega,\mathcal{F},\Pr) \rightarrow (M, \mathcal{M}) \).
Let $N \subset M$ be the space of all $\sigma$-finite counting measures on $\mathcal{B}$.
Furthermore, let \( \mathcal{N} \) denote the restriction of \( \mathcal{M} \) when only considering $\sigma$-finite counting measures on \( \mathcal{B} \), \ie, \( \mathcal{N} = \{S \cap \mathcal{N} \mid S \in \mathcal{M} \} \).
A \emph{point process}~\citep{benes2007stochastic} on $X$ is a measurable map \( \hat{\eta} : (\Omega,\mathcal{F},\Pr) \rightarrow (N, \mathcal{N}) \).
A temporal point process is a random measure on the real line taking values in the non-negative integers or infinity \citep{BGS:2002}. 

More intuitively, temporal point processes are a family of stochastic processes $( \eventOccurTime_n )_{n \in \mathbb{N}_+}$, in which each random variable $\eventOccurTime_n$ represents the occurrence time for an event~\citep{Klebaner:2012}.
Such processes can also be specified as a \emph{counting process} $( N( t ) )_{t \geq 0}$,
where the random variable $N(t)$ represents the number of events that have occurred up to and including time $t$~\citep[pg. 59]{Ross:1995}.

\textbf{The Poisson process.}
Let $\HPcomp$ be a positive measure such that $\HPcomp(A)<\infty$ for every bounded set $A$. 
In the literature of martingale treatments of point processes~\citep{Jacobsen2006}, $\HPcomp$ is often denoted as the \emph{compensator} -- we also use this term throughout this paper. The Poisson process is a point process which is solely defined by the compensator.
In particular, the Poisson process is often characterised by (and can be defined by) the evaluation of the probability on disjoint sets. That is for every collection of disjoint, bounded Borel measurable sets $A_1,\cdots,A_k$, for a Poisson process, the $N(A_i)$ are independent Poisson random variables with rate $\HPcomp(A_i)$:
\begin{align} \label{eq:events-in-interval-prob}
    \Pr(N(A_i)=n_i, i=1,\cdots, k)=\prod^k_{i=1}\frac{\HPcomp(A_i)^{n_i}}{n_i !}e^{-\HPcomp(A_i)}.
\end{align}

Next, we use the notion of counting processes to define the Poisson process, \ie, specifying the compensator \( \HPcomp \). 
Suppose that $\HPcomp(A)=\HPintens l(A)$, where $l(A)$ is the Lebesgue measure of $A$ and $\HPintens>0$ a positive constant. 
We define the \emph{homogeneous Poisson process} $(N(t))_{t\ge 0}$ as

\begin{enumerate}
   	\item $N(0) = 0$ almost surely
	\item for any $s < t$, $N( t ) - N( s ) \sim \mathrm{Poisson}( \HPintens(t-s) )$
	\item for any $s < t \leq s' < t'$, $N( t ) - N( s ) \independent N( t' ) - N( s' )$,
\end{enumerate}
where $X \independent Y$ denotes that $X$ and $Y$ are independent.

We say a Poisson process is non-homogeneous when $\HPintens$ is no longer a constant, but a non-negative function.
Here, the compensator is redefined as $\HPcomp(A)=\int_A \HPintens(x)\, \dx$, and the \emph{non-homogeneous Poisson process} (NHPP) is

\begin{enumerate}[label={(\alph*)},itemsep=0pt]
	\item $N(0) = 0$ almost surely
	\item for any $s < t$, $N( t ) - N( s ) \sim \mathrm{Poisson}( \HPcomp( s, t ) )$
	\item for any $s < t \leq s' < t'$, $N( t ) - N( s ) \independent N( t' ) - N( s' )$.
\end{enumerate}

Associated with the counting process are two additional sequences of interest.
The first is the sequence of \emph{event histories} $( \HSf( t ) )_{t \geq 0}$,
where each random variable $\HSf( t )$
records the times of all events that have occurred up to and including time $t$.
Strictly speaking $\HSf( t )$ is a filtration, that is, an increasing sequence of $\sigma$-algebras~\citep{Laub:2015}.
The second is the sequence of \emph{conditional intensities} $( \HPintens( t \mid \HSf(t^-) ) )_{t \geq 0}$ --- denoted for simplicity as $\HPintens( t)$ ---, where each random variable
$\HPintens( t )$ measures the expected rate of events occurring in an infinitesimal window around the time $t$.
Formally, \citep[Equation 3.1]{Cox:1972},
\begin{align*}
	\HPintens( t ) &\defEq \lim_{h \to 0^+} \frac{1}{h} \cdot \E{}{N( t^{-} + h ) - N( t^{-} ) \mid \HSf( t^{-} )} \\
	&= \lim_{h \to 0} \frac{1}{h} \cdot \Pr\left( {N( t^{-} + h ) - N( t^{-} ) = 1 \mid \HSf( t^{-} )} \right),
\end{align*}
where the second equality holds for \emph{simple} processes wherein multiple events cannot occur at the same time, that is the jump size $N(t)-N(t^-)$ at each time $t$ can be either 0 or 1.
Finally, we can generalize the definition of the {compensator} $\HPcomp(s,t)$ as the integral of the intensity between two moments of time:
\begin{align} \label{eq:compensator-def}
    \HPcomp(s, t) \defEq \int_s^t\HPintens(\tau) \d\tau,
\end{align}
where $0 \leq s < t$.
We further shorthand $\HPcomp( t ) \defEq \HPcomp(0, t)$.

\textbf{Processes with independent increments and Markov processes.}
The process \( (N_{t})_{t \geq 0} \) is said to have the \emph{independent increments} property when, for any $u \le s\le t$, the random variables $N_t - N_s$ and $N_u$ are independent.
The process $N=(N_t)_{t\ge 0}$ is said to be a \emph{Markov process} with respect to $\HSf(t)$ if $\mathbb{E} [X_s \mid \HSf(t)]=\mathbb{E}[X_s \mid X_t]$ whenever $s\ge t$.
In simple terms, this means that the future of a Markov process is independent of its past. 
Moreover, if $N$ is a Markov process and conditional stationary, \ie, the conditional distribution is invariant to time translation, it results that it is a homogeneous Markov process.
That is, the transition probability function $\Pr(N(t+s)\le y \mid N(t)=x)$ is independent to $t$. 
Processes with independent increments are Markov processes. However, the converse is not true --- not every Markov process obeys the property of independent increments. 
One such example is the Ornstein-Uhlenbeck process~\citep{uhlenbeck1930theory}, a Markov process with the ``return to mean'' property -- intuitively, the process is dragged towards zero when it is far away from zero.
As a result, the increments are not independent and even negatively correlated.
Given the definition in \cref{eq:events-in-interval-prob}, the Poisson process has the independent increments property and, therefore, is a Markov process.

\subsection{Hawkes process}
\label{ssect:the poisson and HP}

The \emph{Hawkes process} is a temporal point process with the property of \emph{self-excitation}.
Intuitively, self-excitation means that the occurrence of one event increases the likelihood of further events in the near future.
The process was widely applied in analyzing social media~\citep{Kobayashi2016,Cao2017,Zhang2019}, 
earthquake aftershocks~\citep{Ogata1988,Helmstetter2002}, 
nuclear physics~\citep{Snyder:1991},
neuronal activity~\citep{apostolopoulou2019mutually}, online advertising~\citep{parmar2017forecasting} and finance~\citep{Bacry:2015}.

The Hawkes process is a doubly stochastic point process -- i.e., the event intensity function is also stochastic and depends upon the previous values of the process itself --
\citep{Hawkes:1971}:
\begin{align}
    \HPintens( t ) =& \, \immiIntens( t ) + \int_0^{t^-} \kernel( t - s ) \, \d N( s ) \nonumber \\ 
    =& \underbrace{\immiIntens( t )}_\text{exogenous} + \underbrace{\sum_{\eventOccurTime_i < t} \kernel( t - \eventOccurTime_i  )}_{\text{endogenous}}, \label{eq:hawkes-intensity}
\end{align}
where 
$\immiIntens( \cdot )$ is the (deterministic) \emph{background intensity} function which controls the arrival of external events into the system --- dubbed as \emph{immigrants};
$\immiIntens( \cdot )$ is also known as the exogenous intensity.
$\kernel( \cdot )$ is the \emph{kernel} function which controls the arrival of \emph{offspring} --- events spawned through self-excitement {--- by any one given event}.
{The arrival of offspring generated by any previous event is controlled by} the second term of \cref{eq:hawkes-intensity}, also known as the endogenous intensity. \( \{ \eventOccurTime_i \} \) denotes a realisation of the Hawkes process, with the immigrants and the offspring are collectively denoted as events.
In the rest of this paper, we use $\eventOccurTime_i$ interchangeably to denote both events and event times.

The Hawkes process is said to have an exponential kernel when 
$\kernel( \tau ) = \kappa \theta e^{-\theta \tau}$,
with 
$\kappa$ the \emph{excitation level}
and
$\theta > 0$ the \emph{decay rate};
similarly, the Hawkes process has a power-law kernel when $\kernel( \tau ) = \kappa (\tau + c)^{-(1+\theta)}$,
with $c$ a time-warping parameter to keep the kernel bounded when $\tau$ approaches zero, and $\kappa$ and $\theta$ have similar meanings as for the exponential kernel.
Note that by the endogenous summation in \cref{eq:hawkes-intensity}, $\HPintens$ depends on the history of events that have occurred up to time $t$. Thus given a fixed history \( \HSf(t^{-}) \), the intensity before time \( t \) is deterministically calculated by \cref{eq:hawkes-intensity}. For times greater of equal to \( t \), the intensity is a random variable.

\textbf{Relation to the Poisson process and processes with independent increments.}
The Hawkes process is not a Poisson process but rather a non-Markovian extension of the Poisson process. 
A Hawkes process can also be viewed as a cluster of Poisson processes~\citep{hawkes1974cluster}, in which each event spawns offspring following a non-homogeneous Poisson process of intensity $\kernel(t)$.
The link between events and their offspring induces the \emph{branching structure} (see \citep{Rizoiu:2017c} for more details).
Because the Hawkes process's intensity function depends on the previous events, it follows that the Hawkes process is not a Markov process and, as a result, does not have the independent increments property (unlike the Poisson point process).
In particular, the number of jumps in a given interval is thus a function of the number of jumps that happened earlier.
We will leverage this result in \cref{sec:interval-censored point processes} for fitting Hawkes processes in the interval-censored setting.

\subsection{Parameter estimation}
\label{subsec:parameter estimation}
We are given a sequence $\HSf(T) \defEq \{ \eventOccurTime_n \mid t_n \leq T \}$ of observed events up to an arbitrary maximum time \( T > 0 \), usually $T \defEq \max_n t_n$.
We try to find the set of parameters $\theta$ of a given point process which represent best the sequence.
As shown by \citet{Daley:2003}, this amounts to minimizing the negative log-likelihood with respect to $\theta$, which in turn maximizes the probability of events occurrence at the observed times $\eventOccurTimeSet_T$:
\begin{equation}
    \label{eqn:NHPP}
    \LCal( \theta; T ) \defEq - \sum_{t_n \in \eventOccurTimeSet_T} \log \HPintens( t_n; \theta ) + \int_{0}^T \HPintens( u; \theta ) \, \du. 
\end{equation}
As we can see, the log-likelihood requires observing the timing of events, which may not be the case in real application scenarios in which only volumes of events are observed (see \cref{sec:interval-censored point processes}).
We denote the standard negative log-likelihood loss in \cref{eqn:NHPP} as the \emph{point process log-likelihood} (PP-LL) loss function for the remainder of the paper.

\subsection{Hawkes intensity process}
\label{subsec:hip}

\citet{Rizoiu:2017} proposed the Hawkes intensity process (HIP), a Hawkes-derived process that operates in the interval-censored setting -- i.e., the set of observation times \( \obsTimeSet = \{ \obsTime_i \mid i = 0, \ldots, m, \, \obsTime_i \leq T \} \) partitions the time line into non-overlapping segments and only the number of events in each segment is observed (see \cref{sect:interval-censored and observation} for more details).
The HIP fitting objective is
\begin{equation}
	\label{eqn:hip-objective}
	\min_{\theta} \sum_{i = 1}^m ( \HPObserVolume(\eventObservTime_{i-1}, \eventObservTime_i] - \MBPintens( \eventObservTime_i; \theta ) )^2, 
\end{equation}
where $\HPObserVolume(\eventObservTime_{i-1}, \eventObservTime_i]$ {are} the observed number of events having occurred in the $i^\text{th}$ interval  $(\eventObservTime_{i-1}, \eventObservTime_i]$  and
$\MBPintens( t ) \defEq \E{}{ \HPintens( t ) }$ is the \emph{expected intensity} of the Hawkes process.

The key ingredient of HIP is being able to efficiently compute $\MBPintens$:
Theorem 2.1 of~\citep{Rizoiu:2017} provides a integral equation for $\MBPintens( t )$, namely
\begin{align}
   	\MBPintens( t ) &= \immiIntens( t )+ \int_0^t \kernel( s ) \cdot \MBPintens( t - s ) \, \d s \nonumber \\
   	&= \immiIntens( t ) + \int_0^t \kernel( t - s ) \cdot \MBPintens( s ) \, \d s. \label{eqn:intensity-renewal}
\end{align}
For completeness, we include a derivation of this result in \cref{app:proofs}.
To optimize \cref{eqn:hip-objective}, one must (approximately) solve this integral equation to compute each
$\MBPintens( t; \theta )$.
When $\immiIntens( t ) = \mu_0 + (\mu_1 - \mu_0) \cdot e^{-\gamma \cdot t}$,
and $\kernel$ corresponds to the exponential kernel, \citet{Dassios:2013} showed that
$$ \MBPintens( t ) = \mu_1 \cdot e^{-(\gamma - \alpha) \cdot t} + \frac{\mu_0 \cdot \gamma}{\gamma - \alpha} \cdot (1 - e^{-(\gamma - \alpha) \cdot t}). $$
For more general kernels, such as the power-law kernel, the closed-form solution for $\MBPintens$ often does not exist.
\citet{Rizoiu:2017} proposed to approximately solve \cref{eqn:intensity-renewal} by discretising time using the pre-defined observation endpoints $\{ \eventObservTime_i \}_{i = 0}^m$, yielding
\begin{equation}
    \label{eqn:discrete-hip}
	\hat{\MBPintens}[ \eventObservTime_i; \theta ] = \mu[ \eventObservTime_i ] + \sum_{s = 0}^{i - 1} \phi( \eventObservTime_i - \eventObservTime_s ) \cdot \hat{\MBPintens}[ \eventObservTime; \theta ], 
\end{equation}
where the bracket notation $\hat{\MBPintens}[ \cdot ]$ is used to explicate the discretization.
One can now use \cref{eqn:discrete-hip} to compute each $\hat{\MBPintens}[ \cdot; \theta ]$,
beginning with the base case $\hat{\MBPintens}[ 0; \theta ] = a$.
Equipped with this, one solves the approximation to the HIP objective:
\begin{equation}
    \label{eqn:approx-hip-objective}
	\min_{\theta} \sum_{i = 1}^m ( \HPObserVolume(\eventObservTime_{i-1}, \eventObservTime_i] - \hat{\MBPintens}[ \eventObservTime_i; \theta ] )^2. 
\end{equation}
\citet{Rizoiu:2017} showed that the solution to \cref{eqn:approx-hip-objective} has good empirical performance.
Conceptually, however, the approach has a few limitations:
\begin{enumerate}[label={(\alph*)},itemsep=0pt,topsep=0pt]
    \item[\textbf{L1}:] There is a mismatch between the objective and the observed data; the observations are of event \emph{counts}, while the objective measures closeness to the event \emph{rates}. Intuitively, one should work with $\E{}{N( t )}$ rather than $\E{}{ \HPintens( t ) }$.

	\item[\textbf{L2}:] The HIP objective is not derived as the likelihood of a particular process, raising uncertainty as to precisely what sense it approximates the Hawkes likelihood.
	In the limiting case where each observation interval is made infinitesimally small, for example, it is unclear whether the solution to \cref{eqn:hip-objective} approaches that of the standard Hawkes likelihood.

	\item[\textbf{L3}:] It relies on $\hat{\MBPintens}$, which is derived from an approximation to the underlying integral equation. This approximation is of unclear quality, and cannot handle intervals of unequal length.
\end{enumerate}

\section{The Mean Behavior Poisson process (MBPP)}
\label{sec:mbp}

In this section, we first introduce the Mean Behavior Poisson process (MBPP) -- a Poisson process whose intensity function is the expected event intensity of a Hawkes process of the same parameters (\cref{subsec:MBP-definition}).
Next, we show that its compensator $\MBPcomp(t)$ follows the same self-consistent equation as its intensity $\MBPintens(t)$ (\cref{subsec:MBP-compensator}).
Finally, we develop two methods to solve the self-consistent equation of MBPP for arbitrary exogenous functions (\cref{subsec:solving-mbp-equation}).

\begin{figure}[tbp]
	\centering
	\newcommand\myheight{0.11} \newcommand\mywidth{0.65} 

\includegraphics[width=\mywidth\textwidth]{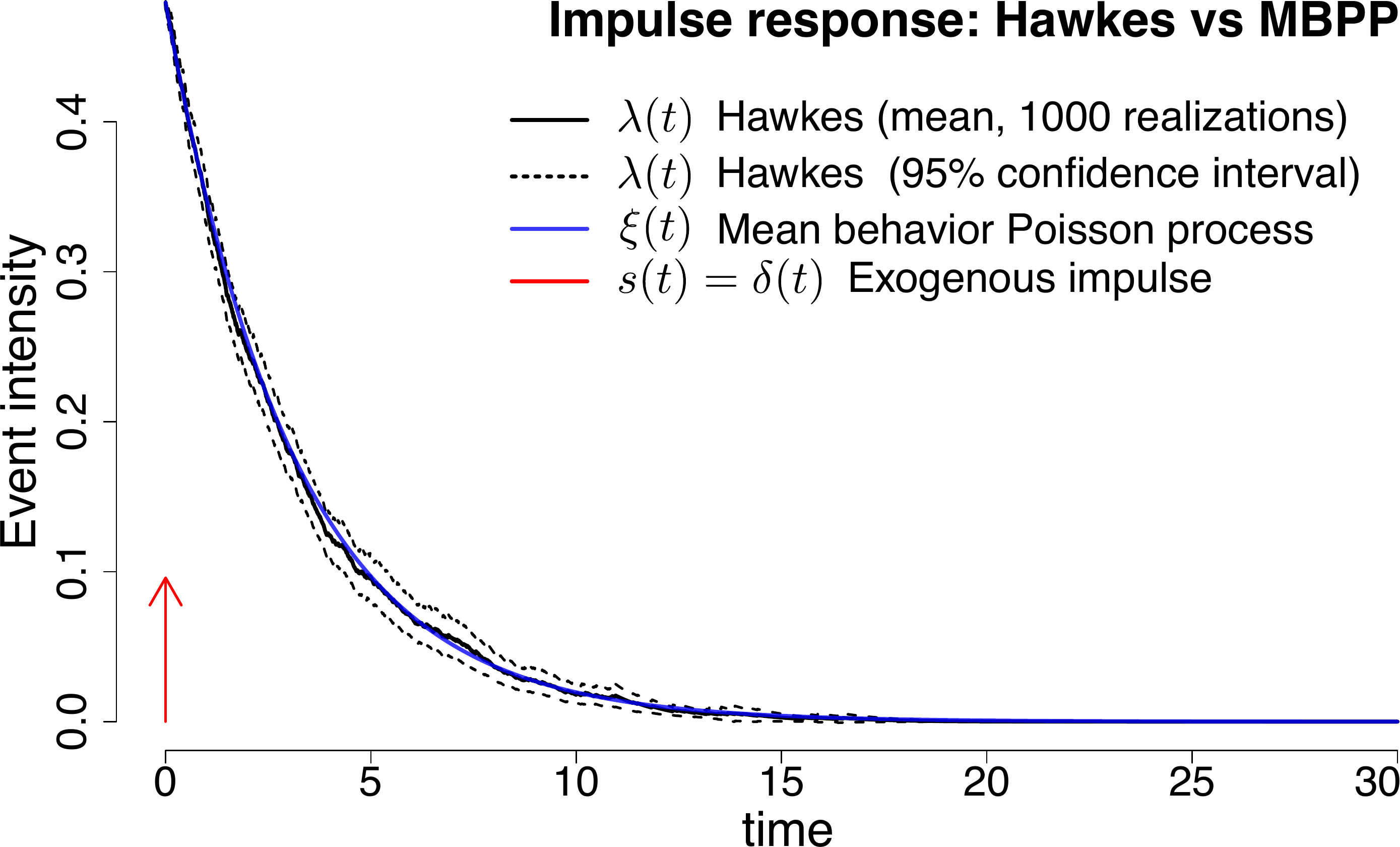}

	\caption{
		\textbf{Event intensity in Hawkes and MBPP} as a result of a single exogenous event (shown in \textcolor{red}{red}).
		We generate 1000 Hawkes realizations using an exponential kernel (\cref{eq:exp-kernel}) and the parameters $\kappa = 0.9, \theta = 1.15$.
		We show the mean and the 95\% confidence interval of the Hawkes intensity $\HPintens(t)$ (in \textcolor{black}{black}), and the (deterministic) MBPP intensity $\MBPintens(t)$ (in \textcolor{blue}{blue}).
		We observe that the mean Hawkes intensity over the 1000 realizations follows closely the MBPP intensity, empirically showing the relation between the two processes.
	}	
	\label{fig:hawkes_mbp_delta}
\end{figure} 
\subsection{Mean Behavior Poisson: model definition}
\label{subsec:MBP-definition}
The intensity of a Hawkes process (defined in \cref{eq:hawkes-intensity}) is a stochastic function -- i.e., at the same future time $t$ the conditional intensity of two Hawkes realizations with the same parameters can take very different values depending on the history of each realization.
Here we introduce a new process --- dubbed the \emph{Mean Behavior Poisson process} (MBPP) --- which captures the expected behavior of all the Hawkes realizations defined by the same set of parameters.
Formally, given a Hawkes process we define MBPP as the non-homogeneous Poisson process with a deterministic intensity $\MBPintens(t)$ --- the expected Hawkes intensity over all possible realizations (\cref{eqn:intensity-renewal}):
\begin{align}  \label{eq:xi-definition}
    \MBPintens(t) \defEq \E{\HSf_{t}}{\HPintens(t)} &= \immiIntens(t) + \int_0^t \MBPintens(\tau)\kernel(t-\tau) \, \d\tau \nonumber \\
    &= \immiIntens(t) + (\MBPintens * \kernel) (t)
\end{align}
where $\HPintens(t)$ denotes the event intensity in the Hawkes process (\cref{eq:hawkes-intensity}); 
$\immiIntens(t)$ denotes exogenous intensity function;
$\phi(t)$ denotes the kernel function of the Hawkes process and
$*$ is the convolution operator.
For the remainder of the paper, when convenient, we suppress the functional dependence on parameters \( \theta \) for intensity functions (\( \HPintens(t) \), \( \MBPintens(t) \)) and compensators (\( \HPcomp(t) \), \( \MBPcomp(t) \)).

We observe that the same objects fully define the MBPP intensity $\MBPintens(t)$ as well as the Hawkes intensity $\HPintens(t)$;
that is, the exogenous intensity $\immiIntens(t)$ and the kernel function $\kernel(t)$.
We, therefore, say that Hawkes processes have a one-to-one correspondence to MBPP. 
In \cref{fig:hawkes_mbp_delta} we visually denote the relation between the two processes by plotting the event intensity in Hawkes and in MBPP.
We sample 10,000 realizations of Hawkes, and we plot its mean event intensity and its 95\% confidence interval.
We also plot the intensity of the mean behavior Poisson, which visually overlaps closely with the mean Hawkes intensity.

\textbf{MBPP and HIP.}
Visibly, MBPP shares the same equation with HIP~\citep{Rizoiu:2017} (see \cref{eqn:intensity-renewal}). 
However, their usages are different: HIP discretizes \cref{eqn:intensity-renewal} and aims to fit volumes of events, while \cref{eq:xi-definition} defines the intensity of a novel point process that, as far as we are aware, has not been studied in the literature.
In the next section, we study an important quantity of MBPP -- its \emph{compensator} -- and we show that it satisfies a nearly identical integral equation to the intensity equation.

\subsection{The compensator of MBPP}
\label{subsec:MBP-compensator}

We define $\MBPcomp(t)$ as the expectation of the Hawkes compensator (see \cref{eq:compensator-def}) over all possible realizations:
\begin{equation*}
    \MBPcomp(t) = \E{\HSf_{t}}{\HPcomp(t)}
\end{equation*}
We show that the $\MBPcomp(t)$ is the compensator of the MBPP using the Fubini theorem and passing the expectation inside the integral since $\HPintens(t)$ is bounded (see~\citet[Theorem 8.4]{Klebaner:2012}):
\begin{align}\label{eq:hawkes compensator process}
    \MBPcomp(t) & = \int_0^{t} \E{\HSf_{t}}{\HPintens(z)}\,  \d z \overset{cf.(\ref{eq:xi-definition})}{=} \int_0^{t}\MBPintens(z)\,  \d z \nonumber \\
    &= \int_0^{t} \left[ \immiIntens(z) + \int_0^z \MBPintens(z-y) \kernel(y)\,  \d y \right]\,  \d z \nonumber \\
    &= \int_0^{t} \immiIntens(z)dz + \int_{z = 0}^{t} \int_{y = 0}^z \MBPintens(z-y) \kernel(y)\,  \d y\,  \d z \nonumber \\
    &\stackrel{(a)}{=}  \int_0^{t} \immiIntens(z)dz + \int_{y = 0}^{t}\int_{z = y}^{t} \MBPintens(z-y) \kernel(y)\,  \d z\,  \d y \nonumber\\
    &= \int_0^{t} \immiIntens(z)dz + \int_0^{t}\kernel(y) \int_y^{t} \MBPintens(z-y)\,  \d z\,  \d y \nonumber\\
    &\stackrel{(b)}{=}  \int_0^{t} \immiIntens(z)dz + \int_0^{t}\kernel(y) \int_0^{t-y} \MBPintens(x)\,  \d x\,  \d y \nonumber\\
    &\stackrel{(c)}{=} \immiComp(t) + \int_0^{t}\MBPcomp(t-y) \kernel(y)\,  \d y \nonumber \\
    &= \immiComp(t) + (\MBPcomp * \kernel) (t)
\end{align}
where $(a)$ is obtained by reversing the order of the integrals and computing the new integration boundaries;
$(b)$ is obtained by performing the change of variable $x = z - y$;
in $(c)$ we apply $\MBPcomp(t-y) \defEq \int_0^{t-y} \MBPintens(x) \, \dx$;
and $\immiCounts(t) \defEq \int_0^{t} \immiIntens(z) \, \d z $ is the cumulative immigrant intensity up to time $t$.

A cursory observation of \cref{eq:hawkes compensator process} shows a distinct resemblance with \cref{eq:xi-definition}:
MBPP's compensator $\MBPcomp(t)$ follows a very similar self-consistent equation as its intensity $\MBPintens (t)$, with the compensator's value at time $t$ being a function of its values at every previous time $\tau < t$, decayed by the corresponding kernel function value.
The similarity in the functional form of the two equations implies that similar methods can be applied to deduce their analytical solutions, detailed in the next section.

\textbf{Computational requirements.}
Both MBPP's compensator and Hawkes process intensity functions are quadratic to compute.
The Hawkes process intensity function (unraveling the summation in \cref{eq:hawkes-intensity}) is quadratic in the number of events in the history $\HSf_{t}$. 
The MBPP's compensator function is quadratic in the number of interval-censored intervals (see \cref{sec:interval-censored point processes}).
However, calculating MBPP's compensator function is more computationally effective than computing the intensity function of a Hawkes process. 
Typically, the number of observational intervals is significantly smaller than the number of events.

\subsection{Solving the Mean Behavior Poisson equation}
\label{subsec:solving-mbp-equation}

To fit the MBPP (for both the event time and interval-censored setups), we need to have its intensity available in a tractable form.
In this section, we propose two methods to solve \cref{eq:xi-definition}, for arbitrary $\immiIntens( t )$ and $\kernel(t)$ functions.
This type of equation is known as a Volterra integral equation of the second kind~\citep{arfken1985mathematical}, which admits a solution only in particular cases.
The first method is based on the Laplace transform and has the downside of applying only to functions for which the transform and its inverse exist.
The second method is a novel solution to the Volterra equation of the second kind, which employs notions of distribution theory first to compute the system's impulse response and afterward the solution.

\textbf{Laplace transform.}
One natural approach to solving the MBPP equation is using the Laplace transform.
Let $\mathcal{L}\{\cdot\}$ and $\mathcal{L}^{-1}\{\cdot\}$ denote the Laplace and the inverse of the Laplace transform respectively.

\begin{theorem}
    \label{thm:laplacesolve}
    Given the functions $\immiIntens( t )$ and $\kernel(t)$,
    then
\begin{align}\label{eqn:MBP-equation-laplace-solution}
        \MBPintens(t) = \mathcal{L}^{-1}\left\{ \frac{\mathcal{L}\left\{\immiIntens(t)\right\}(\omega)}{1-\mathcal{L}\left\{\phi(t)\right\}(\omega)} \right\}(t)
    \end{align}
    is a closed-form solution for \cref{eq:xi-definition} if the Laplace and inverse Laplace transform exist.
\end{theorem}
The derivation of \cref{eqn:MBP-equation-laplace-solution} is shown in \cref{app:proofs}.

This Laplace transform-based method has several drawbacks. 
First, it requires that the Laplace transform exists for both $\immiIntens( t )$ and $\kernel(t)$.
This requirement is not always fulfilled by the exogenous intensity functions $\immiIntens(t)$ that we construct in \cref{def:multi-impulse}.
These involve Dirac delta functions, for which the Laplace transform does not exist in the classical sense.
Note that, while there are extensions for the Laplace transform to generalized functions (we refer the curious reader to~\citet[Ch.~8]{VanDijk2013}), their direct application to solving \cref{eq:xi-definition} is not obvious, and they are outside the scope of this work.

Second, the method requires that the inverse Laplace function exists, which does not hold when $\kernel$ involves a power-law or a Rayleigh time decay function.
This is particularly limiting since the power-law kernel ($\kernel(t) = \kappa (t + c)^{-(1+\theta)}$) has been shown to be the best performing in applications involving social media data~\citep{Rizoiu:2017,Mishra2016};
the Rayleigh function ($\kernel(t) = \kappa \frac{t}{\theta^2} e^{- \frac{t^2}{2 \theta^2}}$) is widely used in epidemics modeling as it allows for a period of increasing intensity corresponding to disease incubation~\citep{Unwin2021}.

Lastly, this method involves re-computing \cref{eqn:MBP-equation-laplace-solution} every time that $\immiIntens(t)$ changes -- with all the difficulties stated above occurring at each recomputation -- rendering the method difficult to apply when the $\immiIntens(t)$ changes often, or when it is not known before-hand.

\textbf{The impulse response solution.}
We propose a novel method to solve the MBPP equation in \cref{eq:xi-definition}, which deals with the shortcomings of the Laplace transform-based method.
We first show that the \cref{eq:xi-definition} defines a linear, continuous-time, time-invariant system~\citep{phillips2003signals}, and we discuss a method to derive its impulse response solution (\cref{theorem:MBP-LTI}).
Next, we show how to use the impulse response to easily compute the solution for any arbitrary $\immiIntens(t)$ (\cref{corollary:arbritrary-s}).

\begin{theorem} \label{theorem:MBP-LTI}
    Let $H$ be a Hawkes process whose behavior is defined by the kernel function $\kernel(t)$ satisfying $\int_0^{\infty} \phi(t) \d t < 1$, and let $M$ be the mean behavior Poisson process (MBPP) associated (\cref{eq:xi-definition}) with $H$, with the event intensity denoted by $\MBPintens(t)$, and defined in \cref{eq:xi-definition}.
    Then the intensity of $M$ defines a causal, linear, continuous-time, time-invariant system of input $\immiIntens(t)$ and output $\MBPintens(t)$.
    The system is completely characterized by its impulse response defined by
    \begin{align*} 
    E(t) = \delta(t) + h(t), \text{ where }
    h (t) = \sum^{\infty}_{n=1}\kernel^{\otimes n} (t)
    \end{align*}
    where $\otimes n$ signifies $n$ times convolution.
\end{theorem}

\begin{proof}
We structure the proof into two parts.
In the first part, we show that MBPP is an LTI system, and in the second part, we derive its impulse response.

For the first part of the proof, \citet[Corollary~2.2]{Rizoiu:2017} have shown that \cref{eq:xi-definition} defines a causal, linear, continuous-time, time-invariant system of input $\immiIntens(t)$ and output $\MBPintens(t)$.
For completeness reasons, we outline this proof here.
It is easy to see that linearity holds by multiplying both sides of \cref{eq:xi-definition} by the same constant. 
The time-invariance property states that the response to a time-delayed input is identical and similarly time-delayed: if $\immiIntens(t)\rightarrow\MBPintens(t)$ then $\immiIntens(t-t_0)\rightarrow\MBPintens(t-t_0)$.
Using \cref{eq:xi-definition} we compute the response for $\immiIntens(t-t_0)$ as follows:
\begin{align*}
    \MBPintens(t-t_0) &= \immiIntens(t-t_0) + \int_0^t \MBPintens(\tau) \kernel(t-t_0 \;-\tau)\, \d\tau \\
    &\stackrel{(a)}{=} \immiIntens(t') + \int_0^{t'+t_0} \MBPintens(\tau) \kernel(t' \;-\tau)\, \d\tau \\
    &\stackrel{(b)}{=} \immiIntens(t') + \int_0^{t'} \MBPintens(\tau) \kernel(t' \;-\tau)\, \d\tau
+ \underbrace{\int_{t'}^{t'+t_0} \MBPintens(\tau) \kernel(t' \;-\tau)\, \d\tau}_{= 0} \\
    &\stackrel{(c)}{=} \immiIntens(t') + \int_0^{t'} \MBPintens(\tau) \kernel(t' \;-\tau)\, \d\tau
\end{align*}
\noindent where at $(a)$ we make a change of variable $t'=t-t_0$;
at line $(b)$ we write the integral into two parts, i.e., $(0,t)$ and $(t', t'+t_0)$;
at line $(c)$ we observe that $\kernel(t)$ is a causal function, i.e., $\kernel(t)=0$ for $t<0$, or $\kernel(t'-\tau)=0$ for $\tau>t'$, and the integral term from $t'$ to $t'+t_0$ disappears.
$\MBPintens(t)$ is also a causal function, which together with the linearity and time-invariance properties renders the MBPP intensity system a causal, linear time-invariant (LTI) system.

In the second part of this proof, we concentrate on the impulse response function corresponding to the MBPP system.
The impulse response of a dynamic system is its output when presented with a unit impulse input signal. 
The impulse response completely characterizes the system, as the response to a linear combination of time-delayed impulses is a linear combination of time-delayed impulse responses.
For continuous time dynamic systems, the impulse is denoted by the Dirac delta function $\delta(t)$.
We denote as $E(t)$ the impulse response of MBPP, with $E(t) = \delta(t) + h(t)$.
From \cref{eq:xi-definition}, it follows that
\begin{align*}
    E(t) &= \delta(t) + \int_0^t E(\tau) \kernel(t-\tau) \, \d\tau \\
    &= \delta(t) + (E * \kernel) (t) \\
    &\stackrel{(a)}{=} \delta(t) + ( (\delta + h) * \kernel) (t) \\
    &= \delta(t) + (\delta * \kernel) (t) + (h * \kernel) (t) \\
    &\stackrel{(b)}{=} \delta(t) + \underbrace{\kernel (t) + (h * \kernel) (t)}_{h(t)}
\end{align*}
\noindent where $(a)$ is obtained by substituting $E(t)$ with $\delta(t) + h(t)$;
$(b)$ is obtained by noticing that the Dirac $\delta$ function is the neutral element for the convolution operation.
Consequently and by dropping the time notation, we obtain
\begin{align*}
    h &= \kernel  + h * \kernel \\
    &= \kernel  + ( \kernel  + h * \kernel)* \kernel \\
    &= \kernel + \kernel^{\otimes 2} + h * \kernel^{\otimes 2} \\
    &= \kernel + \kernel^{\otimes 2} + \kernel^{\otimes 3} + h * \kernel^{\otimes 3} \\
    &= \ldots \\
    &= \sum^{\infty}_{n=1}\kernel^{\otimes n}.
\end{align*}
For the above equation to hold, the infinite sum must converge.
That is, we require that $\underset{n \rightarrow \infty}{\lim} \kernel^{\otimes n} = 0$. Observe that
\begin{align*}
    \underset{n \rightarrow \infty}{\lim} \int_0^\infty \kernel^{\otimes n}(t)\d t &\stackrel{(a)}\leq \underset{n \rightarrow \infty}{\lim} \left( \int_0^\infty \kernel(t) \d t \right)^n \\
    &\stackrel{(b)}= 0,
\end{align*}
where for (a) we use Young's convolution inequality on $\kernel(t)$ under the $L^1$ norm $\int_0^\infty (\cdot) \d t$, and for (b) we use $\int_0^{\infty} \phi(t) \d t < 1$. Since $\underset{n \rightarrow \infty}{\lim} \int_0^\infty \kernel^{\otimes n}(t)\d t \geq 0$, we have  $\underset{n \rightarrow \infty}{\lim} \int_0^\infty \kernel^{\otimes n}(t)\d t = 0$. By $L^1$ convergence, we then have $\underset{n \rightarrow \infty}{\lim} \kernel^{\otimes n} = 0$.
\end{proof}

\begin{corollary} \label{corollary:arbritrary-s}
Let $M$ be a mean behavior Poisson process associated with a Hawkes process of kernel $\kernel(t)$, and let $E(t) = \delta(t) + h(t)$ be its impulse response function.
    Let $\immiIntens(t)$ be an arbitrary generalized function denoting an external stimuli.
    Then $\MBPintens(t) = (E * \immiIntens) (t)$ is a solution to the MBPP equation (\cref{eq:xi-definition}).
\end{corollary}

\begin{proof}
\begin{align*}
    \MBPintens &= E * \immiIntens = (\delta + h) * \immiIntens = \delta * \immiIntens + h * \immiIntens \\
    &\stackrel{(a)}{=} \immiIntens + E * \kernel * \immiIntens = \immiIntens + (E * \immiIntens) * \kernel \\
    &= \immiIntens + \MBPintens * \kernel 
\end{align*}
where $(a)$ is obtained by replacing $h = E * \kernel$, knowing that $E = \delta + h$ and that $E$ is the impulse response, i.e. a solution to $E = \delta + E * \kernel$. \end{proof}

\bheader{For example}. Suppose
\begin{equation}
    \label{eq:dassio response}
    \immiIntens( t ) = \kappa \theta + (u_0 - \kappa \theta)  e^{-\theta t}
\end{equation}
and $\kernel(t)$ is exponentially time-decaying.
These particular forms for the $\immiIntens(t)$ and $\kernel(t)$ were proposed by \citet{Dassios:2013}, for which they also find a closed-form solution.
We define the exponential kernel as
\begin{equation} \label{eq:exp-kernel}
    \kernel(t) = \kappa \theta e^{-\theta t} \cdot \llbracket t > 0 \rrbracket.\end{equation}

Following \cref{theorem:MBP-LTI} we first compute the impulse response as $E(t) = \delta(t) + h(t)$, with $h\in L^1(\mathbb{R})$ given by $h=\sum^{\infty}_{n=1}\phi^{\otimes n}$. 
For $t \geq 0$ and $\kernel$ defined in \cref{eq:exp-kernel} we have 
\begin{align*}
    \kernel^{\otimes 2}&=(\kernel*\kernel)(t)\\
    &=\int^t_0 \kappa\theta e^{-\theta\tau} \kappa\theta e^{-\theta(t-\tau)}\, \d\tau\\
    &= \kappa^2\theta^2 e^{-\theta t}t\end{align*}
and one can easily show that 
\begin{equation} \label{eq:infinite-conv-sum-analytical-solution}
    \kernel^{\otimes n}(t)= \kappa^n \theta^n e^{-\theta t}\frac{t^{n-1}}{(n-1)!}
\end{equation}
By summing up, we compute $h(t)$ as 
\begin{align*}
    h(t) =\sum^{\infty}_{n=1} \phi^{\otimes n}(t)&=e^{-\theta t} \kappa\theta\sum^{\infty}_{n=0}\frac{( \kappa\theta t)^n}{n!}\\
    &= e^{-\theta t} \kappa\theta e^{ k\theta t}\\
    &= \kappa\theta e^{( \kappa-1)\theta t}, \text{        for } t \geq 0
\end{align*}
Since $\kernel(t < 0) = 0$, we have $h(t < 0) = 0$.
It follows that the impulse response for MBPP with the exponential kernel defined in \cref{eq:exp-kernel} is 
\begin{align} \label{eq:impulse-response-exp}
    E(t) &= \delta(t) + \kappa\theta e^{(\kappa-1)\theta t} \cdot \llbracket t > 0 \rrbracket
\end{align}
From \cref{corollary:arbritrary-s} it immediately follows that the solution for \cref{eq:xi-definition} with $\immiIntens( t ) = \kappa \theta + (u_0 - \kappa \theta)  e^{-\theta t}$ is
\begin{align}
    \MBPintens(t) &= (E * \immiIntens) (t) = \immiIntens(t) + (h * \immiIntens) (t) \nonumber \\
    &= \frac{\kappa \theta}{1-\kappa} \left( 1 - e^{-(1 - \kappa) \theta t} \right) + u_0 e^{-(1 - \kappa) \theta t} \label{eq:Dassios-response}
\end{align}
The full details of the derivation (and derivations of other exogenous functions) are shown in \cref{app:mbp_intensity}.

Note that the infinite convolutions sum has a closed-form solution only for particular kernel functions (such as in \cref{eq:infinite-conv-sum-analytical-solution}).
For all the other kernels, one can apply an approximation based on finite sums.
For example, knowing that $\underset{n \rightarrow \infty}{\lim} \kernel^{\otimes n} = 0 $ when $\int_0^{\infty} \phi(t) \d t < 1$ (see above), we can use the approximation $h \approx \sum^{T}_{n=1}\kernel^{\otimes n}$, where $T$ controls the approximation error and the computation speed.
Note also that an analytical solution for the convolution $h * h$ does not exist for most kernels, requiring numerical solutions.

\begin{figure*}[htb]
	\centering
	\newcommand\myheight{0.11} \newcommand\mywidth{0.33} 

\subfloat[]{
		\includegraphics[width=\mywidth\textwidth]{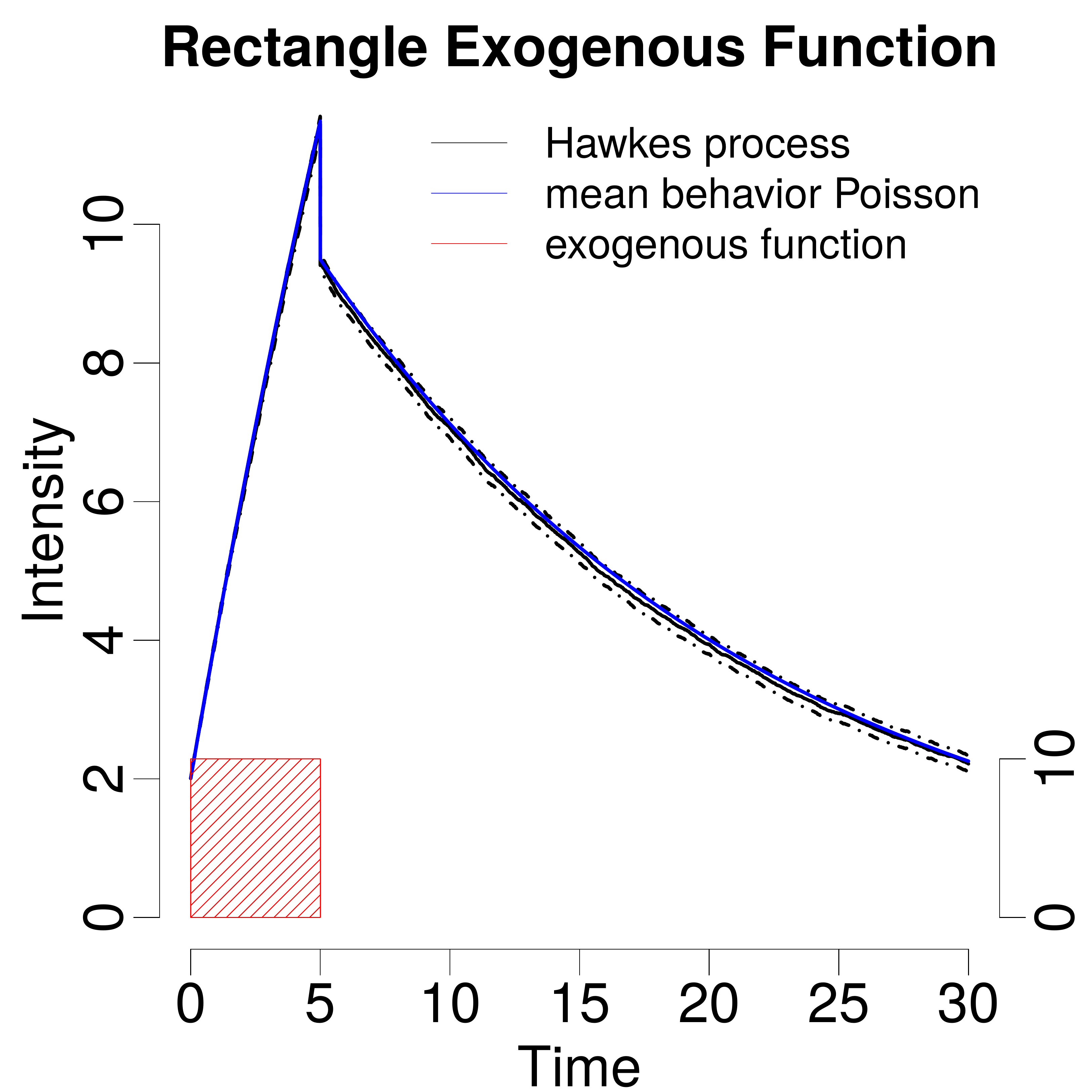}\label{subfig:hawkes_mbp_rect}}\subfloat[]{
		\includegraphics[width=\mywidth\textwidth]{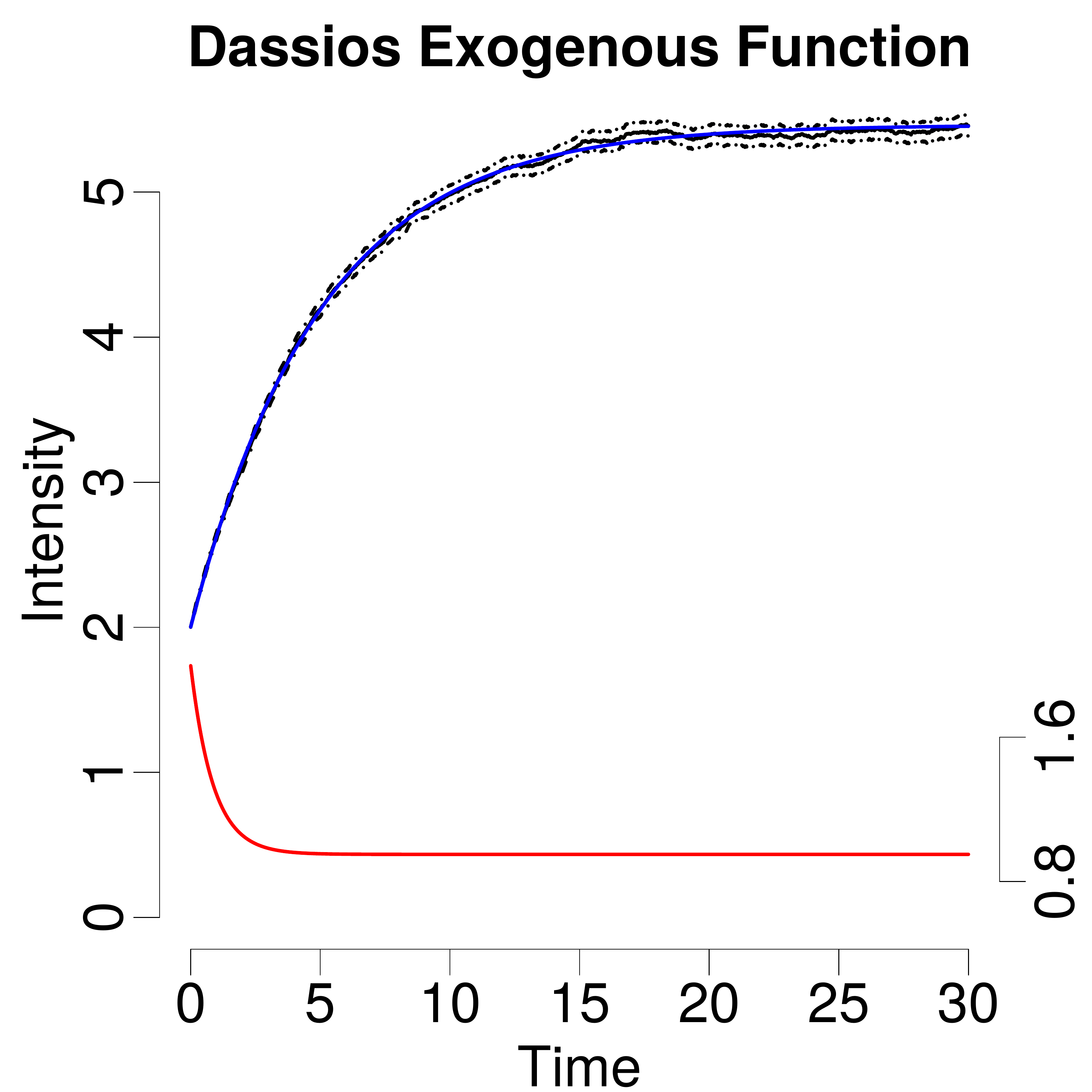}\label{subfig:hawkes_mbp_dassios}}\subfloat[]{
		\includegraphics[width=\mywidth\textwidth]{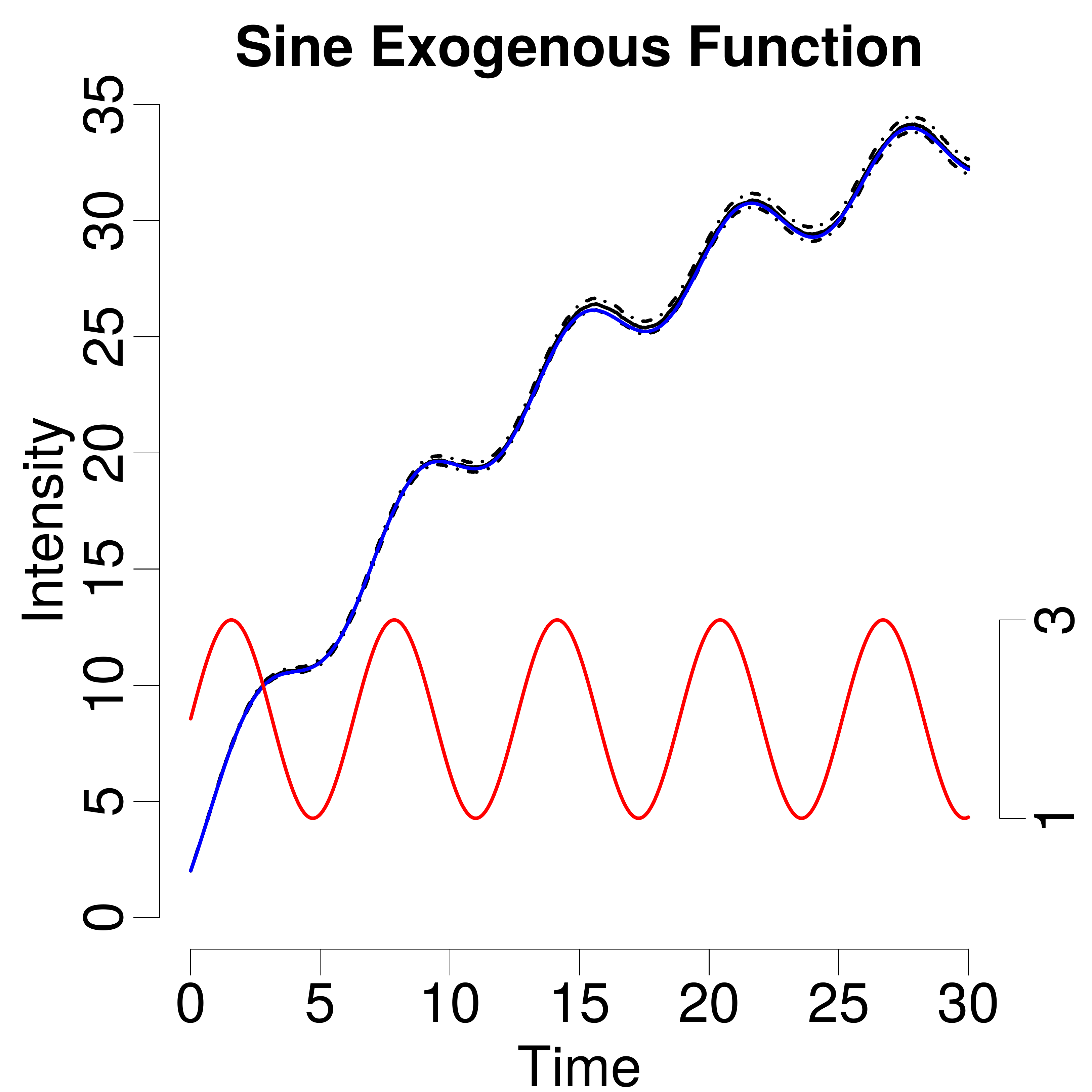}\label{subfig:hawkes_mbp_sin}}\caption{
		The intensity of 10,000 simulated realizations of a Hawkes process with exogenous function as a rectangle response, Dassios response~\citep{Dassios:2013}, and sine response. 
		Here the \textcolor{red}{red} part is the exogenous function both in the Hawkes process and MBPP. 
		The \textcolor{blue}{blue} line is the MBPP intensity, the solid black line is the mean intensity of the corresponding Hawkes process, and the black dashed lines indicate the 95\% confidence interval.
	}
	\label{fig:hawkes_mbp_other}
\end{figure*} %
 
\section{Interval-censored point processes}
\label{sec:interval-censored point processes}

In this section, we discuss fitting MBPP in the interval-censored setting (detailed in \cref{sect:interval-censored and observation}),
also known in some areas of the literature as panel count data~\citep{Sun2013,Ding2018a,Moreno2020}.
The Hawkes process generally cannot be fit in the interval-censored setting as it does not have the {\it independent increment} property.
Given the correspondence in parameters between MBPP and the Hawkes process, we can instead fit an MBPP using the interval-censored log-likelihood (IC-LL) as the MBPP does have the independent increment property.
Finally, we present a method for numerically calculating the MBPP compensator function, which is needed to calculate the IC-LL loss function.

\subsection{Interval-censored Hawkes processes}
\label{sect:interval-censored and observation}

\begin{figure}[tbp]
	\centering

	\includegraphics[width=0.75\textwidth]{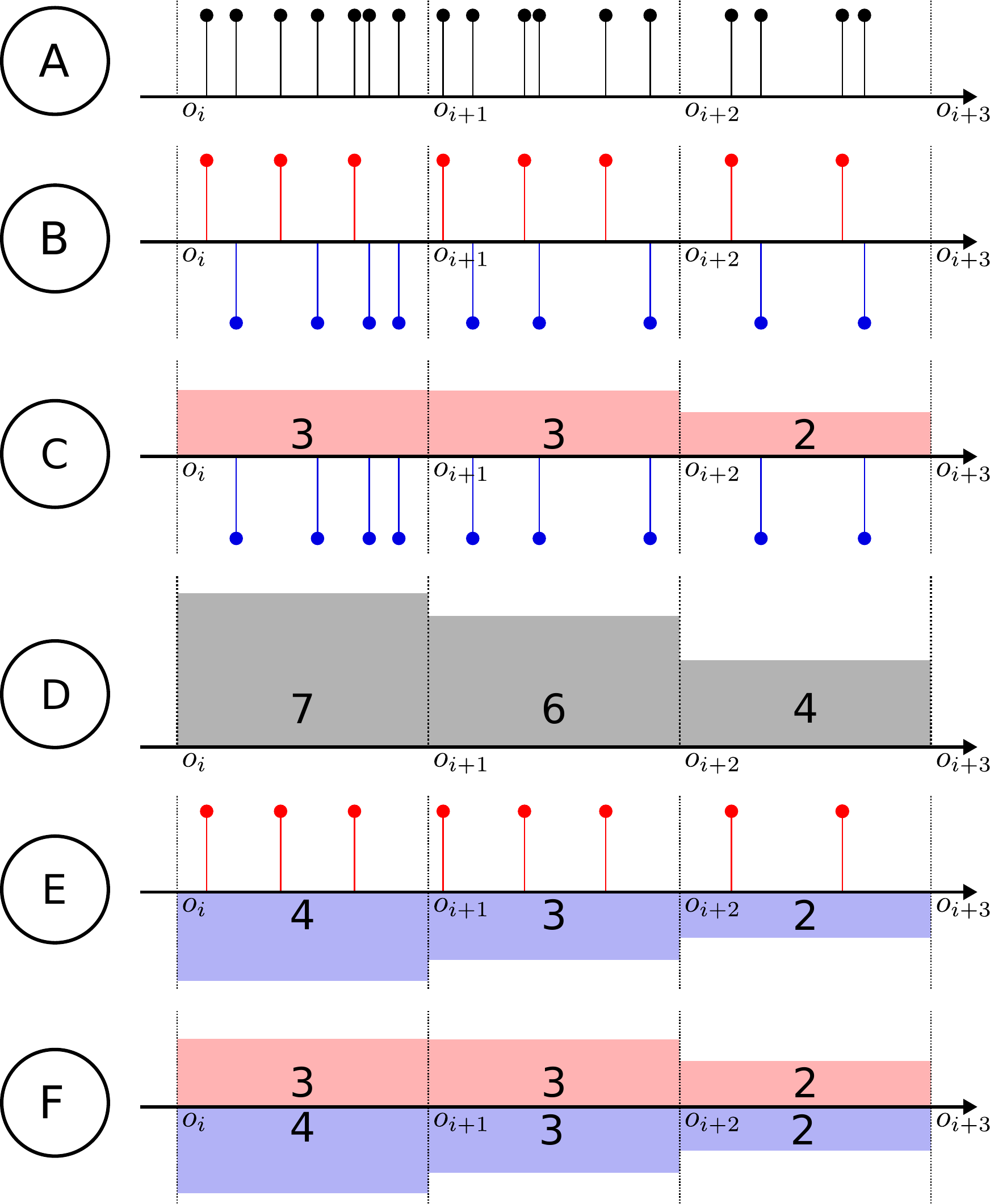}
	\caption{
		\textbf{The same point process realization is shown in six scenarios of data presentation -- from event times, to separable event times and interval censored data.}
		Each row corresponds to a setting in~\cref{tab:settings}. 
		\textcolor{red}{Red components} correspond to exogenous events; and \textcolor{blue}{blue components} correspond to endogenous events.
		Black color indicates that we cannot distinguish between exogenous and endogenous.
		$\obsTime_{i}$ denote the observation times based on which the data is interval-censored.
		The lollipops markers represent event times, while the shaded areas represent event counts within intervals.
	}
	\label{fig:interval-censored}
\end{figure}

Suppose that we have the set of observation times \( \obsTimeSet = \{ \obsTime_i \mid i = 0, \ldots, m, \, \obsTime_i \leq T \} \) which partitions into non-overlapping segments the time extent that the point process is observed \( t \in (0, T] \). 
Without loss of generality, we assume that the time points in \( \obsTimeSet \) are ordered by index, \( \obsTime_{0} = 0 \), and \( \obsTime_{m} = T \). 
Further, for each segment defined by the half-open interval of consecutive points \( ( \obsTime_{i - 1}, \obsTime_i ] \) we observe the number of events which occurs in the interval, denoted by \( \HPObserVolume( \obsTime_{i - 1}, \obsTime_i ] \). We define this setting as the \emph{interval-censored setting}. \cref{fig:interval-censored} illustrates the interval-censored setting for a simulated Hawkes process realization. 
Panel A shows the realization as event times;
panel D shows the same Hawkes process realization as interval-censored.

If we want to estimate the parameters of a Hawkes process in this setting, we cannot use the standard point process log-likelihood in \cref{eqn:NHPP} as it is defined for event times. 
Instead, we aim to maximize the joint probability of the observed event volumes in each interval:
\begin{equation}
    \label{eqn:sec4_prob_example}
    \max_{\theta} \Pr(N(\obsTime_{i - 1}, \obsTime_i] = \HPObserVolume( \obsTime_{i - 1}, \obsTime_i ], i \in \{ 1, \ldots, m \}),
\end{equation}
where \( \HPCounts(\obsTime_{i-1}, \obsTime_{i}] = \HPCounts(\obsTime_{i}) - \HPCounts(\obsTime_{i-1}) \) is defined as the number of events in a realization on the half-open interval and \( \theta \) are the Hawkes process's parameters.

However, there are several complications for computing \cref{eqn:sec4_prob_example} for Hawkes processes. 
First, unlike the homogeneous and homogeneous Poisson processes, the Hawkes process does not have the independent increments property. 
As a result, we cannot evaluate \cref{eqn:sec4_prob_example} as we did in \cref{eq:events-in-interval-prob}, i.e., the product of disjoint Poisson distributions with the expected number of events set to the point process compensator.
An alternative method for breaking down the Hawkes process is to decompose the process into the event count of each generation of offspring, formally known as a Galton-Watson branching process~\citep{hawkes1974cluster}. 
This provides a decomposition of Borel distributions~\citep{borel1942emploi}.
\citet{kong2020exploiting} derive the closed-form solution for the case where the offspring distribution is evaluated at \( t \rightarrow \infty \). 
However, there is no known analytical solution for computing the Borel distributions at finite times. 
A Monte-Carlo solution is available, but the computation cost does not scale well~\citep{OBrien2020}.
Further, there is no known analytic or numeric solution for computing the conditional factorization of the joint probability of all offspring generation Borel distributions.
Thus, there is no closed-form solution for a Hawkes process likelihood function in the interval-censored setting in the general case.

\textbf{Using the MBPP equivalence.}
Given the above difficulties, we propose in this section a solution based on the MBPP.
In a nutshell, we replace the Hawkes process with an MBPP of equivalent parameters and fit the MBPP parameters in the interval-censored setting.
Finally, we use the obtained MBPP parameters to approximate the Hawkes parameters. 
Our approach has two sources of information loss (that we study on synthetic data in \cref{sec:synthetic_experiments}), which prevent obtaining the exact original Hawke parameters.
First, 
the information contained in the interval-censored version of the data -- when the observation periods are not infinitesimal in size -- will necessarily have less information than the original event time data. 
Second, while having parameter equivalence, MBPP and Hawkes are different models -- for example, MBPP is not a self-exciting model.
As a result, the fitted MBPP parameters can only be approximations of the true Hawkes parameters.

\subsection{Negative Poisson log-likelihood: Interval-censored loss}
\label{subsec:IC-LL-loss-function}

Unlike the Hawkes process, the MBPP introduced in \cref{sec:mbp} is an non-homogeneous Poisson process. 
Thus the MBPP has the independent increment property, allowing \cref{eqn:sec4_prob_example} to be evaluated as the likelihood of disjoint Poisson distributions~\citep[Chapter 2.4]{Daley:2003}. 
The corresponding negative log-likelihood function of the MBPP is dubbed as the \emph{interval-censored log-likelihood (IC-LL)}.
The IC-LL was used with non-homogeneous Poisson processes by prior literature~\citep{Ding2018a}.
In this work, we use IC-LL as a building block to estimate the parameters of Hawkes models (via MBPP) and to build new loss functions via the Bregman divergence interpretation (see \cref{subsec:bregman_divergence}).

The Poisson distributions which compose the negative Poisson log-likelihood are defined with respect to the MBPP compensator function \( \MBPCounts(t) \), as per \cref{eq:events-in-interval-prob}. 
Similar to the counts \( N(t) \), define \( \MBPCounts (\obsTime_{i-1}, \obsTime_{i}] = \MBPCounts(\obsTime_{i}) - \MBPCounts(\obsTime_{i-1})\). 
Note that \( M(t) \) is stochastic, as it is the counting process of the MBPP.
Using this joint distribution, we can calculate the negative log-likelihood for an MBPP by decomposing over each observation interval.
\begin{align}
        \LCal(\theta)
        &= - \log \mathbb{P}(\MBPCounts(\obsTime_{i-1}, \obsTime_{i}] = \HPObserVolume(\obsTime_{i-1}, \obsTime_{i}], \, i = 1, \ldots, m) \nonumber\\
        &= - \log \prod_{i=1}^{m} \mathbb{P}(\MBPCounts(\obsTime_{i-1}, \obsTime_{i}] = \HPObserVolume(\obsTime_{i-1}, \obsTime_{i}]) \nonumber\\
        &= - \log \prod_{i=1}^{m} \left( \frac{\MBPcomp(\obsTime_{i-1}, \obsTime_{i}]^{\HPObserVolume(\obsTime_{i-1}, \obsTime_{i}]}}{\HPObserVolume(\obsTime_{i-1}, \obsTime_{i}]!} \exp(- \MBPcomp(\obsTime_{i-1}, \obsTime_{i}]) \right) \nonumber\\
        &= - \sum_{i=1}^{m} \left( \HPObserVolume(\obsTime_{i-1}, \obsTime_{i}] \cdot \log \MBPcomp(\obsTime_{i-1}, \obsTime_{i}] - \log (\HPObserVolume(\obsTime_{i-1}, \obsTime_{i}]!) - \MBPcomp(\obsTime_{i-1}, \obsTime_{i}] \right) \nonumber\\
        &= \sum_{i=1}^{m} \MBPcomp(\obsTime_{i-1}, \obsTime_{i}] - \sum_{i=1}^{m} \HPObserVolume(\obsTime_{i-1}, \obsTime_{i}] \log \MBPcomp(\obsTime_{i-1}, \obsTime_{i}] + \sum_{i=1}^{m} \log (\HPObserVolume(\obsTime_{i-1}, \obsTime_{i}]!) .\nonumber \\ \label{eq:MBP-interval-censored-likelihood}  
\end{align}

Notably, as we are aiming to optimise a parameter set \( \theta \) of MBPP, the constant which only depends on the event counts \(\sum_{i=1}^{m} \log (\HPObserVolume(\obsTime_{i-1}, \obsTime_{i}]!) \) can be ignored. As such, we further rewrite the above loss function to the following equivalent loss function when minimised.

\begin{proposition}
\label{prop:nhpp-censored}
The interval-censored log-likelihood (IC-LL) for an MBPP with corresponding compensator \( \MBPcomp \) under interval-censored data is
\begin{equation}
    \label{eq:nhpp-censored}
    \begin{aligned}
        \npll( \theta ) = \sum_{i=1}^{m} \MBPcomp(\obsTime_{i-1}, \obsTime_{i}; \theta] - \sum_{i=1}^{m} \HPObserVolume(\obsTime_{i-1}, \obsTime_{i}] \log \MBPcomp(\obsTime_{i-1}, \obsTime_{i}; \theta].
    \end{aligned}
\end{equation}
\end{proposition}

Note that partial similar results exist in literature. 
For example, \citep[Sec. 2.1]{Ding2018a} derive a similar log-likelihood of a multivariate non-homogeneous point process.
In this work, we further link IC-LL with the
Bregman divergence (see~\cref{subsec:connection-bregman-icll}), and we show how other loss functions can be obtained by varying the generator function.

\subsection{Calculating the negative Poisson log-likelihood}
\label{subsec:npllcalculate}

The loss functions proposed above both require the computation of the compensator \( \MBPcomp(\obsTime_{i-1}, \obsTime_{i}] \), as defined in \cref{eq:hawkes compensator process}. 
From the definition of the compensator over an interval, we have
\begin{equation}
    \label{eq:compensatoroverinterval}
    \begin{aligned}
        \MBPcomp(\obsTime_{i-1}, \obsTime_{i}]
        &\defEq \int_{\obsTime_{i-1}}^{\obsTime_{i}} \MBPintens(s) \, ds \\
        &= \int_{0}^{\obsTime_{i}} \MBPintens(s) \, ds - \int_{0}^{\obsTime_{i-1}} \MBPintens(s) \, ds \\
        &= \MBPcomp(\obsTime_{i}) - \MBPcomp(\obsTime_{i-1}).
    \end{aligned}
\end{equation}

We explore the two cases where we either have an analytical solution for the MBPP compensator or when we have to use numeric approximation to evaluate it.

\textbf{Analytical solution.}
As we have foreshadowed in \cref{subsec:MBP-compensator}, the compensator of MBPP follows the same functional form as the MBPP intensity function \( \MBPintens \). 
Thus, we can use the same analytical methods for solving the compensator evaluated at a point.
Specifically, we can apply \cref{corollary:arbritrary-s} to derive an analytical solution:

\begin{corollary}[to \cref{theorem:MBP-LTI}]
    Let \( \kernel(t) \) be the kernel of a Hawkes process and \( E(t) = \delta(t) + h(t) \) be its impulse response function, then for \( \immiComp(t) \) a closed-form solution for the MBPP compensator over a half-open interval \( (x, y] \) is
    \begin{equation}
        \MBPcomp(x, y] = (E * S)(y) - (E * S)(x).
    \end{equation}
\end{corollary}
\begin{proof}
    Follows immediately from application of \cref{eq:compensatoroverinterval} and \cref{corollary:arbritrary-s}.
\end{proof}

\textbf{Numerical approximation.}
The cases in which the compensator \( \MBPcomp(\obsTime_{i-1}, \obsTime_{i}] \) cannot be solved analytically are generally due to the lack of a closed-form solution for the $h$ term in \cref{theorem:MBP-LTI}.
While we can approximate $h$ using a finite sum of numerical convolutions (see discussion in \cref{subsec:solving-mbp-equation}), here we consider a numeric lower bound approximation of the MBPP compensator -- see discussion at the end of this section for additional advantages.
Specifically, we approximate the MBPP counting process with a set of approximation points \( \{ \approxpoint_{j} \}_{j=0}^{\approxupper} \), where \( \approxpoint_{0} = 0 \), \( \approxpoint_{j-1} < \approxpoint_{j} \), and \( 0 < \approxpoint_{j} < T \) for \( j \in \{ 1, \ldots, \approxupper \} \). Consider the following lower-bound on the MBPP counting process using \( \approxupper \) approximation points:
\begin{equation}
    \MBPCounts(t) \geq \MBPCounts^{-}_{D}(t) \defEq \sum_{j=1}^{\approxupper} \MBPCounts(\approxpoint_{j-1}, \approxpoint_{j}] \cdot \llbracket \approxpoint_{j} < t \leq \approxpoint_{\approxupper} \rrbracket.
\end{equation}
Considering that the event count $\MBPCounts(t)$ is a monotonically increasing function, the lower bound $\MBPCounts^{-}_{D}(t)$ is intuitively a piecewise constant function, which gets updated at each $\approxpoint_{j}$, and is constant in between approximation points.
For equidistant approximation points
\begin{equation*}
	\approxpoint_{j} = \frac{T}{D}i,
\end{equation*}
it can be shown that we can become arbitrarily accurate to \( \MBPCounts(t) \) as the number of approximation points increases. 

\begin{proposition}\label{prop:approx_converge}
    As \( D \rightarrow \infty \), then \( \MBPCounts^{-}_{D}(t) \rightarrow \MBPCounts(t) \).
\end{proposition}

Using this lower bound counting process, we can approximate the compensator of MBPP, which follows a self-consistent equation (as per \cref{eq:hawkes compensator process}).
We consider the following formulation of the compensator which uses the equivalence between compensator and expected counts \( \MBPcomp(t) = \mathbb{E}[\MBPCounts(t)] \) for non-homogeneous Poisson processes:
\begin{equation}
    \begin{aligned}
        \MBPcomp(t) 
&= \immiCounts(t) + \int_{0}^{t} \MBPcomp(t-y) \kernel(y) \d y \\
        &= \immiCounts(t) + \int_{0}^{t} \MBPcomp(y) \kernel(t-y) \d y \\
        &= \immiCounts(t) + \int_{0}^{t} \mathbb{E}[\MBPCounts(y)] \kernel(t-y) \d y.
    \end{aligned}
\end{equation}

We now define the following lower-bound compensator using the counting process approximation \( \MBPCounts^{-}_{D}(t) \):
\begin{equation}
    \label{eq:numapproxlowerbound}
    \begin{aligned}
        \MBPcomp^{-}(t) 
        &\defEq \immiCounts(t) + \int_{0}^{t} \mathbb{E}[\MBPCounts^{-}_{D}(y)] \kernel(t-y) \d y \\
        &= \immiCounts(t) + \int_{0}^{t} \mathbb{E}\left[\sum_{j=1}^{\approxupper} \MBPCounts(\approxpoint_{j-1}, \approxpoint_{j}] \cdot \llbracket \approxpoint_{j} < y \leq \approxpoint_{\approxupper} \rrbracket\right] \kernel(t-y) \d y \\
        &= \immiCounts(t) + \sum_{j=1}^{\approxupper} \mathbb{E}\left[\MBPCounts(\approxpoint_{j-1}, \approxpoint_{j}]\right] \int_{0}^{t}  \llbracket \approxpoint_{j} < y \leq \approxpoint_{\approxupper} \rrbracket \kernel(t-y) \d y \\
        &\stackrel{(a)}{=} \immiCounts(t) + \sum_{j=1}^{\bar{\approxupper}(t)} \mathbb{E}\left[\MBPCounts(\approxpoint_{j-1}, \approxpoint_{j}]\right] \int_{\approxpoint_{j}}^{\min(t, \approxpoint_{\approxupper})} \kernel(t-y) \d y,
    \end{aligned}
\end{equation}
where \( \bar{\approxupper}(t) = \max \{ i \in \{ 1, \ldots, \approxupper \} : \approxpoint_{i} < t \} \). Step \( (a) \) follows from simultaneously changing the integral upper-bound and the summation bound with respect to the Iverson brackets.

An upper-bound approximation equivalent is given by
\begin{equation}
    \label{eq:numapproxupperbound}
    \MBPcomp^{+}(t)  = 
    \immiCounts(t) + \sum_{j=1}^{\bar{\approxupper}(t)+1} \mathbb{E}\left[\MBPCounts(\approxpoint_{j-1}, \approxpoint_{j}]\right] \int_{\approxpoint_{j-1}}^{\min(t, \approxpoint_{\approxupper})} \phi(t-y) \d y.
\end{equation}

The negative Poisson log-likelihood loss function requires to compute the compensator over intervals \( \MBPcomp(\obsTime_{i-1}, \obsTime_{i}] \),
which we approximate using the compensator lower-bound \( \MBPcomp^{-}_{D} \):
\begin{equation} \label{eq:approx-compensator}
    \MBPcomp(\obsTime_{i-1}, \obsTime_{i}] \approx \MBPcomp^{=}_{D}(\obsTime_{i-1}, \obsTime_{i}] = \MBPcomp_{D}^{-}(\obsTime_{i}) - \MBPcomp^{-}_{D}(\obsTime_{i-1}).
\end{equation}
By using \cref{eq:numapproxlowerbound}, we can compute this approximation as follows,

\begin{proposition}
    \label{prop:compensator-approx}
    The numeric approximation of the compensator over an interval \( \MBPcomp(\obsTime_{i-1}, \obsTime_{i}] \approx \MBPcomp^{=}_{D}(\obsTime_{i-1}, \obsTime_{i}] \) can be computed as
    \label{prop:numapprox}
    \begin{equation}
        \begin{aligned}
        &\MBPcomp^{=}_{D}(\obsTime_{i-1}, \obsTime_{i}]
        = \immiCounts(\obsTime_{i}) - \immiCounts(\obsTime_{i-1}) 
        + \sum_{j=1}^{\bar{\approxupper}(\obsTime_{i})} \mathbb{E}\left[\MBPCounts(\approxpoint_{j-1}, \approxpoint_{j}]\right] \int_{\obsTime_{i-1} - \approxpoint_{j}}^{\obsTime_{i} - \approxpoint_{j}} \phi(y) \d y \\
        &\hspace{4cm}
        + \sum_{j=\bar{\approxupper}(\obsTime_{i-1}) + 1}^{\bar{\approxupper}(\obsTime_{i})} \mathbb{E}\left[\MBPCounts(\approxpoint_{j-1}, \approxpoint_{j}]\right] \int_{0}^{\obsTime_{i-1} - \approxpoint_{j}} \phi(y) \d y.
        \end{aligned}
    \end{equation}
\end{proposition}

The proof of \cref{prop:numapprox} can be found in \cref{app:proofs}.

Leveraging \cref{prop:numapprox} has several advantages.
First, it allows approximating the compensator's value for functional forms that do not allow for closed-form solutions of the infinite convolution proposed in \cref{subsec:solving-mbp-equation}.
Second, it allows leveraging observed history (the number of observed events in past time intervals) to compute future compensator values.
We use the latter property in our real-world experiments in~\cref{sec:real_world_experiments}.

\section{Bregman generalization of IC-LL}
\label{sec:bregman-gen}

Previously in \cref{sec:interval-censored point processes}, we introduced the NPLL loss function for fitting in interval-censored data settings. 
This section shows that the IC-LL loss function can be interpreted as a Bregman divergence-- a generalized notion of distances. 
We consider alternative loss functions that arise when changing the convex generator function of the Bregman divergence. 
This leads to a generalization and correction of the HIP loss function \citep{Rizoiu:2017}. In particular, we present specific conditions in which the generalization simplifies to the HIP loss function.

\subsection{Bregman divergences}
\label{subsec:bregman_divergence}

We first begin with a brief introduction of Bregman divergences.
Bregman divergences are a widespread tool for designing and analyzing algorithms within machine learning. Bregman divergences have been used to construct generalizations of algorithms to use different definitions of distances, i.e., clustering~\citep{banerjee2005clustering}, boosting~\citep{collins2002logistic}, and computational geometry~\citep{nielsen2007bregman}. 
Similarly, we will use Bregman divergences to generalize IC-LL by interpreting the loss function as the generalized Kullback-Leibler divergence (KL-divergence) between compensator values and interval-censored event counts.

A Bregman divergence is a generalized notion of distances between points. Suppose we are given a strictly convex generator \( \varphi : S \rightarrow \mathbb{R} \), where \( \textrm{dom}(\varphi) =  S \) is a convex set \( S \subseteq \mathbb{R}^{d} \) and \( \varphi(\cdot) \) is differentiable on the relative interior \( \textrm{ri}(S) \). Then the Bregman divergence \( \bregdiv_{\varphi} : S \times \textrm{ri}(S) \rightarrow [0, \infty) \) is defined as \citep{banerjee2005clustering}:
\begin{equation}\label{eq:bregman_def}
    \bregdiv_{\varphi}(\bf{x}, \bf{y}) \defEq \varphi(\bf{x}) - \varphi(\bf{y}) - (\bf{x} - \bf{y})^{\top} \nabla\varphi(\bf{y}).
\end{equation}
A wide selection of popular divergences can be obtained by picking the appropriate generator function. In particular, we consider two specific convex generators for Bregman divergences.

\bheader{KL-divergence.} The KL-divergence can be obtained by considering the generator function
\begin{equation}
    \label{eq:generator_kl}
    \varphi({\bf x}) \doteq \sum_{i=1}^{d} x_{i} \log x_{i},
\end{equation}
yielding, for \( \bf x, \bf y \in \mathbb{R}^{+} \),
\begin{equation}
    \label{eq:divergence_kl}
    {\rm KL}({\bf x}, {\bf y}) \doteq \bregdiv_{\varphi}({\bf x}, {\bf y}) = \sum_{i=1}^{d} x_{i} \log \frac{x_{i}}{y_{i}} - {\bf 1}^{\top}({\bf x} - {\bf y}).
\end{equation}

\bheader{Squared loss.} The squared loss function is obtained if we consider a squared Euclidean norm generator function
\begin{equation}
    \label{eq:generator_sse}
    \hat\varphi({\bf x}) \doteq \Vert {\bf x} \Vert^{2},
\end{equation}
yielding
\begin{equation}
    {\rm SSE}({\bf x}, {\bf y}) \doteq \bregdiv_{\hat\varphi}({\bf x}, {\bf y}) = \Vert {\bf x} - {\bf y} \Vert^{2}.
\end{equation}

\subsection{Connections to IC-LL}
\label{subsec:connection-bregman-icll}

Given a single observation sequence, the IC-LL loss function is equivalent to the generalized KL-divergence with an additional constant factor. As a result, IC-LL is equivalent to the KL-divergence under minimization. First, let us define \( {\bf C} \defEq (\HPObserVolume(\obsTime_{0}, \obsTime_{1}], \ldots, \HPObserVolume(\obsTime_{m-1}, \obsTime_{m}]) \) and \( \mathbf{\MBPcomp}(\theta) \defEq (\MBPcomp(\obsTime_{0}, \obsTime_{1}], \ldots, \MBPcomp(\obsTime_{m-1}, \obsTime_{m}]) \) as an observed volume of interval-censored events and corresponding interval compensator values using parameters \( \theta \), respectively.

Starting from a simple observation, we express the IC-LL using the KL-divergence.
\begin{proposition}
    \label{prop:bregman}
    The IC-LL loss function is the given by
    \begin{equation}
        \label{eq:bregman_icll}
        \npll(\theta) = {\rm KL}({\bf C}, {\bf \MBPcomp}(\theta)) + \Gamma({\bf C}),
    \end{equation}
    where \( \Gamma({\bf C}) \) is a constant only dependent on the observed events.
Furthermore,
    \begin{equation}
        \arg \min_{\theta} \npll(\theta) = \arg \min_{\theta} {\rm KL}({\bf C}, {\bf \MBPcomp}(\theta)).
    \end{equation}
\end{proposition}

The KL-divergence term in \cref{eq:bregman_icll} can be rewritten as a Bregman divergence as follows:
\begin{equation}
    \label{eq:general_minimise}
    \bregdiv_{\varphi}({\bf C}, {\bf \MBPcomp}(\theta)) + \Gamma({\bf C}),
\end{equation}
where \( \varphi \) is given by \cref{eq:generator_kl}. Thus the natural question arises of what happens when we switch the choice of Bregman divergence. 

We now consider switching the KL-divergence generator function to that of the squared loss function given by \cref{eq:generator_sse}. This gives a sum of the squared error loss function:
\begin{align}
    \label{eq:sse_loss}
    \LCal_{\rm{SSE}} &= {\rm SSE}({\bf c}, {\bf \MBPcomp}) + \Gamma({\bf C})
    \nonumber \\
    &= \sum_{i=1}^{m} (\HPObserVolume(\obsTime_{i-1}, \obsTime_{i}] - \MBPcomp(\obsTime_{i-1}, \obsTime_{i}; \theta])^{2} + \Gamma({\bf C}).
\end{align}

This SSE loss function is actually a generalized version of the HIP loss function~\citep{Rizoiu:2017}; differing in only that the intensity function \( \MBPintens \) is used instead of \( \MBPcomp \) and the additional constant \( \Gamma({\bf C}) \).

A notable connection is the bijection between (regular) Bregman divergences and (regular) exponential families~\citep[Theorem 6]{banerjee2005clustering}. 
In particular, the choice of Bregman divergence dictates what distribution the data is assumed to be sampled from for each observational interval. 
For example, by choosing the KL-divergence \cref{eq:bregman_icll}, each observational interval is assumed to be Poisson distributed, giving the IC-LL loss function. 
By considering SSE divergence \cref{eq:sse_loss}, we are assuming that the event volume is Gaussian distributed; with constant standard deviation.

One way of establishing a connection between the two loss functions is by considering the SSE loss function as an approximation of the KL-divergence, where each interval is Poisson distributed. Concretely, let us define a Poisson distribution with true rate \( \MBPcomp(\obsTime_{i-1}, \obsTime_{i}] \) for an arbitrary event interval. Then suppose we have \( N \) many i.i.d. volume observations over this interval \( \HPObserVolume_{1}(\obsTime_{i-1}, \obsTime_{i}], \ldots, \HPObserVolume_{N}(\obsTime_{i-1}, \obsTime_{i}] \). It follows that given the empirical mean of these observations \( \hat\HPObserVolume(\obsTime_{i-1}, \obsTime_{i}] \), by the central limit theorem, in the limit we have that
\begin{align*}
    \hat\HPObserVolume(\obsTime_{i-1}, \obsTime_{i}] \sim \N(\MBPcomp(\obsTime_{i-1}, \obsTime_{i}], \MBPcomp(\obsTime_{i-1}, \obsTime_{i}] / N)
\end{align*}
as a Poisson distribution has variance equal to its rate.
Thus, the averaging of event sequence provides an SSE loss function by the bijection between Bregman divergences and exponential families~\citep[Theorem 6]{banerjee2005clustering}. 
Consequently, we expect the KL loss and SSE loss functions to perform similarly when the number of samples is large.

Notably, optimal compensator parameters for our Bregman loss functions are sensitive to the generator function chosen. This contrasts the unconstrained minimization of the expectation of Bregman divergence, which was shown to have a unique minimum invariant of the generator function~\citep[Proposition 1]{banerjee2005clustering}. Thus the functional form of the compensator (the chosen triggering kernel) can be considered an additional constraint to the Bregman divergence minimization problem.

\subsection{HIP as an MBPP approximation} \label{subsec:connections_to_ObsTime_hip}

We note that the SSE loss function given by \cref{eq:sse_loss} is similar to the HIP loss function~\citet{Rizoiu:2017}. 
This section shows that the SSE loss function is a generalization of the HIP loss function, which is not theoretically justified in the general case. 
In particular, it incorrectly attempts to match the point process intensity with interval-censored volumes. 
However, by using the framework of Bregman divergence losses presented in~\cref{subsec:bregman_divergence}, we prove that the HIP loss function is equivalent to an SSE loss function with three additional assumptions:
(1) that observation intervals are unit sized; 
(2) the MBPP intensity function is constant over the interval; and 
(3) a discrete convolution approximation is used on the intensity function.

One simple, practical method of using Hawkes processes with interval-censored data is to convert intervals into event times through uniformly sampling within each interval. 
The average intensity (over multiple samples of intervals) would amount to the MBPP intensity function (by \cref{eq:xi-definition}). 
In essence, this simplification assumes that events are equally likely in the intervals; 
thus, the simplification is equivalent to assuming that the underlying MBPP intensity function is constant over each interval.

In addition, by assuming that we have unit time observation intervals, we can show that the compensator function over that interval is equivalent to the intensity function of the MBPP process.

\begin{lemma}
    \label{lem:hip_assumption}
    Suppose that the MBPP intensity function is constant over unit interval \( (i - 1, i] \), where \( i \in \mathbb{Z} \). Then \( \MBPcomp(i - 1, i] = \MBPintens(i) \).
\end{lemma}
\begin{proof}
    As the MBPP intensity function is constant over the interval, let \( c \) the value it obtains. Thus,
    \begin{equation*}
        \MBPcomp(i - 1, i] = \int_{i-1}^{i} \MBPintens(s) \, \d s = \int_{i-1}^{i} c \, \d s = c = \MBPintens(i).
    \end{equation*}
\end{proof}
\cref{lem:hip_assumption} shows that under the specific conditions we have specified, the MBPP intensity function can be considered instead of the MBPP compensator --- which can be cumbersome to calculate/approximate for certain kernel functions chosen (as shown in \cref{subsec:npllcalculate}).
If, in addition, we use the SSE loss function, we recover the exact HIP loss function.

\begin{theorem}
    \label{thm:hip_eq_sse}
    Suppose that observation intervals are unit length with constant MBPP intensity. Then the SSE loss function in~\cref{eq:sse_loss} with the MBPP intensity approximated with discrete convolution recovers the HIP loss function~\cite[Eq. (6)]{Rizoiu:2017}, \ie,
    \begin{equation}
        \label{eq:hiploss}
        \frac{1}{2} \sum_{i=1}^{t} (\MBPintens(i) - \HPObserVolume(i-1, i])^{2}.
    \end{equation}
\end{theorem}
\begin{proof}
    Given the observation intervals are unit length and \cref{lem:hip_assumption}, the SSE loss function in~\cref{eq:sse_loss} is given by 
\begin{equation*}
        {\rm SSE}({\bf c}, {\bf \MBPcomp}) = \sum_{i=1}^{t} (\HPObserVolume(i-1, i] - \MBPcomp(i-1, i])^{2} = \sum_{i=1}^{t} (\HPObserVolume(i-1, i] - \MBPintens(i))^{2},
    \end{equation*}
    which when minimised, is equivalent to \cref{eq:hiploss}.
\end{proof}

With the HIP loss function recovered, to recover the HIP model we consider an MBPP with power-law kernel function. Additionally, we also approximate the intensity function by discrete convolution. This approximation step changes the integral in~\cref{eq:xi-definition} to a summation
\begin{equation}
    \label{eq:discrete_approx}
    \immiIntens(i) + \int_{0}^{i} \MBPintens(i) \kernel(t - \tau) \, \d \tau \approx 
    \immiIntens[i] + \sum_{\tau = 1}^{i} \MBPintens[i - \tau] \kernel[\tau],
\end{equation}
where the square brackets indicate that the function is constant over the interval \( (i-1, i] \).

Leaving the choice of kernel aside, the approximation via discrete convolution provides an MBPP intensity function which adheres to the assumption in~\cref{thm:hip_eq_sse}. Thus the entire HIP model can be realised using the loss function framework proposed in this section, generalising~\cref{eqn:discrete-hip}.

\begin{theorem}
    \label{thm:hip}
    Suppose that we have unit length interval-censored event volumes.
    Suppose that we have an MBPP with power-law kernel and that the MBPP intensity is approximated with a discrete convolution,~\cref{eq:discrete_approx}.
    Then by using the SSE loss function,~\cref{eq:sse_loss}, we have the HIP model from~\cite{Rizoiu:2017}.
\end{theorem}
\begin{proof}
    An MBPP with power-law kernel and discrete convolution approximation is equivalent to the MBPP introduced in~\cite[Eq. (4)]{Rizoiu:2017}.

    By definition of the discrete convolution in \cref{eq:discrete_approx}, the intensity function is constant over each interval \( (i - 1, i] \). Thus by~\cref{thm:hip_eq_sse}, the SSE loss function (which uses the MBPP compensator) is equivalent to the HIP loss function~\cite[Eq. (6)]{Rizoiu:2017}.
\end{proof}

\cref{thm:hip} shows that our framework can generalise the method proposed by~\citet{Rizoiu:2017}. In particular, we can consider different kernel functions or use the compensator approximation method in \cref{subsec:npllcalculate} when observation intervals are not uniformly unit length. Importantly, using this framework, we can provide a closed-form solution of HIP when the power-law kernel is replaced with the exponential kernel --- as the discrete convolution and uniform unit length observations are no longer required to approximate the MBPP compensator.

\section{Processes with observed exogenous stimuli}
\label{sec:processes_with_observed_exogenous_stimuli}

In \cref{sec:mbp,sec:bregman-gen,sec:interval-censored point processes}, we used the underlying assumption that the exogenous events of the point process -- i.e., the events not generated through self-excitation -- are solely determined by a corresponding exogenous intensity function \( s(t) \).
Furthermore, we did not differentiate between exogenous and endogenous events.
However, in many real-world scenarios (including our real-world experiments in \cref{sec:real_world_experiments}), the exogenous and endogenous are both separable and directly observed; 
while the underlying functional form of \( s(t) \) is unknown. 
We denote the setting in which the exogenous events can be differentiated as a \emph{separable scenario}. 
Furthermore, in the separable setting, the exogenous and endogenous events can be observed at different granularity levels (event time vs. interval-censored).
For example, consider the problem of predicting flu trends using Twitter data~\citep{Achrekar2011}.
The tweets are observed as event times, while flu cases are recorded as aggregated volumes (for patient privacy reasons~\citep{Rizoiu2016}).

\begin{table}
    \caption{Tabulation of possible data scenarios. ET = ``event times'' and IC = ``interval-censored''. The last column specifies if an endogenous loss function can be used.}
    \centering
    \label{tab:settings}
    \begin{tabular}{cccclcclc}
        \toprule
        \multirow{2}{*}{Scenario} & \multicolumn{2}{c}{Setting} & & \multicolumn{5}{c}{Point Process} \\
\cmidrule{2-3}
        \cmidrule{5-9}
        & Immigrant & Offspring & & Exog. Func. & HP & MBPP & Loss Func. & Endo.? \\
        \midrule
        A & - & ET && Any & \tick & \tick & PP-LL & \cross \\
        B & ET & ET && Multi-Impulse & \tick & \tick & PP-LL & \tick \\
        C & IC & ET && LHPP & \tick & \tick & PP-LL & \tick \\
        D & - & IC && Any & \cross & \tick & IC-LL/SSE & \cross \\
        E & ET & IC && Multi-Impulse & \cross & \tick & IC-LL/SSE & \tick \\
        F & IC & IC && LHPP & \cross & \tick & IC-LL/SSE & \tick \\
        \bottomrule
    \end{tabular}
\end{table} 
\cref{tab:settings} lists the six possible scenarios in which a point process can be observed.
\cref{fig:interval-censored} exemplifies how the same realization is presented in each of these scenarios.
Scenarios A and D are not separable, and events are observed as event times (ET) or interval-censored (IC), respectively (shown in black color).
Scenarios B, C, E, and F are separable and define all the possible combinations of ET and IC for exogenous and endogenous events.
For these separable scenarios, \cref{fig:interval-censored} exemplifies how the exogenous events (upper red events) and endogenous events (lower blue events) can be observed with either one or both types of events as interval-censored.

This section introduces two exogenous functions to approximate $s(t)$, which allow us to characterize the exogenous contribution of a Hawkes process, or MBPP, in the separable scenarios.
We introduce the \textit{multi-impulse exogenous function} and the \textit{latent homogeneous Poisson process exogenous function} which can be used when the exogenous data is observed as event times or interval-censored, respectively. 
We also build endogenous loss functions that account for only endogenous events, allowing us to fit model parameters in all possible regimes of observed data.
\cref{tab:settings} summarises, for each scenario, the corresponding exogenous function, loss function, whether it is applicable for Hawkes, MBPP, or both, and whether an endogenous loss function should be used.

\subsection{Exogenous Event Times -- Multi-Impulse}
\label{subsec:exogenous_event_times}

Consider a separable scenario in which we observe exogenous events as event times (scenarios B and E in \cref{tab:settings}).
That is, we have a set of exogenous event times \( \immiOccurTime_{z} \in \immiOccurTimeSet_{T} \). 
We define the corresponding exogenous intensity as the sum of Dirac delta functions, one for each observed exogenous event, i.e., \( \immiMDelta(t) \) the multi-impulse function.
\begin{definition}
    \label{def:multi-impulse}
    The \emph{multi-impulse} exogenous function for a set of immigrant event times \( \immiOccurTime_{z} \in \immiOccurTimeSet_{T} \) is defined as
    \begin{equation}
        \label{eq:multi-impulse exogenous function}
        \immiMDelta(t) = \sum_{\immiOccurTime_{z} \in \immiOccurTimeSet_{T}} \delta(t - \immiOccurTime_{z}).
    \end{equation}
\end{definition}

Notably, the multi-impulse exogenous function is entirely deterministic for a set of immigrant event times and does not have any parameters to fit. 
Furthermore, \( \immiMDelta(t) \) is not a proper function, as the Dirac delta function is a generalized function not directly computable, \ie, it typically is computed under an integral.
Thus, loss functions which require the multi-impulse exogenous to be directly evaluated are incomputable. 
For these cases, we introduce in~\cref{subsec:end\obsTime_loss_function} the endogenous loss functions.
We use the multi-impulse exogenous function defined in \cref{def:multi-impulse} in \cref{sec:synthetic_experiments} for the scenarios where the immigrants are directly observed, and the intensity function generating them is unknown.

\subsection{Exogenous Interval-Censored -- Latent Homogeneous Poisson Process}
\label{subsec:exogenous_interval_censored}

Here we consider the separable scenarios in which the immigrant events are observed as interval-censored data, with no assumption on the endogenous response of the process (scenarios C and F in \cref{tab:settings}).
Let $\immiObsTimeSet = \{ \immiTime_{0}, \ldots, \immiTime_{m} \}$ be a pre-specified set of observations times for exogenous events.
For each interval \( (\immiTime_{i-1}, \immiTime_{i}] \) we only observe the total immigrant counts \( \immiObserVolume(\immiTime_{i-1}, \immiTime_{i}] \), similarly to \cref{sec:interval-censored point processes}.
Note that for scenario F (with interval-censored endogenous response), the immigrant observation times $\immiTime_i \in \immiObsTimeSet$ can be distinct from the endogenous observation time $ \obsTime_j \in \obsTimeSet$, although they are often identical in practice and in our experiments in \cref{sec:synthetic_experiments,sec:real_world_experiments}.
The rest of this section builds on the more general case where we do not assume that the immigrant observation times \( \immiObsTimeSet \) are the same as the endogenous events \( \obsTimeSet \).

To account for this scenario in which the immigrants are observed as interval-censored data, we assume that the volume of events for each observation interval \( (\immiTime_{i-1}, \immiTime_{i}] \) is determined by a Poisson process. In particular, we define a \emph{latent homogenous Poisson process (LHPP)} with constant intensity \( s(t) = \lambda_{i} \) for each interval; which subsequently defines a piece-wise constant non-homogenous Poisson process as an exogenous function
\begin{equation}
    \label{eq:latent homogeneous poisson}
    \sum_{i=1}^{m} \lambda_{i} \cdot \llbracket \immiTime_{i-1} < t \leq \immiTime_{i} \rrbracket.
\end{equation}

Notably, the rates \( \lambda_{i} \) are free parameters to be chosen. We consider the parameters of maximum likelihood.
\begin{proposition}
Given a latent homogeneous Poisson process defined over observation intervals \( (\immiTime_{i-1}, \immiTime_{i}] \) and corresponding immigrant event counts \( \immiObserVolume(\immiTime_{i-1}, \immiTime_{i}] \), the most likely intensity values are
    \begin{equation}
    \label{eq:latentbestparam}
       \lambda_{i} = \frac{\immiObserVolume(\immiTime_{i-1}, \immiTime_{i}]}{\immiTime_{i} - \immiTime_{i-1}}.
   \end{equation}
\end{proposition}
\begin{proof}
    Suppose we have a homogeneous Poisson point process \( N(t) \) defined over arbitrary half-open interval \( (x, y], \, x \leq y \) with a realization of \( C \) many events occurring on that interval.
The number of events over a given interval in a homogeneous Poisson process follows a Poisson distribution of parameter $\lambda(y-x)$.
    Thus, the most likely intensity value is given by
\begin{align*}
        \lambda^{\star}
        &= \argmax{\lambda}{\mathbb{P}(N(x, y] = C]; \, \lambda)} \\
        &= \argmax{\lambda}{\frac{(\lambda(y-x))^{C}}{C!} \exp(- \lambda (y-x))} \\
        &= \argmax{\lambda}{C \log(\lambda(y-x)) - \log(C!) - \lambda (y-x)} \\
        &= \argmax{\lambda}{C \log(\lambda(y-x)) - \lambda (y-x)}.
    \end{align*}
    The function we are maximizing is concave, thus by setting the derivative to zero we have,
    \begin{equation}
        \lambda^{\star} = \frac{C}{y-x}.
    \end{equation}
    Thus for each of the intervals \( (\immiTime_{i-1}, \immiTime_{i}] \) and corresponding immigrant event counts \( \immiObserVolume(\immiTime_{i-1}, \immiTime_{i}] \) we have \cref{eq:latentbestparam}.
\end{proof}

Thus using the optimal parameters for the latent homogeneous Poisson process, we have the LHPP exogenous function.

\begin{definition}
    Given the interval-censored exogenous volumes \( \immiObserVolume(\immiTime_{i-1}, \immiTime_{i}] \) over observation times \( \{ \immiTime_{0}, \ldots, \immiTime_{m} \} \in \immiObsTimeSet \), then the \emph{latent homogeneous Poisson process (LHPP)} exogenous function is
\begin{equation}
        \label{eq:latentexogfunc}
        \immiIntensStep(t) = \sum_{i=1}^{m} \frac{\immiObserVolume(\immiTime_{i-1}, \immiTime_{i}]}{\immiTime_{i} - \immiTime_{i-1}} \cdot \llbracket \immiTime_{i-1} < t \leq \immiTime_{i} \rrbracket.
    \end{equation}
\end{definition}

If more information is available about the process in which immigrant events are generated, an inductive bias can be added into the point process by changing the distribution of the exogenous function (for example, by considering a non-homogenous Poisson process in each interval). Furthermore, the LHPP exogenous function given in \cref{eq:latentexogfunc} can also be derived by first considering the volume distribution of exogenous events for each interval \( ( \immiTime_j, \immiTime_{j+1}] \) (see \cref{app:lhpp_prob}). In particular, considering a uniform distribution in \cref{app:lhpp_prob} results in the LHPP exogenous function.

\subsection{Endogenous Loss Functions}
\label{subsec:end\obsTime_loss_function}

When the exogenous and endogenous events are separable, both the intensity function and observed data can be separated in to endogenous and exogenous categories. As such, we propose the \emph{endogenous loss function}, which is defined as only the endogenous contribution of the point process to any loss function. The motivation to do so is two fold. Firstly conceptually, as the exogenous events are already observed we only need to account for the endogenous response of the point process. Secondly computationally, the endogenous loss function does not require us to compute the exogenous function when evaluating the loss function. As a side effect, this allows us to circumvent the incomputability problems of the multi-impulse exogenous function, as per~\cref{subsec:exogenous_event_times}.

Suppose that we have a valid point process with corresponding intensity function \( \arbintens(t) \doteq \arbintens(t; \theta) \), where \( \theta \) denotes the parameterisation of the intensity function. Then for a loss function \( \LCal(\theta; \arbintens, \immiOccurTimeSet \cup \offspringOccurTimeSet) \), the intensity function \( \arbintens(\cdot; \theta) \) is compared and evaluated against both the immigrant events \( \immiOccurTimeSet \) and the offspring events \( \offspringOccurTimeSet \), whether event times or interval-censored. Suppose that the data is separable. Then similar to the separability of the data, we can decompose the intensity function into a sum of exogenous and endogenous function components:
\begin{equation}
    \arbintens(t; \param) = \exointens(t; \exoparam) + \edointens(t; \edoparam),
\end{equation}
where \( \exointens(t) \doteq \exointens(t; \exoparam) \) solely characterizes the immigrant events and \( \edointens(t) \doteq \edointens(t; \edoparam) \) solely characterizes the offspring events. The exogenous parameters \( \exoparam \) and endogenous parameters \( \edoparam \) corresponds to decomposition of the parameter \( \param \) as a result of \cref{eq:intens_decomp}.

\bheader{Hawkes process.} An example of a valid decomposition can be found in the definition of the standard Hawkes process intensity function, \cref{eq:hawkes-intensity}. Here, the Hawkes process intensity function is explicitly defined as the addition of an exogenous function \( s(t) \) and endogenous intensity contribution \( \sum_{t_i < t} \kernel( t - t_i ) \) determined by the past events and kernel function \( \kernel(\cdot) \).

\bheader{MBPP.} For MBPP, we can define the exogenous-endogenous function decomposition as
\begin{align}
    \MBPintens(t) &= \exoIntensity(t) + \edoIntensity(t) \\
    \nonumber \\
    \exoIntensity(t) &= \immiIntens(t) \\
    \edoIntensity(t) &= \int_{0}^{t} \MBPintens(y) \kernel(t - y) \d y.
\end{align}
Additionally, if a closed-formed solution of the intensity function \( \MBPintens(t) \) (as per \cref{subsec:solving-mbp-equation}) and exogenous function \( \exoIntensity(t) \) exists, we are able to find a closed-formed solution of the endogenous function
\begin{equation}
    \label{eq:intens_decomp}
    \edoIntensity(t) = \MBPintens(t) - \exoIntensity(t).
\end{equation}

Notably, the endogenous function \( \edointens(t) \) corresponds to the intensity function of a well defined point process. As such, instead of evaluating a loss function on the entire data \( \immiOccurTimeSet \cup \offspringOccurTimeSet \) and the intensity function \( \arbintens(t) \), we can instead evaluate the endogenous intensity function \( 
\edointens(t) \) and its corresponding events; which is exactly the endogenous events \( \offspringOccurTimeSet \) we can identify from the separability of the data. Thus we define the \emph{endogenous loss function} from the original loss function and point process intensity function:
\begin{equation}
    \edoLL(\edoparam) \doteq \LCal(\edoparam; \edointens, \offspringOccurTimeSet).
\end{equation}
In particular, for the rest of this paper, we will assume that \( \exoparam = \emptyset \), as per the exogenous function we have introduced in this section, i.e., the multi-impulse (\cref{eq:multi-impulse exogenous function}) and LHPP (\cref{eq:latentexogfunc}) exogenous functions. Thus, we have that the endogenous parameters account for all model parameters of the original point process, \( \edoparam = \param \).

\bheader{MBPP + PP-LL.} For example, the endogenous loss function of MBPP using the PP-LL loss function, as per \cref{eqn:NHPP}, is defined as
\begin{equation}
    \label{eq:edopointlikelihood}
    \edoLL(\param) =  - \sum_{\offspringOccurTime_{i} \in \offspringOccurTimeSet_{T}} \log \edoIntensity( \offspringOccurTime_i) + \int_{0}^T \edoIntensity( s ) \, \d s,
\end{equation}
where \( \offspringOccurTimeSet_{T} \) is the set of offspring events.
Notably, this endogenous loss function is appropriate when the offspring data is in point format (i.e., scenarios B and C).

\bheader{MBPP + IC-LL.} In the scenarios when the offspring data is in an interval-censored format (i.e., scenarios E and F), we can no longer use \cref{eq:edopointlikelihood}. 
Instead, we can consider the endogenous version of an IC-LL loss function introduced in \cref{sec:interval-censored point processes,sec:bregman-gen}. 
Consider the Bregman divergence interpretation of a loss function, with convex generator \( \varphi(z) \). The endogenous version of the loss function is defined as,
\begin{align} \label{eq:bregman_end\obsTime_example}
    \edoLL(\param) &= \sum_{i = 1}^m \bregdiv_\varphi( \hat{f}_i, \edoMBPcomp(\obsTime_{i-1}, \obsTime_{i}] ), \\
    & \nonumber \\
    \text{where } \edoMBPcomp(t) &= \int^{t}_{0} \edoIntensity(s) \d s \nonumber \\
    \text{and   } \hat{f}_i &= \OffspringVolume(\obsTime_{i-1}, \obsTime_{i}]. \nonumber
\end{align}
Here \( \OffspringVolume(\obsTime_{i-1}, \obsTime_{i}] \) denotes offspring volumes on the interval \( (\obsTime_{i-1}, \obsTime_{i}] \).

\Cref{tab:settings} outlines when using an endogenous loss function is suitable. 
In addition, experiments in \cref{sec:synthetic_experiments} show that even across the different settings where the endogenous loss function is not required, the performances to the non-endogenous standard loss functions are similar.

\section{Synthetic Experiments}
\label{sec:synthetic_experiments}

In this section, we empirically validate our proposals by refitting a synthetically generated dataset with each of the six scenarios described in \cref{tab:settings}.
In \cref{subsec:synthetic-datasets}, we generate datasets of separable Hawkes realizations.
We study in detail two parameter combinations -- one clearly subcritical (branching factor $n^* \ll 1$) and another approaching criticality ($n^* \simeq 1$) -- and two exogenous functions.
We also perform a broad parameter sweep and fit over the parameter grid to analyze fitting stability.
Next, in \cref{subsec:synthetic-setup}, we selectively interval-censor types of events in the datasets to render them compatible with the different scenarios in \cref{tab:settings}.
Last, we present our results in \cref{subsec:synthetic-results}.

\subsection{Synthetic Datasets}
\label{subsec:synthetic-datasets}

We generate synthetic datasets usable in all scenarios (A-F in \cref{tab:settings}).
We start from two exogenous functions and two sets of Hawkes parameters.
We generate separable Hawkes realizations for each of the four combinations -- i.e., immigrants are distinguishable from the offspring.
This allows testing all defined scenarios (for example, to fit in scenario C we hide the exogenous generating function, and we interval-censor the immigrants, more details in \cref{subsec:synthetic-setup}).

\textbf{Exogenous functions.}
We select two functions for $\immiIntens(t)$ the exogenous events intensity.
The first function is a piece-wise constant function, inspired from the latent homogeneous Poisson process defined in \cref{subsec:exogenous_interval_censored}.
We select this function to allow control over the number of exogenous events in each observation interval.
We denote the synthetic datasets generated with this exogenous function as \hprect, and the resulted sampling event intensity is defined as:
\begin{equation}
    \label{eq:hp_rect}
    \HPintens(t) = \sum_{j=1}^n \frac{\immiObserVolume(\immiTime_{j-1}, \immiTime_{j}]}{\immiTime_{j} - \immiTime_{j-1}} \llbracket \immiTime_{j-1} < t \leq \immiTime_{j} \rrbracket + \sum_{\eventOccurTime_i < t} \kappa \theta e^{-\theta (t - \eventOccurTime_i)},
\end{equation}
where \( \immiTime_j \in \immiObsTimeSet_T = \{0, 5, 10, 15\} \), \( n = 2 \); and \( \immiObserVolume(\immiTime_{0}, \immiTime_{1}] = 7 \), \( \immiObserVolume(\immiTime_{1}, \immiTime_{2}] = 6 \), and \( \immiObserVolume(\immiTime_{2}, \immiTime_{3}] = 8 \).

The second immigrant function is a sinusoidal function, which emulates the seasonality often observed in real-world processes.
We denote this dataset as \hpsin, and realizations are sampled from the event intensity:
\begin{equation}
    \label{eq:hp_sine}
    \HPintens(t) = \sin{(t)} + 2 + \sum_{\eventOccurTime_i < t} \kappa \theta e^{-\theta (t - \eventOccurTime_i)}
\end{equation}

\textbf{Hawkes kernel parametrization.}
For each of the above-defined exogenous function, we consider a
Hawkes process with exponential memory kernel \( \kernel(t) = \kappa \theta e^{-\theta t} \). 
We consider two parameter combinations.
The first combination defines a Hawkes process in a clearly subcritical regime (i.e. each event spawn significantly less than one offspring):
\( \kappa = 0.6 \), \( \theta = 0.8 \), resulting in a branching factor $n^* = 0.6$.
The second combination is in a nearly-critical regime, where each event spawns, on average, close to one offspring.
This results in realizations with more significant numbers of events and pronounced clustering effects.
We use the parameter set:
\( \kappa = 0.95 \), \( \theta = 1.15 \), resulting in a branching factor $n^* = 0.95$.

\textbf{Simulation algorithm.}
To be usable with all scenarios in \cref{tab:settings}, we generate sequences in which the exogenous and endogenous event times are distinguishable.
We use a simulation algorithm that leverages the cluster structure of Hawkes process~\citep{hawkes1974cluster,Reinhart2018}, by first sampling the exogenous events and then all their direct and indirect offspring.
The steps for generating the synthetic event sequences are:
\begin{enumerate}
    \item First, we sample immigrants from the fixed exogenous function.
    
    \item Second, we obtain offspring by sampling a Hawkes process with no background intensity and with the immigrant as the first event.
    We use rejection sampling~\citep{Dassios:2013}.
    We denote the collection of an immigrant together with its offspring as a cascade~\citep{Rizoiu:2017}.
    
    \item Finally, we combine all cascades to obtain a sequence, which is a realization of a Hawkes process with the benchmark parameters.
\end{enumerate}

We combine the exogenous and endogenous time points for scenarios where we do not require the separation between immigrants and offspring (i.e., A and D).
For scenarios that require one or both dimensions interval-censored, we compute the corresponding volumes by counting the number of events over pre-specified observation intervals.

\subsection{Experimental Setup}
\label{subsec:synthetic-setup}

\textbf{Pre-process synthetic realizations.}
In our experiments in \cref{subsec:synthetic-results}, the aim is to recover the parameters \( \kappa \) and \( \theta \) of the \emph{true} model used to generate the synthetic datasets. 
We test over all scenarios in \cref{tab:settings}, and for each scenario, we use the exact same set of realizations constructed in \cref{subsec:synthetic-datasets}.
Prior to fitting, for each scenario, we perform the following pre-processing on the realizations:
\begin{itemize}
    \item[\textbf{A.}] We remove the distinction between immigrants and offspring, and we provide the immigrant intensity $\immiIntens(t)$;
    \item[\textbf{B.}] We hide the immigrant intensity $\immiIntens(t)$, and we observe immigrants and offspring separately;
    \item[\textbf{C.}] We interval-censor the immigrants, and we observe offspring as event times;
    \item[\textbf{D.}] We remove the distinction between immigrants and offspring, we interval-censor the resulting event stream, and we provide the immigrant intensity $\immiIntens(t)$;
    \item[\textbf{E.}] We hide the immigrant intensity $\immiIntens(t)$, we interval-censor the offsprings, and we observe immigrants (event times) and offspring (interval-censored) separately;
    \item[\textbf{F.}] We hide the immigrant intensity $\immiIntens(t)$ and we individually interval-censor the immigrant stream and the offspring stream
\end{itemize}

\noindent\textbf{Models and loss functions.}
For the scenarios where the offspring events are observed as event times (scenarios A-C), we fit both Hawkes and MBPP using the PP-LL loss (\cref{eqn:NHPP}).
This allows estimating the performances of MBPP to estimate Hawkes parameters.
For scenarios in which the offspring are interval-censored (D-F), we only fit MBPP using both IC-LL (\cref{eq:nhpp-censored}) and SSE (\cref{eq:sse_loss}) loss functions.
This allows comparing the performances of the two loss functions.
We prefer fitting using the endogenous version of the loss function for the separable settings (B, C, E, F).
However, we compare the standard point process and the endogenous version of the loss function for scenario C.
Finally, for the non-separable scenarios (A and D), we use the non-endogenous loss functions as there is no distinction between immigrants and offspring.

\noindent\textbf{Closed-form vs. approximate loss functions.}
We test the fitting performance gap between the closed-form and approximate versions of our models, using the approximation method specified in \cref{subsec:npllcalculate}. 
Scenario E is used to test our numerically approximated models when a high number of discretization points are given, additionally presenting both endogenous and non-endogenous versions of the IC-LL and SSE loss function.
Additionally, discretization tests which do not involve parameter estimation can be found in~\cref{app:experiments_num_approx} and further non-endogenous loss functions experiments can be found in~\cref{app:experiments_non_endogenous}.

\noindent\textbf{Additional setup.}
For Scenarios A-C, we consider \( 1,000 \) event sequences, which are further split into \( 50 \) groups to estimate parameters. 
For Scenarios D-F, we consider \( 10, 000 \) sequences, also split into \( 50 \) groups to be jointly fit. The additional volume of sequences is to account for the loss of information from interval censoring.
For the scenarios where endogenous events are observed as event time (scenarios A-C), we consider both \hprect and \hpsin datasets. 
For interval-censored endogenous events (scenarios D-F), only the more challenging \hpsin datasets are considered to limit computation time.
In both cases, the sequences we generate are sampled until time-step \(T = 30 \). When we are testing interval-censored settings, we use uniform observation intervals with 5, 10, 15, 30, 60, and 100 observation periods --- for example, with 10 observation intervals we calculate volumes over intervals \( (0, 3], (3, 6], \ldots, (27, 30] \).

In each scenario and model tested we fit \( 50 \) of that model corresponding to the 50 groups of synthetic sequences. We then calculate the average of the estimated parameters and compare the aggregated statistic against the true parameters.

\subsection{Results}
\label{subsec:synthetic-results}

Our investigations in this section have five goals.
First, we test the performance of MBPP for retrieving the Hawkes parameters used to generate the datasets for scenarios A-C (the Hawkes process cannot be used for scenarios D-F).
Second, we seek to quantify the performance loss due to increasingly larger data granularity when moving from scenario A to F.
Third, we investigate the effects of using the endogenous loss introduced in \cref{subsec:end\obsTime_loss_function} for the separable scenarios B, C, E, and F.
Fourth, we study the effects of using the lower bound approximation in \cref{subsec:npllcalculate} for scenarios D, E, F, and the impact of the number of discretization points.
Fifth and last, we study the MBPP fitting stability (the accuracy of retrieving generating parameters) over a wide parameter grid.

\subsubsection{Fitting for scenarios A to F}

\noindent\textbf{Scenario A.}
\cref{tab:experiment_a} shows the results of fitting the Hawkes realizations with both the Hawkes process and MBPP for non-separable data and with knowledge of the parametrization of the exogenous function.
Visibly, the parameters fitted with Hawkes are closest to the true parameters, which is expected as the data is generated from a Hawkes model---i.e., this is the upper bound of how well parameters can be retrieved on each dataset.
Perhaps surprisingly, MBPP has very similar fitting performances, and we only observe a degradation on the \hpsin dataset.
This suggests that MBPP can be used as a drop-in alternative for fitting the parameters of Hawkes processes.

\begin{table}[htbp]
    \parbox{0.48\columnwidth}
    {
    \centering
    \caption{Synthetic experiments for Scenario A: non-separable data, event times.}
    \label{tab:experiment_a}
    \resizebox{0.48\columnwidth}{!}{\begin{tabular}{ccccp{17mm}lc}
        \toprule
        True model && True Params. && \multicolumn{3}{c}{Model Fit} \\
        \cmidrule{1-1}
        \cmidrule{3-3}
        \cmidrule{5-7}
        \multirow{8}{*}{\hprect} && \multirow{4}{*}{\shortstack[1]{\( \kappa = 0.6 \) \\ \( \theta = 0.8 \)}} && \multirow{2}{*}{HP} & \( \kappa \) & \( 0.58 \pm 0.043 \) \\
& & & & & \( \theta \) & \( 0.80 \pm 0.087 \) \\
        \cmidrule{5-7}
        & & & & \multirow{2}{*}{MBPP} & \( \kappa \) & \( 0.58 \pm 0.043 \) \\
& & & & & \( \theta \) & \( 0.78 \pm 0.145 \) \\
        \cmidrule{3-7}
        & & \multirow{4}{*}{\shortstack[1]{\( \kappa = 0.95 \) \\ \( \theta = 1.15 \)}} && \multirow{2}{*}{HP} & \( \kappa \) & \( 0.95 \pm 0.015 \) \\
& & & & & \( \theta \) & \( 1.17 \pm 0.073 \) \\
        \cmidrule{5-7}
        & & & & \multirow{2}{*}{MBPP} & \( \kappa \) & \( 0.95 \pm 0.016 \) \\
& & & & & \( \theta \) & \( 1.21 \pm 0.246 \) \\
        \midrule
        \multirow{8}{*}{\hpsin} && \multirow{4}{*}{\shortstack[1]{\( \kappa = 0.6 \) \\ \( \theta = 0.8 \)}} && \multirow{2}{*}{HP} & \( \kappa \) & \( 0.60 \pm 0.019 \) \\
& & & & & \( \theta \) & \( 0.80 \pm 0.098 \) \\
        \cmidrule{5-7}
        & & & & \multirow{2}{*}{MBPP} & \( \kappa \) & \( 0.60 \pm 0.018 \) \\
& & & & & \( \theta \) & \( 0.75 \pm 0.192 \) \\
        \cmidrule{3-7}
        & & \multirow{4}{*}{\shortstack[1]{\( \kappa = 0.95 \) \\ \( \theta = 1.15 \)}} && \multirow{2}{*}{HP} & \( \kappa \) & \( 0.95 \pm 0.008 \) \\
& & & & & \( \theta \) & \( 1.14 \pm 0.062 \) \\
        \cmidrule{5-7}
        & & & & \multirow{2}{*}{MBPP} & \( \kappa \) & \( 0.95 \pm 0.010 \) \\
& & & & & \( \theta \) & \( 1.22 \pm 0.227 \) \\
        \bottomrule
    \end{tabular}}
    }
\hfill\vline\hfill
\parbox{0.48\columnwidth}
    {
    \centering
    \caption{Synthetic experiments for Scenario B: separable data, exogenous and endogenous event times.}
    \label{tab:experiment_b}
    \resizebox{0.48\columnwidth}{!}{\begin{tabular}{lclcp{17mm}lc}
        \toprule
        True model && True Params. && \multicolumn{3}{c}{Model} \\
        \cmidrule{1-1}
        \cmidrule{3-3}
        \cmidrule{5-7}
        \multirow{8}{*}{\hprect} && \multirow{4}{*}{\shortstack[1]{\( \kappa = 0.6 \) \\ \( \theta = 0.8 \)}} && \multirow{2}{*}{HP-Endo} & \( \kappa \) & \( 0.59 \pm 0.056 \) \\
& & & & & \( \theta \) & \( 0.81 \pm 0.135 \) \\
        \cmidrule{5-7}
        & & & & \multirow{2}{*}{\small MBPP-Endo} & \( \kappa \) & \( 0.59 \pm 0.056 \) \\
& & & & & \( \theta \) & \( 0.8 \pm 0.093 \) \\
        \cmidrule{3-7}
        & & \multirow{4}{*}{\shortstack[1]{\( \kappa = 0.95 \) \\ \( \theta = 1.15 \)}} && \multirow{2}{*}{HP-Endo} & \( \kappa \) & \( 0.94 \pm 0.025 \) \\
& & & & & \( \theta \) & \( 1.17 \pm 0.067 \) \\
        \cmidrule{5-7}
        & & & & \multirow{2}{*}{\small MBPP-Endo} & \( \kappa \) & \( 0.94 \pm 0.025 \) \\
& & & & & \( \theta \) & \( 1.37 \pm 0.228 \) \\
        \hline
        \multirow{8}{*}{\hpsin} && \multirow{4}{*}{\shortstack[1]{\( \kappa = 0.6 \) \\ \( \theta = 0.8 \)}} && \multirow{2}{*}{HP-Endo} & \( \kappa \) & \( 0.59 \pm 0.024 \) \\
& & & & & \( \theta \) & \( 0.74 \pm 0.11 \) \\
        \cmidrule{5-7}
        & & & & \multirow{2}{*}{\small MBPP-Endo} & \( \kappa \) & \( 0.59 \pm 0.024 \) \\
& & & & & \( \theta \) & \( 0.75 \pm 0.098 \) \\
        \cmidrule{3-7}
        & & \multirow{4}{*}{\shortstack[1]{\( \kappa = 0.95 \) \\ \( \theta = 1.15 \)}} && \multirow{2}{*}{HP-Endo} & \( \kappa \) & \( 0.95 \pm 0.021 \) \\
& & & & & \( \theta \) & \( 1.07 \pm 0.28 \) \\
        \cmidrule{5-7}
        & & & & \multirow{2}{*}{\small MBPP-Endo} & \( \kappa \) & \( 0.95 \pm 0.027 \) \\
& & & & & \( \theta \) & \( 1.1 \pm 0.124 \) \\
         \bottomrule
    \end{tabular}}
    }
\end{table}

\noindent\textbf{Scenario B.}
We distinguish between immigrants and offspring in this scenario, but we do not know the immigrant intensity.
We fit the two synthetic datasets using the endogenous versions of MBPP and Hawkes process with the multi-impulse exogenous function. 
\cref{tab:experiment_b} shows the obtained fitted parameters indicating very similar performances for both the Hawkes and MBPP. The 
MBPP even shows a slight advantage for all combinations of parameter and data sets, except for \hprect dataset and parameters \( \kappa = 0.95 \) and \( \theta = 1.15 \).
Furthermore, MBPP even shows a lower standard deviation indicating that it is more stable when predicting parameters.
These results also show the effectiveness of using the multi-impulse exogenous function in the presence of separable data, together with the endogenous loss function.

\noindent\textbf{Scenario C.}
\cref{tab:experiment_c} shows the results of fitting in Scenario C, in which immigrants and offspring are separable, immigrants are interval-censored, and offspring are event times.
Again, we can conclude that the MBPP and the Hawkes process have very similar performances
Similar to Scenario B, the only degradation in approximation quality is for \hprect with the parameters \( \kappa = 0.95 \) and \( \theta = 1.15 \). 
Note that \cref{tab:experiment_c} also presents results for the non-endogenous version of the loss function, which we discuss in \cref{subsubsec:end\obsTime_loss_function_tests}.

\begin{table}[htbp]
    \caption{Synthetic experiments for Scenario C: separable data, interval-censored exogenous events and endogenous event times.}
    \label{tab:experiment_c}
    \parbox{0.49\columnwidth}
    {
    \centering
    \resizebox{0.49\columnwidth}{!}
    {\begin{tabular}{lclcllc}
        \toprule
        True model && True Params. && \multicolumn{3}{c}{Model} \\
        \cmidrule{1-1}
        \cmidrule{3-3}
        \cmidrule{5-7}
        \multirow{16}{*}{\hprect} && \multirow{8}{*}{\shortstack[1]{\( \kappa = 0.6 \) \\ \( \theta = 0.8 \)}} && \multirow{2}{*}{HP} & \( \kappa \) & \( 0.57 \pm 0.064 \) \\
& & & & & \( \theta \) & \( 0.74 \pm 0.097 \) \\
        \cmidrule{5-7}
        & & & & \multirow{2}{*}{HP-Endo} & \( \kappa \) & \( 0.59 \pm 0.056 \) \\
& & & & & \( \theta \) & \( 0.8 \pm 0.093 \) \\
        \cmidrule{5-7}
        & & & & \multirow{2}{*}{MBPP} & \( \kappa \) & \( 0.59 \pm 0.059 \) \\
& & & & & \( \theta \) & \( 0.83 \pm 0.138 \) \\
        \cmidrule{5-7}
        & & & & \multirow{2}{*}{MBPP-Endo} & \( \kappa \) & \( 0.59 \pm 0.056 \) \\
& & & & & \( \theta \) & \( 0.81 \pm 0.139 \) \\
        \cmidrule{3-7}
        & & \multirow{8}{*}{\shortstack[1]{\( \kappa = 0.95 \) \\ \( \theta = 1.15 \)}} && \multirow{2}{*}{HP} & \( \kappa \) & \( 0.94 \pm 0.025 \) \\
& & & & & \( \theta \) & \( 1.12 \pm 0.065 \) \\
        \cmidrule{5-7}
        & & & & \multirow{2}{*}{HP-Endo} & \( \kappa \) & \( 0.94 \pm 0.025 \) \\
& & & & & \( \theta \) & \( 1.17 \pm 0.067 \) \\
        \cmidrule{5-7}
        & & & & \multirow{2}{*}{MBPP} & \( \kappa \) & \( 0.94 \pm 0.025 \) \\
& & & & & \( \theta \) & \( 1.37 \pm 0.237 \) \\
        \cmidrule{5-7}
        & & & & \multirow{2}{*}{MBPP-Endo} & \( \kappa \) & \( 0.94 \pm 0.025 \) \\
& & & & & \( \theta \) & \( 1.37 \pm 0.229 \) \\
        \bottomrule
    \end{tabular}}
    }
\hfill
\parbox{0.49\columnwidth}
    {
    \centering
\resizebox{0.49\columnwidth}{!}
    {\begin{tabular}{lclcllc}
        \toprule
        True model && True Params. && \multicolumn{3}{c}{Model} \\
        \cmidrule{1-1}
        \cmidrule{3-3}
        \cmidrule{5-7}
        \multirow{16}{*}{\hpsin} && \multirow{8}{*}{\shortstack[1]{\( \kappa = 0.6 \) \\ \( \theta = 0.8 \)}} && \multirow{2}{*}{HP} & \( \kappa \) & \( 0.59 \pm 0.025 \) \\
& & & & & \( \theta \) & \( 0.52 \pm 0.08 \) \\
        \cmidrule{5-7}
        & & & & \multirow{2}{*}{HP-Endo} & \( \kappa \) & \( 0.59 \pm 0.024 \) \\
& & & & & \( \theta \) & \( 0.75 \pm 0.098 \) \\
        \cmidrule{5-7}
        & & & & \multirow{2}{*}{MBPP} & \( \kappa \) & \( 0.59 \pm 0.024 \) \\
& & & & & \( \theta \) & \( 0.72 \pm 0.138 \) \\
        \cmidrule{5-7}
        & & & & \multirow{2}{*}{MBPP-Endo} & \( \kappa \) & \( 0.59 \pm 0.023 \) \\
& & & & & \( \theta \) & \( 0.73 \pm 0.107 \) \\
        \cmidrule{3-7}
        & & \multirow{8}{*}{\shortstack[1]{\( \kappa = 0.95 \) \\ \( \theta = 1.15 \)}} && \multirow{2}{*}{HP} & \( \kappa \) & \( 0.95 \pm 0.021 \) \\
& & & & & \( \theta \) & \( 0.99 \pm 0.117 \) \\
        \cmidrule{5-7}
        & & & & \multirow{2}{*}{HP-Endo} & \( \kappa \) & \( 0.95 \pm 0.021 \) \\
& & & & & \( \theta \) & \( 1.1 \pm 0.124 \) \\
        \cmidrule{5-7}
        & & & & \multirow{2}{*}{MBPP} & \( \kappa \) & \( 0.95 \pm 0.03 \) \\
& & & & & \( \theta \) & \( 1.07 \pm 0.296 \) \\
        \cmidrule{5-7}
        & & & & \multirow{2}{*}{MBPP-Endo} & \( \kappa \) & \( 0.95 \pm 0.028 \) \\
& & & & & \( \theta \) & \( 1.07 \pm 0.282 \) \\
        \bottomrule
    \end{tabular}}
    }
\end{table} 
\noindent\textbf{Scenario D.}
In this scenario, we do not distinguish between immigrants and offspring, and we observe the process as interval-censored.
We use for the first time the novel loss functions introduced in \cref{sec:interval-censored point processes}.
I.e., we fit the synthetic data using the IC-LL and SSE loss functions together with the MBPP. 
Note that we cannot use the standard Hawkes process when the events are observed as interval-censored (see \cref{sec:interval-censored point processes}). 
\cref{tab:experiment_d} shows the obtained fitting results for an increasing number of observation intervals.
We fit using both the closed-form solution and the approximation introduced in \cref{subsec:npllcalculate}.
For the approximation, we use as many approximation points as the number of intervals.

We observe that the closed-form MBPP performs exceptionally well, with no noticeable differences between IC-LL and SSE loss. 
In particular, we obtain highly accurate parameter fittings for closed-form models even for small numbers of observations.
However, when we use the numeric approximation of compensators, we observe a decrease in the accuracy of the recovered parameters.
The degradation is prevalent only for \( \theta \), which is consistently overestimated.
Interestingly, an increase in observation intervals does not yield sizable improvements for the approximate models. 
Thus, for this scenario, knowing the functional form of the underlying interval-censored Hawkes process (the exogenous function and kernel function) appears more important than having finer-grained observation periods.

\begin{table}[tbp]
    \centering
    \caption{
        Synthetic experiments for Scenario D: non-separable data, interval-censored events. 
        Note that only MBPP can fit this kind of data.
    }
    \label{tab:experiment_d}
    \resizebox{\columnwidth}{!}{\begin{tabular}{lclcllcccccccccccc}
        \toprule
        \multirow{2}{*}{True Params.} && \multirow{2}{*}{Loss Func.} && \multirow{2}{*}{Model} & \multirow{2}{*}{} && \multicolumn{11}{c}{Number of Observation Intervals} \\
        \cmidrule{8-18}
        && && & && 5 && 10 && 15 && 30 && 60 && 100 \\
        \cmidrule{1-1}
        \cmidrule{3-3}
        \cmidrule{5-6}
        \cmidrule{8-8}
        \cmidrule{10-10}
        \cmidrule{12-12}
        \cmidrule{14-14}
        \cmidrule{16-16}
        \cmidrule{18-18}
        \multirow{8}{*}{\shortstack[1]{\( \kappa = 0.6 \) \\ \( \theta = 0.8 \)}} && \multirow{4}{*}{SSE} && \multirow{2}{*}{Closed} & \( \kappa \) && \(0.6 \pm 0.007\) && \(0.6 \pm 0.007\) && \(0.6 \pm 0.007\) && \(0.6 \pm 0.007\) && \(0.6 \pm 0.006\) && \(0.6 \pm 0.006\) \\
& & && & \( \theta \) && \(0.81 \pm 0.092\) && \(0.81 \pm 0.087\) && \(0.8 \pm 0.07\) && \(0.8 \pm 0.076\) && \(0.8 \pm 0.072\) && \(0.8 \pm 0.075\) \\
        \cmidrule{5-18}
        & & && \multirow{2}{*}{Approx} & \( \kappa \) && \(0.6 \pm 0.007\) && \(0.6 \pm 0.007\) && \(0.6 \pm 0.007\) && \(0.6 \pm 0.007\) && \(0.6 \pm 0.006\) && \(0.6 \pm 0.006\) \\
& & && & \( \theta \) && \(0.85 \pm 0.099\) && \(0.85 \pm 0.096\) && \(0.84 \pm 0.077\) && \(0.84 \pm 0.082\) && \(0.84 \pm 0.079\) && \(0.84 \pm 0.081\) \\
        \cmidrule{3-18}
        & & \multirow{4}{*}{IC-LL} && \multirow{2}{*}{Closed} & \( \kappa \) && \(0.6 \pm 0.007\) && \(0.6 \pm 0.007\) && \(0.6 \pm 0.007\) && \(0.6 \pm 0.007\) && \(0.6 \pm 0.006\) && \(0.6 \pm 0.006\) \\
& & && & \( \theta \) && \(0.82 \pm 0.091\) && \(0.81 \pm 0.086\) && \(0.81 \pm 0.072\) && \(0.8 \pm 0.076\) && \(0.8 \pm 0.073\) && \(0.8 \pm 0.076\) \\
        \cmidrule{5-18}
        & & && \multirow{2}{*}{Approx} & \( \kappa \) && \(0.6 \pm 0.007\) && \(0.6 \pm 0.007\) && \(0.6 \pm 0.007\) && \(0.6 \pm 0.007\) && \(0.6 \pm 0.006\) && \(0.6 \pm 0.006\) \\
& & && & \( \theta \) && \(0.85 \pm 0.098\) && \(0.85 \pm 0.095\) && \(0.84 \pm 0.08\) && \(0.84 \pm 0.083\) && \(0.84 \pm 0.08\) && \(0.84 \pm 0.083\) \\
         \midrule
        \multirow{8}{*}{\shortstack[1]{\( \kappa = 0.95 \) \\ \( \theta = 1.15 \)}} && \multirow{4}{*}{SSE} && \multirow{2}{*}{Closed} & \( \kappa \) && \(0.95 \pm 0.004\) && \(0.95 \pm 0.004\) && \(0.95 \pm 0.004\) && \(0.95 \pm 0.004\) && \(0.95 \pm 0.004\) && \(0.95 \pm 0.005\) \\
& & && & \( \theta \) && \(1.16 \pm 0.082\) && \(1.16 \pm 0.091\) && \(1.16 \pm 0.065\) && \(1.16 \pm 0.1\) && \(1.16 \pm 0.079\) && \(1.16 \pm 0.088\) \\
        \cmidrule{5-18}
        & & && \multirow{2}{*}{Approx} & \( \kappa \) && \(0.95 \pm 0.004\) && \(0.95 \pm 0.004\) && \(0.95 \pm 0.004\) && \(0.95 \pm 0.004\) && \(0.95 \pm 0.004\) && \(0.95 \pm 0.005\) \\
& & && & \( \theta \) && \(1.24 \pm 0.093\) && \(1.24 \pm 0.103\) && \(1.24 \pm 0.074\) && \(1.24 \pm 0.114\) && \(1.24 \pm 0.09\) && \(1.24 \pm 0.1\) \\
        \cmidrule{3-18}
        & & \multirow{4}{*}{IC-LL} && \multirow{2}{*}{Closed} & \( \kappa \) && \(0.95 \pm 0.004\) && \(0.95 \pm 0.004\) && \(0.95 \pm 0.004\) && \(0.95 \pm 0.004\) && \(0.95 \pm 0.004\) && \(0.95 \pm 0.005\) \\
& & && & \( \theta \) && \(1.16 \pm 0.074\) && \(1.16 \pm 0.078\) && \(1.16 \pm 0.06\) && \(1.16 \pm 0.09\) && \(1.16 \pm 0.07\) && \(1.16 \pm 0.082\) \\
        \cmidrule{5-18}
        & & && \multirow{2}{*}{Approx} & \( \kappa \) && \(0.95 \pm 0.004\) && \(0.95 \pm 0.004\) && \(0.95 \pm 0.004\) && \(0.95 \pm 0.004\) && \(0.95 \pm 0.004\) && \(0.95 \pm 0.005\) \\
& & && & \( \theta \) && \(1.24 \pm 0.083\) && \(1.24 \pm 0.089\) && \(1.24 \pm 0.068\) && \(1.24 \pm 0.103\) && \(1.24 \pm 0.079\) && \(1.24 \pm 0.093\) \\
         \bottomrule
    \end{tabular}}
\end{table} 
\noindent\textbf{Scenario E.}
Here, we separately observe immigrants (as event times) and offspring (interval-censored).
We fit using the multi-impulse exogenous function and the SSE and IC-LL loss functions, with both the closed-form solution and the numerical approximation.
\Cref{tab:experiment_e} shows that both loss functions achieve highly accurate parameter fittings for the closed-form models, even for low numbers of observation intervals.
However, the approximate models require higher numbers of observation intervals to recover the parameters. 
For \( \kappa = 0.6 \) and \( \theta = 0.8 \), we require at least \( 30 \) observation intervals to converge close to the correct \( \theta \) value. 
We need at least \( 60 \) intervals for the other parameter setting. 
Notably, the \( \kappa \) parameter is easier to fit and can be approximated using fewer intervals. 
We further explore the dependence of parameter estimation for the numerically approximated models in~\cref{subsubsec:end\obsTime_loss_function_tests}.

\begin{table}[tbp]
    \centering
    \caption{Synthetic experiments for Scenario E: separable data, event time exogenous events and interval-censored endogenous events.}
    \label{tab:experiment_e}
    \resizebox{\columnwidth}{!}{\begin{tabular}{lclcllcccccccccccc}
        \toprule
        \multirow{2}{*}{True Params.} && \multirow{2}{*}{Loss Func.} && \multirow{2}{*}{Model} & \multirow{2}{*}{} && \multicolumn{11}{c}{Number of Observation Intervals} \\
        \cmidrule{8-18}
        && && & && 5 && 10 && 15 && 30 && 60 && 100 \\
        \cmidrule{1-1}
        \cmidrule{3-3}
        \cmidrule{5-6}
        \cmidrule{8-8}
        \cmidrule{10-10}
        \cmidrule{12-12}
        \cmidrule{14-14}
        \cmidrule{16-16}
        \cmidrule{18-18}
        \multirow{8}{*}{\shortstack[1]{\( \kappa = 0.6 \) \\ \( \theta = 0.8 \)}} && \multirow{4}{*}{SSE} && \multirow{2}{*}{Closed} & \( \kappa \) && \( 0.60 \pm 0.006 \) && \( 0.60 \pm 0.006 \) && \( 0.60 \pm 0.005 \) && \( 0.60 \pm 0.006 \) && \( 0.60 \pm 0.005 \) && \( 0.60 \pm 0.005 \) \\
&& && & \( \theta \) && \( 0.79 \pm 0.055 \) && \( 0.79 \pm 0.047 \) && \( 0.79 \pm 0.042 \) && \( 0.79 \pm 0.039 \) && \( 0.79 \pm 0.036 \) && \( 0.79 \pm 0.036 \) \\
        \cmidrule{5-18}
         && && \multirow{2}{*}{Approx} & \( \kappa \) && \( 10 \pm 0 \) && \( 0.62 \pm 0.006 \) && \( 0.61 \pm 0.006 \) && \( 0.60 \pm 0.006 \) && \( 0.60 \pm 0.006 \) && \( 0.60 \pm 0.006 \) \\
         && && & \( \theta \) && \( 0.01 \pm 0 \) && \( 8.82 \pm 0.866 \) && \( 9.55 \pm 1.589 \) && \( 1.09 \pm 0.091 \) && \( 0.91 \pm 0.052 \) && \( 0.86 \pm 0.043 \) \\
         \cmidrule{3-18}
         && \multirow{4}{*}{IC-LL} && \multirow{2}{*}{Closed} & \( \kappa \) && \( 0.60 \pm 0.006 \) && \( 0.60 \pm 0.005 \) && \( 0.60 \pm 0.005 \) && \( 0.60 \pm 0.005 \) && \( 0.60 \pm 0.005 \) && \( 0.60 \pm 0.005 \) \\
         && && & \( \theta \) && \( 0.80 \pm 0.053 \) && \( 0.80 \pm 0.042 \) && \( 0.79 \pm 0.037 \) && \( 0.79 \pm 0.035 \) && \( 0.79 \pm 0.032 \) && \( 0.79 \pm 0.032 \) \\
         && && \multirow{2}{*}{Approx} & \( \kappa \) && \( 10 \pm 0 \) && \( 0.63 \pm 0.006 \) && \( 0.61 \pm 0.006 \) && \( 0.60 \pm 0.005 \) && \( 0.60 \pm 0.005 \) && \( 0.60 \pm 0.005 \) \\
         && && & \( \theta \) && \( 0.01 \pm 0 \) && \( 8.22 \pm 1.192 \) && \( 9.98 \pm 0.103 \) && \( 1.17 \pm 0.089 \) && \( 0.94 \pm 0.045 \) && \( 0.87 \pm 0.038 \) \\
         \midrule
        \multirow{8}{*}{\shortstack[1]{\( \kappa = 0.95 \) \\ \( \theta = 1.15 \)}} && \multirow{4}{*}{SSE} && \multirow{2}{*}{Closed} & \( \kappa \) && \( 0.95 \pm 0.004 \) && \( 0.95 \pm 0.004 \) && \( 0.95 \pm 0.004 \) && \( 0.95 \pm 0.004 \) && \( 0.95 \pm 0.004 \) && \( 0.95 \pm 0.004 \) \\
         && && & \( \theta \) && \( 1.15 \pm 0.076 \) && \( 1.15 \pm 0.075 \) && \( 1.15 \pm 0.073 \) && \( 1.15 \pm 0.073 \) && \( 1.15 \pm 0.072 \) && \( 1.15 \pm 0.072 \) \\
         \cmidrule{5-18}
         && && \multirow{2}{*}{Approx} & \( \kappa \) && \( 10 \pm 0 \) && \( 1.14 \pm 0.007 \) && \( 1.03 \pm 0.005 \) && \( 0.96 \pm 0.003 \) && \( 0.95 \pm 0.004 \) && \( 0.95 \pm 0.004 \) \\
         && && & \( \theta \) && \( 0.01 \pm 0 \) && \( 9.41 \pm 0.764 \) && \( 10 \pm 0 \) && \( 10 \pm 0 \) && \( 1.77 \pm 0.177 \) && \( 1.43 \pm 0.111 \) \\
         \cmidrule{3-18}
         && \multirow{4}{*}{IC-LL} && \multirow{2}{*}{Closed} & \( \kappa \) && \( 0.95 \pm 0.004 \) && \( 0.95 \pm 0.004 \) && \( 0.95 \pm 0.004 \) && \( 0.95 \pm 0.004 \) && \( 0.95 \pm 0.004 \) && \( 0.95 \pm 0.004 \) \\
         && && & \( \theta \) && \( 1.16 \pm 0.068 \) && \( 1.16 \pm 0.067 \) && \( 1.16 \pm 0.065 \) && \( 1.16 \pm 0.065 \) && \( 1.16 \pm 0.065 \) && \( 1.16 \pm 0.065 \) \\
         \cmidrule{5-18}
         && && \multirow{2}{*}{Approx} & \( \kappa \) && \( 10 \pm 0 \) && \( 1.17 \pm 0.007 \) && \( 1.04 \pm 0.005 \) && \( 0.96 \pm 0.003 \) && \( 0.95 \pm 0.004 \) && \( 0.95 \pm 0.004 \) \\
         && && & \( \theta \) && \( 0.02 \pm 0 \) && \( 9.95 \pm 0.032 \) && \( 10 \pm 0 \) && \( 10 \pm 0 \) && \( 1.82 \pm 0.164 \) && \( 1.46 \pm 0.101 \) \\
         \bottomrule
    \end{tabular}}
\end{table} 
\noindent\textbf{Scenario F.}
This is a separable scenario in which both immigrants and offspring are interval-censored.
\cref{tab:experiment_f} shows the fitting results, and we observe that the quality of parameter estimation improves as we increase the number of observation intervals. 
Interestingly, the performance improvement is more constant for the IC-LL than SSE, as the number of intervals grows.
When using the SSE, the parameters are poorly estimated for \( 10 \) and \( 15 \) intervals.
We also note that the numerical approximation requires between \( 60 \) and \( 100 \) intervals to approach the true parameters for this scenario.
Knowing that these results are directly comparable with those for Scenario E (\cref{tab:experiment_e}), we conclude that fitting in Scenario F is the most challenging as most information gets lost when interval-censoring.

\begin{table}[tbp]
    \centering
    \caption{Synthetic experiments for Scenario F: separable data, interval-censored exogenous and endogenous events.}
    \label{tab:experiment_f}
    \resizebox{\columnwidth}{!}{\begin{tabular}{lclcllcccccccccccc}
        \toprule
        \multirow{2}{*}{True Params.} && \multirow{2}{*}{Loss Func.} && \multirow{2}{*}{Model} & \multirow{2}{*}{} && \multicolumn{11}{c}{Number of Observation Intervals} \\
        \cmidrule{8-18}
        && && & && 5 && 10 && 15 && 30 && 60 && 100 \\
        \cmidrule{1-1}
        \cmidrule{3-3}
        \cmidrule{5-6}
        \cmidrule{8-8}
        \cmidrule{10-10}
        \cmidrule{12-12}
        \cmidrule{14-14}
        \cmidrule{16-16}
        \cmidrule{18-18}
        \multirow{8}{*}{\shortstack[1]{\( \kappa = 0.6 \) \\ \( \theta = 0.8 \)}} && \multirow{4}{*}{SSE} && \multirow{2}{*}{Closed} & \( \kappa \) && \( 0.726 \pm 0.004 \) && \( 0.682 \pm 0.005 \) && \( 0.668 \pm 0.005 \) && \( 0.642 \pm 0.005 \) && \( 0.62 \pm 0.005 \) && \( 0.612 \pm 0.005 \) \\
         && && & \( \theta \) && \( 0.647 \pm 0.021 \) && \( 0.641 \pm 0.024 \) && \( 0.801 \pm 0.031 \) && \( 0.995 \pm 0.052 \) && \( 0.915 \pm 0.052 \) && \( 0.865 \pm 0.045 \) \\
        \cmidrule{5-18}
         && && \multirow{2}{*}{Approx} & \( \kappa \) && \( 10 \pm 0 \) && \( 0.621 \pm 0.006 \) && \( 0.605 \pm 0.006 \) && \( 0.601 \pm 0.006 \) && \( 0.601 \pm 0.006 \) && \( 0.601 \pm 0.006 \) \\
         && && & \( \theta \) && \( 0.005 \pm 0 \) && \( 8.823 \pm 0.866 \) && \( 9.546 \pm 1.589 \) && \( 1.09 \pm 0.091 \) && \( 0.909 \pm 0.052 \) && \( 0.856 \pm 0.043 \) \\
         \cmidrule{3-18}
         && \multirow{4}{*}{IC-LL} && \multirow{2}{*}{Closed} & \( \kappa \) && \( 0.729 \pm 0.004 \) && \( 0.69 \pm 0.005 \) && \( 0.672 \pm 0.005 \) && \( 0.642 \pm 0.005 \) && \( 0.621 \pm 0.005 \) && \( 0.613 \pm 0.005 \) \\
         && && & \( \theta \) && \( 0.642 \pm 0.019 \) && \( 0.675 \pm 0.023 \) && \( 0.828 \pm 0.029 \) && \( 0.997 \pm 0.045 \) && \( 0.929 \pm 0.044 \) && \( 0.878 \pm 0.039 \) \\
        \cmidrule{5-18}
         && && \multirow{2}{*}{Approx} & \( \kappa \) && \( 10 \pm 0 \) && \( 0.628 \pm 0.006 \) && \( 0.611 \pm 0.006 \) && \( 0.604 \pm 0.005 \) && \( 0.603 \pm 0.005 \) && \( 0.602 \pm 0.005 \) \\
         && && & \( \theta \) && \( 0.006 \pm 0 \) && \( 8.223 \pm 1.192 \) && \( 9.982 \pm 0.103 \) && \( 1.174 \pm 0.089 \) && \( 0.938 \pm 0.045 \) && \( 0.872 \pm 0.038 \) \\
         \midrule
        \multirow{8}{*}{\shortstack[1]{\( \kappa = 0.95 \) \\ \( \theta = 1.15 \)}} && \multirow{4}{*}{SSE} && \multirow{2}{*}{Closed} & \( \kappa \) && \( 0.958 \pm 0.002 \) && \( 1.496 \pm 2.17 \) && \( 1.494 \pm 2.171 \) && \( 0.95 \pm 0.004 \) && \( 0.95 \pm 0.004 \) && \( 0.95 \pm 0.004 \) \\
         && && & \( \theta \) && \( 1.765 \pm 0.098 \) && \( 1.362 \pm 0.358 \) && \( 1.301 \pm 0.344 \) && \( 1.283 \pm 0.089 \) && \( 1.216 \pm 0.08 \) && \( 1.19 \pm 0.076 \) \\
        \cmidrule{5-18}
         && && \multirow{2}{*}{Approx} & \( \kappa \) && \( 10 \pm 0 \) && \( 1.139 \pm 0.007 \) && \( 1.033 \pm 0.005 \) && \( 0.958 \pm 0.003 \) && \( 0.95 \pm 0.004 \) && \( 0.95 \pm 0.004 \) \\
         && && & \( \theta \) && \( 0.014 \pm 0 \) && \( 9.409 \pm 0.764 \) && \( 10 \pm 0 \) && \( 10 \pm 0 \) && \( 1.765 \pm 0.177 \) && \( 1.433 \pm 0.111 \) \\
         \cmidrule{3-18}
         && \multirow{4}{*}{IC-LL} && \multirow{2}{*}{Closed} & \( \kappa \) && \( 0.959 \pm 0.002 \) && \( 0.953 \pm 0.003 \) && \( 0.95 \pm 0.003 \) && \( 0.949 \pm 0.003 \) && \( 0.95 \pm 0.004 \) && \( 0.95 \pm 0.004 \) \\
         && && & \( \theta \) && \( 1.687 \pm 0.085 \) && \( 1.501 \pm 0.087 \) && \( 1.448 \pm 0.092 \) && \( 1.323 \pm 0.081 \) && \( 1.239 \pm 0.073 \) && \( 1.205 \pm 0.069 \) \\
        \cmidrule{5-18}
         && && \multirow{2}{*}{Approx} & \( \kappa \) && \( 10 \pm 0 \) && \( 1.165 \pm 0.007 \) && \( 1.043 \pm 0.005 \) && \( 0.959 \pm 0.003 \) && \( 0.949 \pm 0.004 \) && \( 0.95 \pm 0.004 \) \\
         && && & \( \theta \) && \( 0.019 \pm 0 \) && \( 9.953 \pm 0.032 \) && \( 10 \pm 0 \) && \( 10 \pm 0 \) && \( 1.823 \pm 0.164 \) && \( 1.456 \pm 0.101 \) \\
         \bottomrule
    \end{tabular}}
\end{table}

\subsubsection{Endogenous loss and number of approximation points}
\label{subsubsec:end\obsTime_loss_function_tests}

\noindent\textbf{Endogenous vs. non-endogenous loss.}
\cref{tab:experiment_c} includes results for both the endogenous and non-endogenous loss function for Scenario C. 
We see that the endogenous versions consistently outperform the non-endogenous version.
This is most prominent in the \hpsin dataset with parameters \( \kappa = 0.6 \) and \( \theta = 0.8 \), where the two Hawkes process models have drastic differences in the fitted \( \theta \) parameter.
\cref{tab:end\obsTime_vs_nonendo} further shows comparisons between endogenous and non-endogenous numeric approximation models for Scenario E, with many approximation points. 
Here we see little or no improvement between endogenous and non-endogenous when using the SSE loss function. 
For the IC-LL loss function, we see that the endogenous version is more stable.
Together with Scenario F's conclusions (that IC-LL is more stable than SSE), we conjecture that using the endogenous IC-LL is preferable.

\noindent\textbf{Number of approximation points.}
All results previously reported in this section use the same number of approximation points for the numerical approximation as the number of observation intervals.
In~\cref{tab:endo_vs_nonendo}, we report on the numeric approximation for Scenario E when \( 300 \) approximation points are used irrespective of the number of observation intervals.
Intuitively, a larger number of approximation points should yield better estimates of the compensator and better parameter fits, with the cost of higher computation time.
We see only a slight improvement for the normal IC-LL loss function as the number of observation intervals increases. 
For all other settings, endogenous or not, the estimated parameters are surprisingly stable. 
This suggests that the performances of the approximation method are dependent on the number of approximation points rather than intervals -- as the number of approximation points increases, the error quickly decreases and converges (see \cref{app:experiments_num_approx}).

\begin{table}[tbp]
    \centering
    \caption{Large number of discretization points experiment. Scenario E: separable data, interval-censored exogenous and endogenous events.}
    \label{tab:endo_vs_nonendo}
    \resizebox{\columnwidth}{!}{\begin{tabular}{lcllcccccccccccc}
        \toprule
        \multirow{2}{*}{Loss Func.} && \multirow{2}{*}{Model} & \multirow{2}{*}{} && \multicolumn{11}{c}{Number of Observation Intervals} \\
        \cmidrule{6-16}
        && & && 5 && 10 && 15 && 30 && 60 && 100 \\
        \cmidrule{1-1}
        \cmidrule{3-4}
        \cmidrule{6-6}
        \cmidrule{8-8}
        \cmidrule{10-10}
        \cmidrule{12-12}
        \cmidrule{14-14}
        \cmidrule{16-16}
        \multirow{4}{*}{SSE} && \multirow{2}{*}{Normal} & \( \kappa \) && \( 0.95 \pm 0.003 \) && \( 0.95 \pm 0.003 \) && \( 0.95 \pm 0.003 \) && \( 0.95 \pm 0.003 \) && \( 0.95 \pm 0.003 \) && \( 0.95 \pm 0.003 \) \\
        && & \( \theta \) && \( 1.23 \pm 0.072 \) && \( 1.23 \pm 0.071 \) && \( 1.23 \pm 0.071 \) && \( 1.23 \pm 0.07 \) && \( 1.23 \pm 0.069 \) && \( 1.23 \pm 0.069 \) \\
        \cmidrule{3-16}
        && \multirow{2}{*}{Endogenous} & \( \kappa \) && \( 0.95 \pm 0.003 \) && \( 0.95 \pm 0.003 \) && \( 0.95 \pm 0.003 \) && \( 0.95 \pm 0.003 \) && \( 0.95 \pm 0.003 \) && \( 0.95 \pm 0.003 \) \\
        && & \( \theta \) && \( 1.23 \pm 0.072 \) && \( 1.23 \pm 0.071 \) && \( 1.23 \pm 0.071 \) && \( 1.23 \pm 0.07 \) && \( 1.23 \pm 0.069 \) && \( 1.23 \pm 0.069 \) \\
        \midrule
        \multirow{4}{*}{IC-LL} && \multirow{2}{*}{Normal} & \( \kappa \) && \( 0.95 \pm 0.003 \) && \( 0.95 \pm 0.003 \) && \( 0.95 \pm 0.003 \) && \( 0.95 \pm 0.003 \) && \( 0.95 \pm 0.003 \) && \( 0.95 \pm 0.003 \) \\
        && & \( \theta \) && \( 0.95 \pm 0.054 \) && \( 0.96 \pm 0.071 \) && \( 0.97 \pm 0.071 \) && \( 0.98 \pm 0.06 \) && \( 1 \pm 0.055 \) && \( 1.03 \pm 0.055 \) \\
        \cmidrule{3-16}
        && \multirow{2}{*}{Endogenous} & \( \kappa \) && \( 0.95 \pm 0.003 \) && \( 0.95 \pm 0.003 \) && \( 0.95 \pm 0.003 \) && \( 0.95 \pm 0.003 \) && \( 0.95 \pm 0.003 \) && \( 0.95 \pm 0.003 \) \\
        && & \( \theta \) && \( 1.24 \pm 0.065 \) && \( 1.24 \pm 0.064 \) && \( 1.24 \pm 0.064 \) && \( 1.24 \pm 0.063 \) && \( 1.24 \pm 0.063 \) && \( 1.24 \pm 0.063 \) \\
        \hline
    \end{tabular}}
\end{table} 

\begin{figure}[htb]
    \centering
\newcommand{\myheight}{0.19}
    \subfloat[]{
        \includegraphics[height=\myheight\textheight]{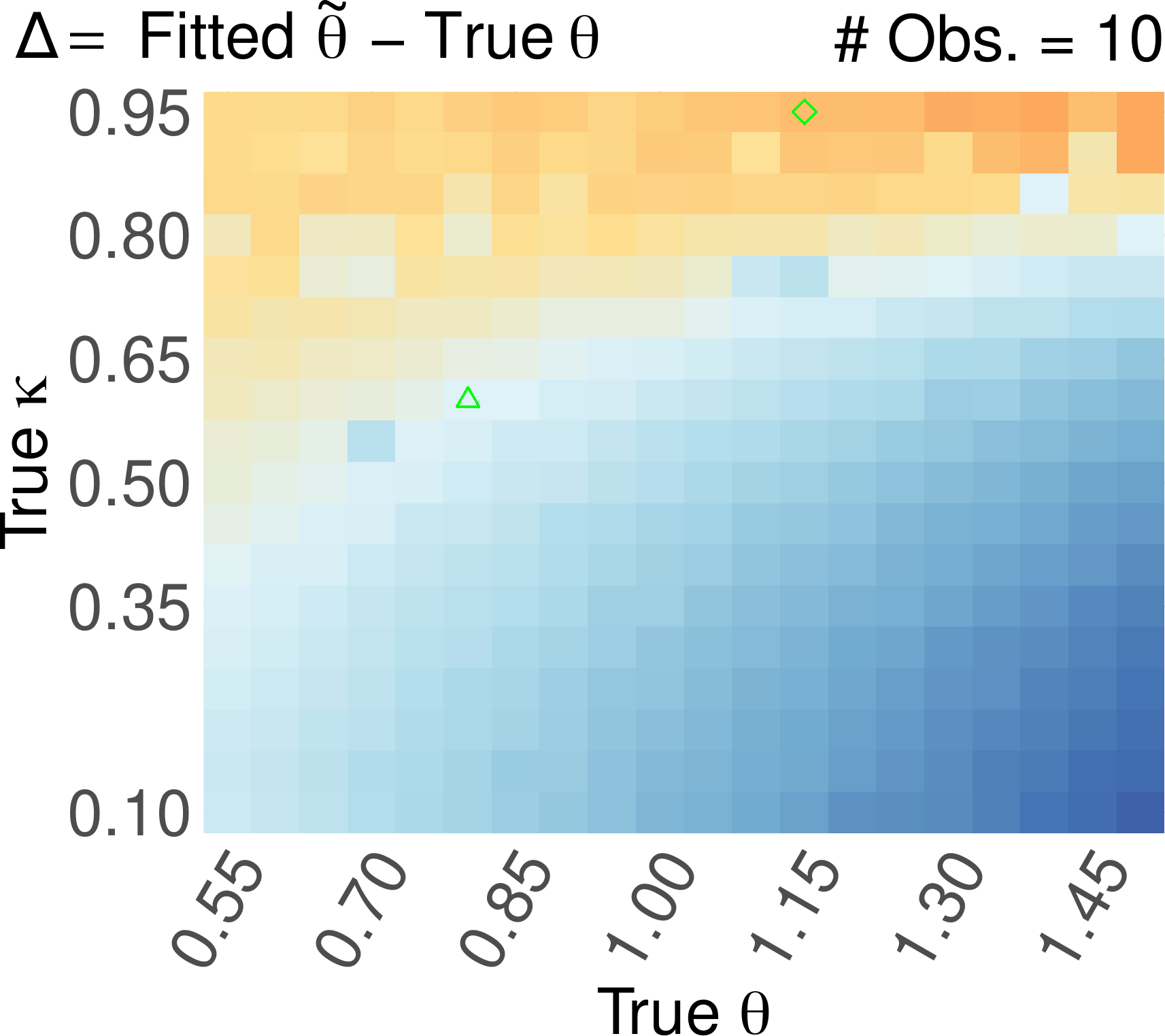}\label{subfig:heatmap_mean_10_obs}}\subfloat[]{
        \includegraphics[height=\myheight\textheight]{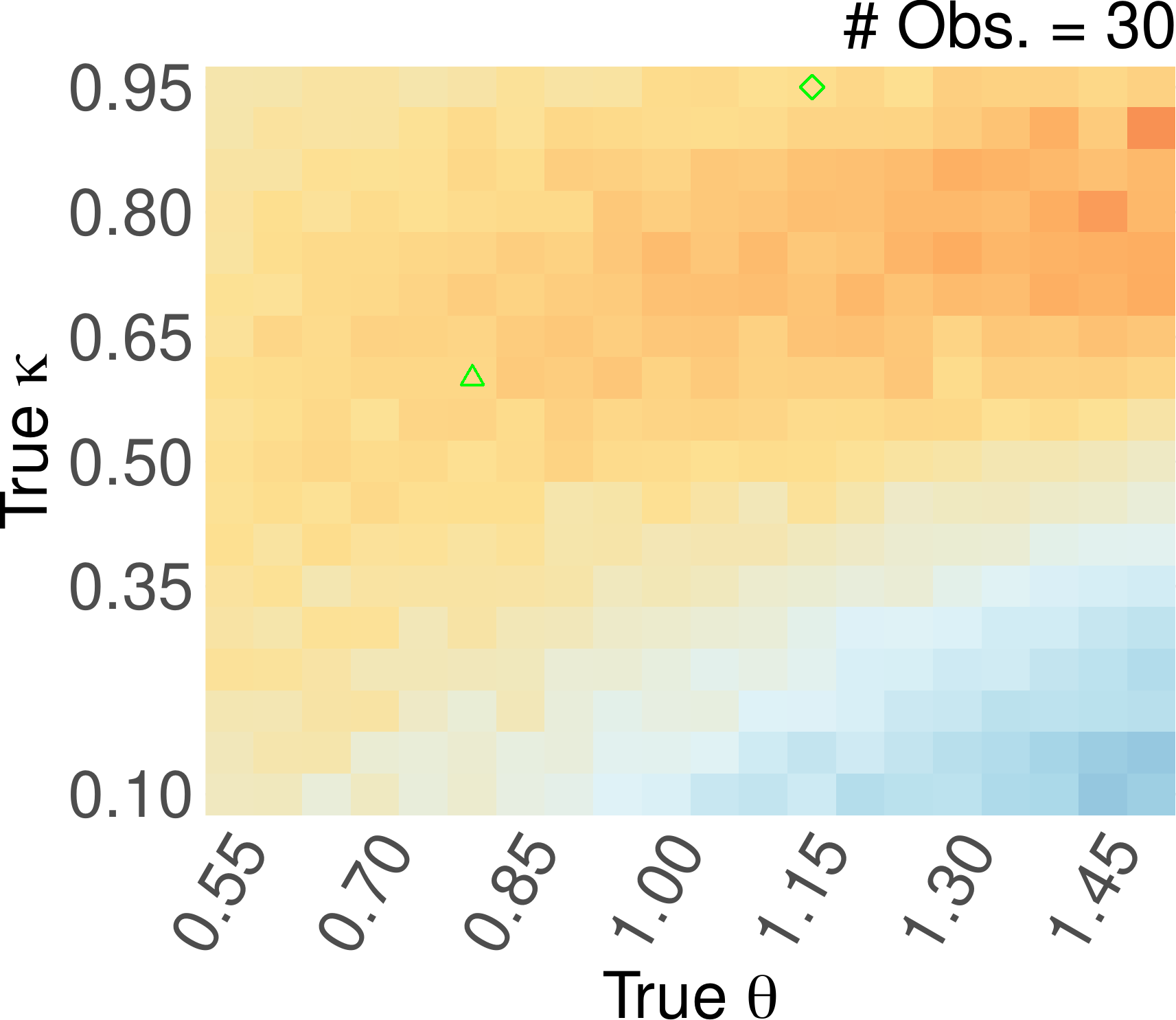}\label{subfig:heatmap_mean_30_obs}}\subfloat[]{
        \includegraphics[height=\myheight\textheight]{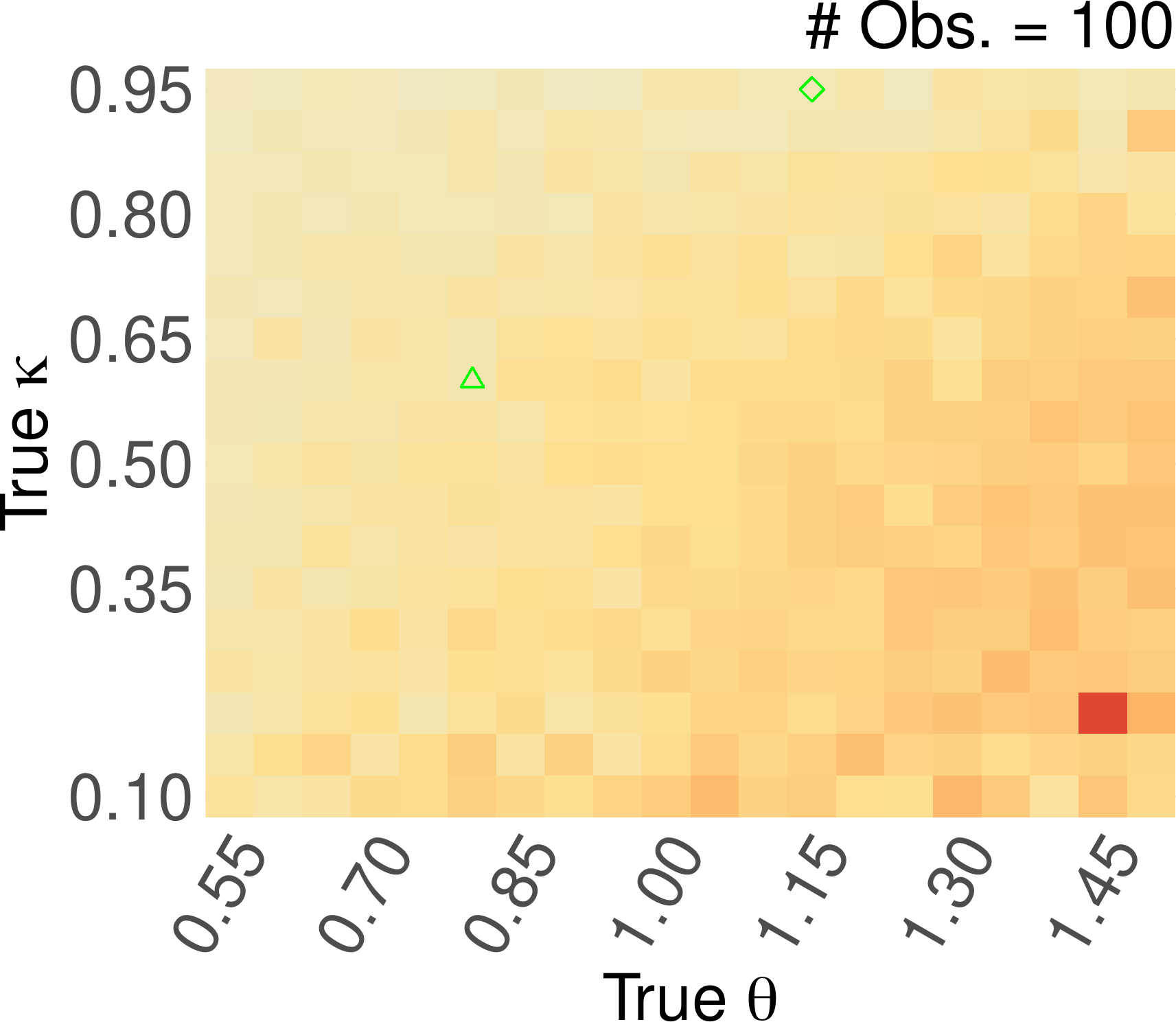}\label{subfig:heatmap_mean_100_obs}}\hspace{1pt}
    {\includegraphics[height=0.23\textheight]{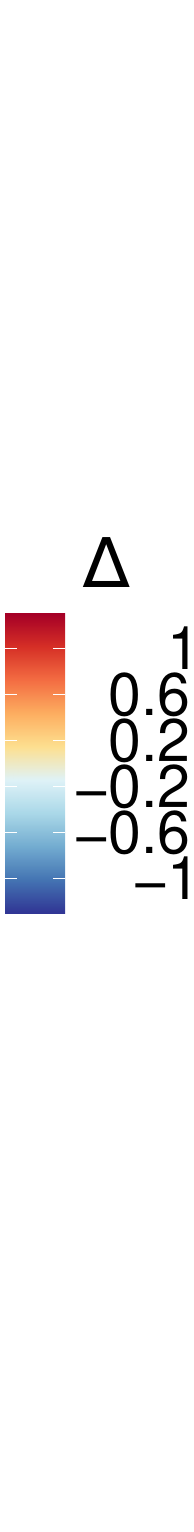}}
    \caption{
        Mean parameter $\theta$ fitting error (\ie, $\text{fitted }\tilde{\theta} - \text{true }\theta$) over a parameter grid $\kappa \in [0.1, 0.95]$ and $\theta \in [0.55, 1.5]$.
        Realizations are sampled using the Hawkes process with the sinusoidal exogenous function and fitted using MBPP in Scenario F with 10 \textbf{(a)}, 30 \textbf{(b)} and 100 \textbf{(c)} observational intervals. 
        The green shapes highlight the specific parameters described in \cref{subsec:synthetic-datasets}: $\{\kappa = 0.6, \theta = 0.8\}$ (green triangle) and $\{\kappa = 0.95, \theta = 1.15\}$ (green diamond).
        For graphics for parameter $\kappa$ and more observational invervals, see \cref{app:mbpp_bias}.
    }
    \label{fig:theta_heatmap_10_30_100_obs}
\end{figure}

\subsubsection{Fitting MBPP with a parameter grid}

The synthetic experiments here above only use two parameter combinations defined in \cref{subsec:synthetic-datasets}.
Here, we verify the stability of the MBPP fitting over a wide parameter grid -- $ \kappa \in [0.1, 0.95]$ (step $0.05$) and $\theta \in [0.5, 1.55]$ (step \( 0.05 \)).
We sample realizations from a Hawkes process with a sinusoidal exogenous function (see \cref{eq:hp_sine}), and we fit MBPP in Scenario F (fully interval-censored).

\cref{fig:theta_heatmap_10_30_100_obs} shows the color map of the fitting error of pameter $\theta$ over the grid.
In every panel, we highlight the two specific parameter combinations used in the synthetic experiments sections above: $\{\kappa = 0.6, \theta = 0.8\}$ (green triangle) and $\{\kappa = 0.95, \theta = 1.15\}$ (green diamond).
Parameter $\kappa$ is consistently retrieved with only minor errors (see \cref{app:mbpp_bias}).
We notice that, for high values of \( \theta \) the fitting under-estimates the true $\theta$ for low values of $\kappa$ (bottom-right corner) and over-estimates it for high values of $\kappa$ (top-right corner).
Note that high $\theta$ corresponds to a faster decay rate for the Hawkes kernel $\phi(t)$, which typically translates to offspring events being tightly clustered close to their parent event.
As $\theta$ decreases (slower kernel decay), the parameter is increasingly well recovered.
As expected, the fitting error decreases as the data granularity (\ie, the number of observation intervals) is increased from 10 (\cref{subfig:heatmap_mean_10_obs}), to 30 (\cref{subfig:heatmap_mean_30_obs}) and 100 (\cref{subfig:heatmap_mean_100_obs}).
For more data granularities, see \cref{app:mbpp_bias}.
Finally, for high data granularity (100 intervals in \cref{subfig:heatmap_mean_100_obs}), we observe a slight over-estimation for $\theta$ due to the model mismatch between the generating model (Hawkes process) and the fitting model (MBPP).
 
\section{Real World Experiments}
\label{sec:real_world_experiments}

We evaluate our proposed MBPP and its variants on \actived~\citep{Rizoiu:2017}, a real-world social media dataset containing views to YouTube videos and the tweets that mention them.
The task is to predict the future popularity of YouTube videos (their future views) by accounting for the exogenous influence of tweets and the endogenous amplification within YouTube (for example, due to word-of-mouth processes).
Both the exogenous (immigrants) and the endogenous (offspring) are in interval-censored format -- counts per day.
Therefore, we fit using Scenario F (also denoted here below as IC-IC) on a subset of days, and we forecast the future.
We compare our MBPP to the HIP process presented in~\cite{Rizoiu:2017}, the current state of the art in predicting popularity under exogenous influence. 
The predicted number of views produced by each model is evaluated using average percentile error (APE) and symmetric mean absolute percentage error (sMAPE).

\subsection{Dataset and prediction setting} 

\noindent\textbf{Dataset.}
We use the \actived~dataset introduced by \citet{Rizoiu:2017} and further studied in \citep{Wu2019,Wu2020}. 
The dataset contains information about more than 14,000 YouTube videos published between 2014-05-29 and 2014-12-26, and the corresponding tweets that share the video in the same time range. 
For each video, we consider the first 120 days of views and tweets. 
We fit model parameters on the first 90 days and we use the later 30 days for evaluation. 
We denote the number of tweets at day \( i \) for a video as \( \left[ \# \texttt{Tweets on day } i \right] \) and the number of YouTube views for a video at day \( i \) as \( \left[ \# \texttt{Views on day } i \right] \).

\noindent\textbf{Prediction setting.}
We replicate the evaluation setup introduced in \citep{Rizoiu:2017}.
To evaluate our models, we predict the number of times a YouTube video is viewed from day 91 to 120. 
We consider that the number of tweets (the exogenous influence) is known in the prediction period.

\noindent\textbf{Compared approaches.}
We evaluate the approximate compensator MBPP, closed-form MBPP, and HIP process, which can be considered an approximation of the MBPP, as per~\cref{subsec:connections_to_ObsTime_hip}. 
We also fit and evaluate these models for both the IC-LL and SSE loss functions; and the exponential and power-law trigger kernel when applicable (the closed-form MBPP only exists for the exponential kernel). 
Furthermore, we use the LHPP exogenous function as the \actived~dataset is in the IC-IC setting. 
It follows that the LHPP exogenous function simplifies to the number of tweets per day as the \actived~dataset presents unit-sized observation intervals (in days). 
Notably, in \citep{Rizoiu:2017}, the HIP process uses only the power-law triggering kernel with the SSE loss function.
Thus, the HIP process with exponential kernel and IC-LL loss function combinations are new to our MBPP framework.

\noindent\textbf{Three fitting procedure augmentations.}
Following \citet{Rizoiu:2017}, we introduce three augmentations to improve fitting and forecasting performances.
The first two augmentations enhance the exogenous intensity function, while the third is a workaround for the local optima in the loss function.

The first two augmentations are required because the models we discuss in this work make the implicit assumption that the exogenous intensity $ \immiIntens(t)$ completely describes all the immigrants that enter the process.
For scenarios B, C, E, F, this implies that the observed exogenous events are the only possible parents for the offspring.
However, this assumption is unrealistic for real data.
For example, the views of YouTube videos can also be driven by other sources of influences outside tweets, such as Facebook or the radio. 
We account for unobserved exogenous influence by augmenting the exogenous intensity $ \immiIntens(t)$.
First, we scale $\immiIntens(t)$ with a fitted constant \( \mu \). 
Second, we add two learnable parameters ($\gamma$ and $\nu$) to account for unobserved external intensity. 
The augmented exogenous intensity for each day $i$ is:
\begin{equation}
    \label{eq:active_exog}
    \hat{s}[i] = \gamma \cdot \llbracket t = 0 \rrbracket + \nu \cdot \llbracket t > 0 \rrbracket + \mu \cdot [\# \texttt{Tweets on day } i],
\end{equation}
where \( \mu \) corresponds to the scaling constant of the observed exogenous influence; and \( \gamma \) and \( \nu \) correspond to parameters accounting for unobserved external influence.

The third augmentation deals with the non-convexity of the loss surface (see \cref{app:convexity analysis} for more details), which can cause the gradient descent minimizer to get stuck into local optima.
We address this by refitting each video $10$ times, starting with randomly selected initial parameters.
We chose as the final parameters the combination with the lowest loss.

\noindent\textbf{Forecasting future views using MBPP.}
We forecast the expected views in the days 91 to 120 using the numeric approximation scheme for the compensators, found in~\cref{prop:numapprox}, and leveraging the observed views in days 1 to 90.
Specifically, the forecasted number of views on day $i$ is:
\begin{align}
    \label{eq:predict}
    &\texttt{Predicted}[i]
    = \hat{s}[90+i]
    + \sum_{j=1}^{90} \left[ \# \texttt{Views on day } j \right] \int_{i-j-1}^{i-j} \phi(y + 90) \d y \nonumber \\
    &\hspace{2cm}+ \sum_{k=91}^{90+i-1} \MBPcomp[k-1, k] \int_{i-k-1}^{i-k} \phi(y + 90) \d y, \enspace \text{for } i \geq 91
\end{align}
where \( \hat{s}[j] \) is defined as per \cref{eq:active_exog}.

We obtain this prediction scheme by replacing the compensator values evaluated between days 1 and 90 with the observed views --- as the expectation of the compensator is equivalent to the expected number of events for a non-homogeneous point process. 
Thus, the fitted compensator only needs to be used when extrapolating past days 90.

\noindent\textbf{Evaluation measures.}
We evaluate the prediction for days 91 to 120 using two evaluation metrics: 
(1) average percentile error (APE) and 
(2) symmetric mean absolute percentage error (sMAPE). 
APE was introduced in~\cite{Rizoiu:2017} and is defined as the absolute difference between the recorded YouTube views percentiles and the predicted YouTube views percentile. 
The measure accounts for the long-tail distribution of online popularity. 
Intuitively, the impact of errors of the same size should be more significant for unpopular videos than for popular videos.
sMAPE is defined as a normalized error of the true YouTube views and the predicted YouTube views over the entire sequence.
Noteworthy, sMAPE is a local error measure (only depends on one video). 
In contrast, APE is a global measure that depends on both the error on the current video and the errors made on all other videos.

\begin{figure}[]
    \centering
	\newcommand\mywidth{0.49} 

	\subfloat[]{
		\includegraphics[width=\mywidth\textwidth]{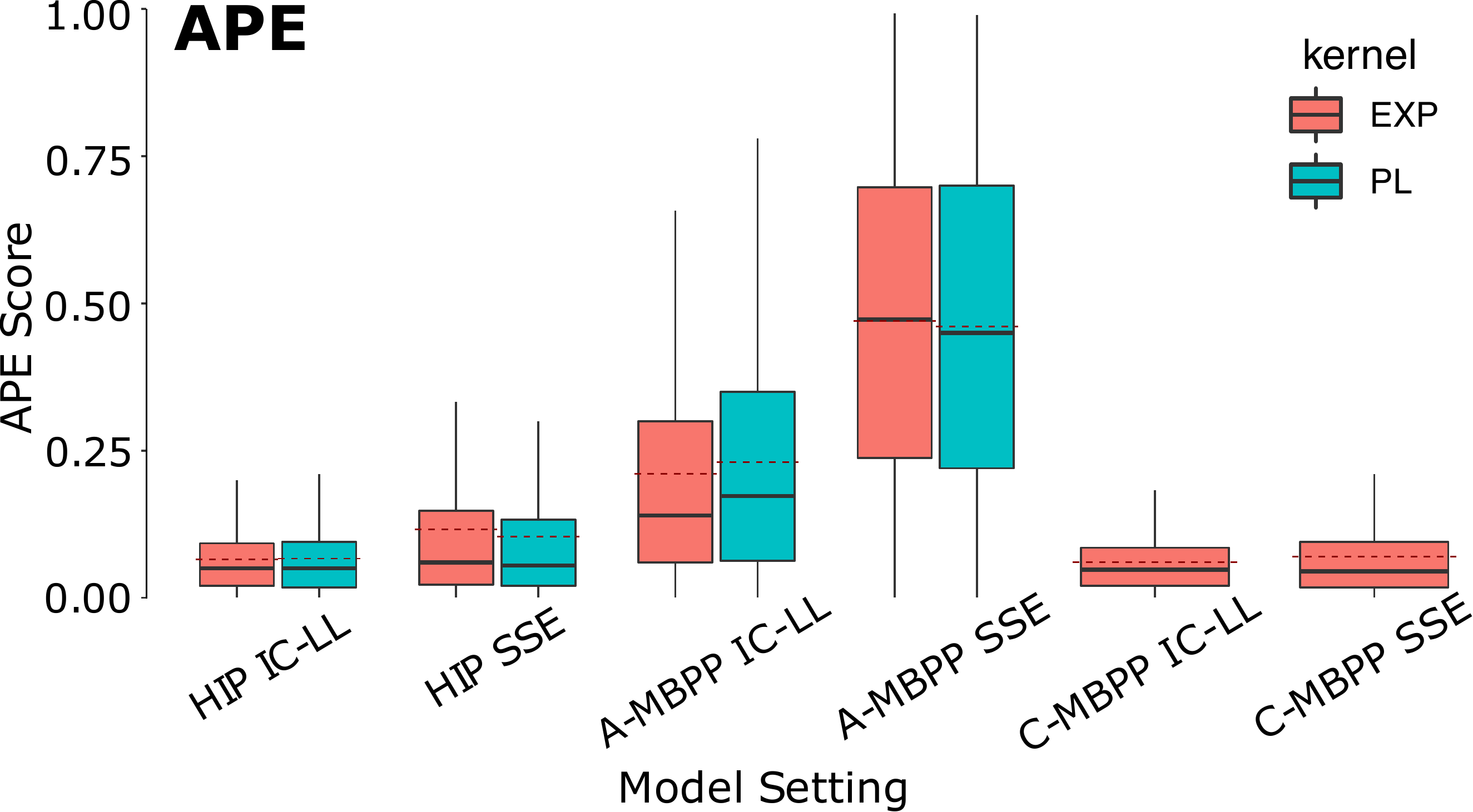}\label{fig:ape_res}}
	\subfloat[]{
		\includegraphics[width=\mywidth\textwidth]{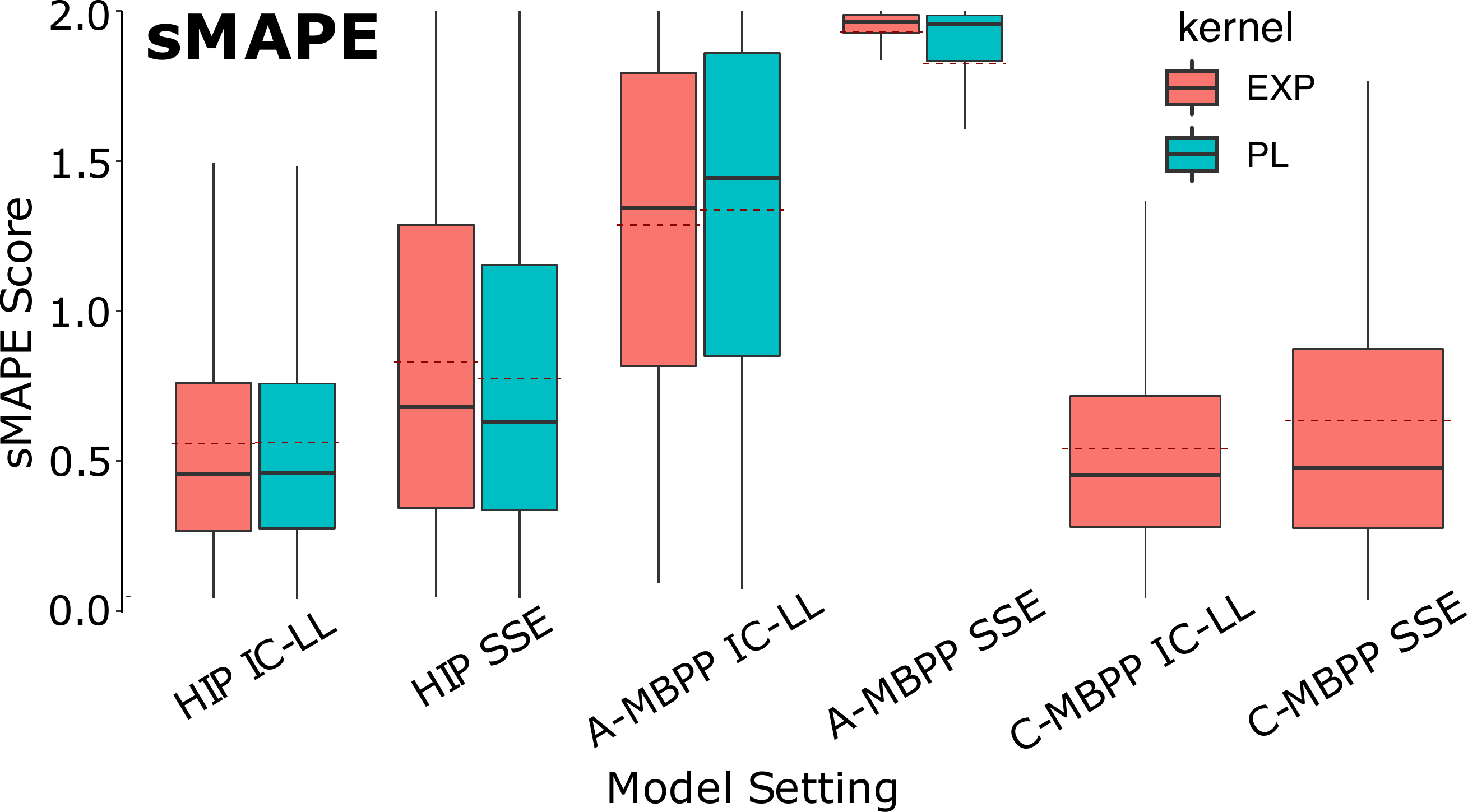}\label{fig:smape_res}}

    \caption{
        \textbf{Evaluation of future popularity prediction using APE (a) and sMAPE (b).}
        For each model, the boxplots aggregate the APE for all videos in the dataset.
        A-MBPP denotes the approximate compensator MBPP variant, and C-MBPP denotes the closed-form compensator MBPP variant. 
        For each boxplot, the solid line denotes the median, and the dashed line denotes the mean.
        The combination HIP SSE PL is the baseline reported in \citep{Rizoiu:2017}
    }
\end{figure}

\subsection{Results}
\label{subsec:xp-results}

We measure the APE (\cref{fig:ape_res}) and sMAPE (\cref{fig:smape_res}) for all HIP and MBPP configurations, summarized as boxplots.
We measure the impact of the choices of three components: 
the kernel (power-law vs. exponential),
the loss functions (IC-LL vs. SSE), and
the model (closed-form MBPP, approximate MBPP, and HIP).
Note that the combination HIP, SSE, and PL kernel is the baseline reported in \citep{Rizoiu:2017}, for which we match the performances claimed by its authors.

We can see that the choice of the kernel does not change the APE value by much, with power-law slightly over-performing exponential (as previously reported in literature~\citep{Mishra2016}).
Visibly, the choice of loss function has a more significant impact on performances, with IC-LL constantly outperforming SSE.
This is particularly visible for the approximate MBPP (A-MBPP) and also present for both HIP and the closed-form MBPP to a lesser extent. 
The result is not surprising as the IC-LL loss function comes from the minimization of the corresponding Poisson distribution defined by the MBPP compensator, whereas the SSE loss function comes from changing the convex generator of a Bregman divergence to obtain an alternative loss function.

The choice of model is also significant, with the closed-form MBPP performing best for each of the loss function choices. 
This was expected, as HIP and the approximate MBPP are both approximations to the closed-form MBPP (for power-law, MBPP not having a closed-form solution). 
Perhaps surprisingly, the HIP has lower APE values than approximate MBPP with the same loss function and triggering kernel.
This is because the HIP approximation is designed explicitly for unit time intervals --- as presented in \cref{subsec:connections_to_ObsTime_hip} --- and would perform better than the more general approximate MBPP.

We draw very similar conclusions from studying sMAPE performance measure (\cref{fig:smape_res}).
The only main difference is that the choice of kernel engenders a larger performance difference;
for example, the 75\% quantile for HIP SSE is much lower for power-law than exponential;
similarly, the approximate MBPP (A-MBPP) SSE with power-law has a lower 25\% quantile than the exponential kernel. 
A-MBPP IC-LL performs slightly worse with a power-law triggering kernel than the exponential triggering kernel, similar to the APE boxplot.

\textbf{Summary.}
Overall, we summarize the three main findings shown by both the local and global error metrics.
First, the IC-LL loss function outperforms SSE.
Second, the closed-form MBPP outperforms numerical approximations (A-MBPP and HIP). 
This follows naturally, given both of those processes are approximations of the closed-form MBPP. 
Third, the HIP approximation~\citep{Rizoiu:2017} -- which is specialized for unit time intervals -- outperforms the general approximation presented in the compensator approximation MBPP.

\section{Related Work}

We structure the related work into three main themes. 
First, we focus on prior work dedicated to estimating the parameters of Hawkes processes in both event-time and interval-censored settings.
Second, we consider alternative theoretic modeling paradigms and augmented Hawkes processes, and we investigate their link to the Hawkes processes.
Finally, we consider the applications of Hawkes processes and other models in various domains.

\subsection{Hawkes model parameter estimation}
\label{subsec:lit-rev-hawkes-estimation}

Several techniques have been proposed for estimating the parameters of a Hawkes model from data.
Here, we concentrate on techniques using event-time data (other than the typical maximum likelihood estimates detailed in \cref{subsec:parameter estimation}) and interval-censored data, both parametric and non-parametric.

\textbf{Event-time models.}
We start with works that estimate Hawkes model parameters starting from event-time data in several settings.
For examples, \citet{Piggott2018} propose an online estimator for the parameters of a Hawkes process by using a Laguerre basis expansion; their work targets an online learning setting.
The RedQueen algorithm~\citep{Zarezade2017} provides a framework for expressing a Hawkes process's counting process and intensity function as a jump stochastic differential equation, providing an online algorithm with provable guarantees~\citep{Zarezade2018}.  Several studies have linked the Hawkes process to the infinite-server queues when the Hawkes process characterizes the arrivals into the queue. 
\citet{Koops2018} obtain a system of differential equations that characterize the joint distribution of arrival intensity and the number of customers. 
Similarly, \citet{Daw2018a} find differential equations to characterize the moments in several different settings. 
Finally, \citet{Gao2018} show that the queue length process is limited to a non-Markovian Gaussian process in the asymptotic regime of a stationary Hawkes process.
More related to the current work, \citet{Xu2017} address the problem of fitting a Hawkes process observed during a finite interval of time -- denoted as a double-censored event sequence. 
Note that their definition of \emph{censored} is different than our interval-censoring in \cref{sect:interval-censored and observation}, i.e., they observe all event times during the given interval and nothing before and after.
To avoid edge effects when the history is missing, \citet{Xu2017} simulate many possible prefixes for the given observed window.

Most of the approaches mentioned above define the Hawkes process using a parametric family of kernels and estimate their parameters.
In contrast, other works estimate the kernel in a non-parametric way.
For example, \citet{Dion2020} use trigonometric basis vectors to provide a non-parametric estimator of one-dimensional diffusion processes characterized by Hawkes processes. 
Similarly, \citet{Li2017} use a Fourier transform to provide a non-parametric estimator for the multidimensional Hawkes process.
\citet{Zhang2019,Zhang2020a,Zhang2020} propose Bayesian estimation procedures for the parameters of the Hawkes process kernel, which leverage the Hawkes process's cluster structure.

The above approaches are typically limited to event time data and cannot estimate model parameters in the presence of interval-censored or mixed data.
Our approach can estimate parameters seamlessly from both event time and interval-censored data and operate on mixtures of the two.

\textbf{Interval-censored models.}
A series of studies show that the Hawkes process can be estimated in the interval-censored setup using an integer-valued auto-regressive (INAR) model.
\citet{Kirchner2016}, and in their follow-up work in \citep{Kirchner2017}, show that the distribution of the resulting `bin-count sequences' (i.e., number of events in each interval) can be approximated by the (multivariate) INAR(p) model.
They establish a least-squares estimation of the INAR(p) model, which yields estimates for the underlying multivariate Hawkes process after appropriate scaling. 
The quality of the estimates is compared against maximum likelihood estimates (on event-times) in \citep{Kirchner2018}.
At first glance, these studies are similar to our work, as the link to the INAR model provides a discrete-time estimator of the Hawkes process.
However, they do not expose the connection between the Hawkes process's event-time and interval-censored formulations as shown in this paper in \cref{sect:interval-censored and observation,subsec:IC-LL-loss-function}.
Therefore, they do not establish a correspondence between the discrete and continuous regimes. 

Additionally, the INAR model and its estimation have been developed under the assumption of a constant exogenous intensity $s(t) = s$. This is incompatible with the synthetic and real-world experiments in \cref{sec:synthetic_experiments} and \cref{sec:real_world_experiments}, respectively, since these experiments involve a non-constant $s(t)$ driving the system.

For other interval-censored settings, \citet{Lu2019} presents several Taylor's expansion approximation algorithms for several thinning-based count processes, including the previously mentioned INAR model.
Similarly, \citet{Manolakis2019} provide a survey of signal processing approaches for interval-censored processes, detailing many different auto-regressive models. 
Like the INAR model, there is no theoretical correspondence between the various discrete models and the underlying true continuous-time process. In the context of survival analysis, \cite{Hudgens2014} propose a scheme for estimating the parameters of the cumulative incidence function in interval-censored and partly interval-censored settings.

Closest to our work, the Hawkes intensity process (HIP)~\citep{Rizoiu:2017} defines an estimator as the Hawkes process intensity function's expectation.
It fits the new process in an interval-censored setting by using a sum of squared errors loss function.
This study has a series of shortcomings already outlined in \cref{subsec:hip}.
In comparison, this work introduces a new point process (the non-homogeneous Mean Behavior Poisson process (MBPP)) whose intensity is the expected intensity of a Hawkes process of equivalent parameters.
The MBPP allows fitting both event-time and interval-censored data.
We further show the specific assumption needed to recover the HIP process (see \cref{subsec:connections_to_ObsTime_hip}).

Among the non-parametric approaches, \citet{Shlomovich2020} propose the estimation of aggregated Hawkes process using a Monte Carlo Expectation-Maximization algorithm.
In aggregated data, event times are rounded to a given precision for data capturing or storage reasons (e.g., only the hour and minute of arrival is recorded, the seconde and below are omitted).
The aggregated process at each time point is the number of events with that rounded time-stamp -- in effect a version of the interval-censored setting defined in \cref{sec:interval-censored point processes,tab:settings}.
A spectral method of approximating stationary Hawkes process, based on Whittle's approach, is established by considering a mixing condition with polynomial decay rates from their Poisson clustering structure~\citep{Cheysson2020}.
There are prior works that propose solutions using panel count data analysis -- an inherently non-parametric approach.
\citet{Ding2018a} propose a Gaussian process-modulated Poisson process that permits panel data observations (i.e., interval-censored).
Later, \citet{Moreno2020} wrap several popular panel count inference methods to deal with incomplete data and the Poisson process assumption's mis-specification.
Note that these approaches estimate Poisson processes, not Hawkes processes, and are not directly linked to the current work apart from the presentation of the data.
Because of the non-parametric estimation of the Poisson intensity, this type of approaches cannot be applied to MBPP either out of the box, as MBPP requires a parametric form to connect back to the Hawkes process as shown in \cref{subsec:MBP-definition}.

Our study differs from the above-mentioned by connecting the Hawkes process with a non-homogeneous Poisson process and providing a generalized treatment of the HIP models.
Notably, our proposed mean behavior Poisson process relates to the Hawkes process by a direct parameter equivalence, which provides a bijection between the models which the previous literature did not have. 
Furthermore, we differentiate from prior studies by providing a singular model to fit both interval-censored and event-time data;
only the log-likelihood function changes to accommodate the type of data. 
Finally, our proposed exogenous functions allow a mixed dataset of interval-censored and event-time data.

\subsection{Other modeling paradigms and their links to the Hawkes model}

The Hawkes processes are not the only option for modeling events in continuous time and various other models have also been used.
Here we present some of these alternatives and how they related to Hawkes.

\textbf{Linking Hawkes to other models.}
In their earlier work, \citet{Lewis1979} propose a foundational connection between homogeneous and non-homogeneous Poisson processes. 
They designed a thinning procedure -- a simple and efficient algorithm for simulating non-homogeneous Poisson processes from homogeneous Poisson processes through thinning. 
This method was also adapted for Hawkes processes and is the de-facto method of simulating Hawkes process event sequences~\citep{ogata1981lewis}.
More recently, the Hawkes process has been linked to epidemic models~\citep{Rizoiu2018}.
More specifically, the expectation of the intensity of the new infection in a stochastic Susceptible-Infectious-Recovery (SIR) epidemic model over all possible recovery processes is equal to the event intensity in a finite population Hawkes process (HawkesN).
HawkesN is a Hawkes process in which the total number of events is limited and for which the remaining number of events modulates the intensity.
\citet{Kong2020} presented a generalized connection between Hawkes processes and epidemic models for arbitrary recovery time distributions and established relationships between the (latent and arbitrary) recovery time distribution, recovery hazard function, and the infection kernel of self-exciting processes.

\textbf{Hawkes properties and augmented models.}
Other studies focus on analyzing other aspects of the Hawkes process that are not directly used to connect different models and estimation procedures. 
Although these studies are only tangentially related to our work, we include some of these theoretical studies for completeness. 
\citet{Cordi2018} show that univariate symmetric Hawkes processes are only weakly causal. They find that reversed event sequences have similar point process log-likelihood scores and sometimes even yield better fits. 
\citet{Zino2018} investigate the epidemic threshold of Susceptible-Infected-Susceptible models in the presence and absence of self-excitation in time-varying networks.
The paper additionally proposes a model where self-excitement govern the temporal evolution of an activity-driven network. 
\citet{Embrechts2018} propose several graph-theoretic objects which summarize the branching structure of a multivariate Hawkes process.
\citet{Daw2018} propose an ephemerally self-excited process which adds an additional stochastic component to the time in which excitement affects the process.
Intuitively, this is motivated by how past events in epidemics only influence the future as long as the event stays active (say, as long as a person is infectious).
Finally, \citet{Dion2019} introduce a new class of diffusion processes where jumps are driven by multivariate nonlinear Hawkes processes.

\textbf{Other point process modeling paradigms.}
Other models have been used in event time modeling as an alternative to Hawkes processes.
Before being used to estimate Hawkes processes (as described above), the integer auto-regressive (INAR) model was proposed as a discrete alternative to the continuous-valued auto-regressive process~\citep{AlOsh1987}. 
Renewal processes are also a common alternative to the Hawkes process.
For example, \citet{Mishra2020} propose a generalized renewal process that arises from the expectation of an age-dependent branching process. 
The renewal process generalization accounts for exogenous events and the time varying-reproduction number for modeling epidemics; 
they tested their approach on COVID-19 case data in South Korea. 
Other point process modeling paradigms make use of Markov processes.
For example, \citet{Soumare2017} propose a two-state Markov switching model to model risk-based pricing models.
Stochastic differential equations are another method of characterizing the dynamics of event time processes, with links to the Hawkes process as per the prior mentioned RedQueen algorithm~\citet{Zarezade2018}. 
A stochastic differential equation framework has been proposed for modeling information diffusion over networks~\citep{Wang2016}. 
Similarly, \citet{Zarezade2017} uses an optimal control problem for jump stochastic differential equations to predict when to post in a social network for visibility.

Our study presents a novel variant of the non-homogeneous Poisson process, the mean behavior Poisson.
Although we mainly use MBPP to approximate the Hawkes process in interval-censored settings, it can be used on its own as an alternative to the Hawkes process with nicer properties (see the independent increments discussion in \cref{sec:prereq}).
Furthermore, the exogenous functions that we introduce can be used for other point process intensity functions.

\subsection{Applications of Hawkes processes}

\textbf{Earthquakes, finance, and epidemics.}
Hawkes processes have long been used for modeling events in various domains, from modeling the aftershocks of earthquakes to predicting social network interactions.
One of the first domains to adapt Hawkes processes was earthquake modeling.
The epidemic-type aftershock (ETAS)~\citep{Ogata1988} model assumes that each earthquake generates aftershocks at a rate governed by a Hawkes model with a power-law decay rate.
Later works show the connection to the Richter intensity scale and provide an analytical solution for Hawkes process variants~\citep{Ogata1999,Helmstetter2002,Helmstetter2002a}.
The Hawkes processes have also been used in financial markets. 
One such work uses a multivariate Hawkes process to account for dynamics of market prices by tailoring four Hawkes kernels for their domain~\citep{Bacry2014}.
Similarly, Hawkes processes have also been used to study epidemic spread. 
\citet{Zino2020} use Hawkes processes to study the bursty activity patterns in social, temporal networks to help determine where targeted behavioral changes could accelerate the erratic of epidemics.

\textbf{Information spread.}
With the rise of social media platforms, many studies use Hawkes processes to characterize how information develops and moves around social networks.
These studies concentrate on either the users of the network or the individual pieces of content (like a post).
Similarly, \citet{Kobayashi2016} predict the time evolution of retweet activity by taking into account the circadian nature of users and the aging of information. 
Hawkes has been effectively applied to social media platforms other than Twitter.
The Hawkes process was found to outperform reinforced Poisson processes for individual micro-blogs on Sina Weibo, a Chinese micro-blogging network. 
\citet{Zarezade2016} propose a spatial-temporal Hawkes process with a periodic decaying kernel, which considers location-based data of users checking in to leverage the observation that users are influenced by locations in which their close friends have recently visited. 
\citet{Wang2019} incorporate the periodic nature of users on social media by switching from a Hawkes process in the day into a non-homogeneous Poisson process at night.

\textbf{Cascade size.}
\citet{Cha2009} analyze the Flickr social network, finding that popular photos do not necessarily spread widely and popular photos spread slowly. 
The study additionally found that information exchanged between friends likely accounted for over 50\% of content sharing, with significant delays between any additional hops.
Similarly, \citet{Goel2012} find that in a variety of social network domains, most of the cascade sizes in network diffusions are small, terminating within one degree of an initial adapting ''seed``.
For large cascades, a study found that in addition to temporal and structural features being key predictors, breadth rather than depth is a better indicator of a large cascade.
\citet{Yu2017} describe a stochastic model based on survival analysis to unify the task of predicting cascade sizes at a given time and the time when a cascade reaches a specific size. 
The authors propose an estimation of the cascading process, and the behavioral dynamics are aggregated.
\citet{Zhao2015} use a doubly stochastic extension of the Hawkes process to examine the Twitter social network and make predictions on a tweet's final number of reshares by looking at its current reshare history. 
\citet{Mishra2016} predict the final popularity of Twitter cascades by combining a marked Hawkes process with a power-law kernel and feature-driven methods.
Finally, \citet{Mishra2018} emulate a Hawkes process unfolding using Recurrent Neural Networks that leverage the Hawkes process intensity function.

\textbf{User influence.}
In a Flickr dataset, \citet{Goyal2010} investigate the influence probabilities between users in a structured social network graph. 
The study presents three classes of models for the influence probabilities: 
the first which assumes that the influence does not change with time;
the second assumes influence is a function of continuous-time;
the third proposes a discrete approximation of the continuous-time models. 
\citet{Matsubara2012} provide an economical alternative to the Hawkes process to model the rise and fall pattern of influence propagation used to investigate the fast propagation of news rumors in social networks.
\citet{Kempe2003} propose a greedy strategy to find the most influential node in a network to determine which nodes can be targeted to trigger a large cascade of idea adaptation.
\citet{Rizoiu2018a} propose a method -- implemented as the software package \texttt{birdspotter}~\citep{Ram2021a} -- to estimate user influence in retweet data sets when the direct retweeting relation (who retweets whom directly) is latent.
They use a Hawkes-inspired retweeting probability and compute user influence as the expected number of retweets over all possible unfolding of retweet cascades.
\citet{Ram2021} further extend this Hawkes process-based method to include social science grounded phenomena, such as network conductance and influence-capital distribution.

\textbf{Content popularity.}
Making predictions about the popularity of online content has been another focus of study in social networks. 
\citet{Szabo2010} present a method for accurately predicting the long-time popularity from early measurements of user access, with empirical results on YouTube and Digg datasets.
Another study found that group-level popularity is a necessary and more practical measurement to predict the popularity of online content~\citep{Hoang2017}.
To predict online video popularity, \citet{Ding2015} propose a dual sentimental Hawkes process that incorporates sentimental features in addition to event counts. 
\cite{Rizoiu:2017} propose the HIP model already discussed in \cref{subsec:hip} and apply it to characterize and predict the popularity of online YouTube videos.
They also use the same model to build optimal promotion schedules to boost online popularity~\citep{Rizoiu2017b}.
\citet{Kong2018} build a dashboard web application that leverages HIP to analyze online videos.

\textbf{Fake users and news.}
There have also been studies on detecting differences in online content, such as differentiating between fake news and regular users. 
\citep{Tschiatschek2018} is one example of such a study, where a Bayesian inference algorithm is used for detecting fake news and jointly learning about the accuracy of users who flag such content over time.
\citet{kong2020exploiting} propose a dual mixture model of Hawkes processes to jointly fit a group of cascades (large and small) relating to content from the same source.
They use the resulted parameters to build diffusion embeddings for news sources -- i.e., a vector that describes a news source based on how the retweet cascades concerning produced articles propagate through Twitter.
The authors show that the diffusion embeddings induce a space in which controversial and reputable news sources are separable.

\textbf{Software tools that implement Hawkes processes.}
There exist several tools that implement the Hawkes process.
These are usually programming language-specific (see \texttt{THAP}~\citep{xu2017thap} in Matlab, \texttt{PoPPy}~\citep{xu2018poppy} in PyTorch) and some allow extensions to the Hawkes process (\texttt{pyhawkes}~\citep{linderman2014discovering} implements network Hawkes models).
The most popular package, \texttt{tick}~\citep{bacry2017tick}, has the most active community and supplies a comprehensive set of models and helper functions for general time-dependent modeling, including Hawkes processes.
Recently, \texttt{evently}~\citep{Kong2021} was developed as a Hawkes process toolbox and designed with an emphasis on online information diffusion modeling.

In this work, we address the application of popularity prediction --- therefore, we classify the present work under the heading content popularity here above.
In our experiments in \cref{subsec:xp-results}, we show that we outperform HIP (the current state of the art in content popularity prediction). 
\section{Conclusion and future work}

This work studies the self-exciting Hawkes point process in an interval-censored setting -- i.e. when we do not observe individual event times but instead counts of events during pre-defined observation intervals.
As Hawkes processes lack the independent increment property, computing the volumes of events is intractable as it depends on all the past observed volumes.
This research overcomes the above difficulties by making several contributions.
First, we establish the Mean Behavior Poisson process (MBPP), a novel Poisson process with a direct parameter correspondence to the popular self-exciting Hawkes process. 
The event intensity function of the MBPP is the expected intensity over all possible Hawkes realizations with the same parameter set.
We fit MBPP in the interval-censored setting using an interval-censored Poisson log-likelihood (IC-LL).
Second, we introduce two novel exogenous functions to distinguish the exogenous from the endogenous events.
We propose the multi-impulse exogenous function when the exogenous events are observed as event time and the latent homogeneous Poisson process exogenous function when the exogenous events are presented as interval-censored volumes.
Third, we provide several approximation methods to estimate the intensity and compensator function of MBPP when no analytical solution exists. 
Fourth and finally, we connect the interval-censored loss of MBPP to a broader class of Bregman divergence-based functions.
Using the connection, we show that the current state of the art in popularity estimation (Hawkes Intensity Process (HIP)~\citep{Rizoiu:2017}) is a particular case of the MBPP. 
We verify our models through empirical testing on synthetic data and real-world data.
We find that on real-world datasets that our MBPP outperforms HIP for the task of popularity prediction.

\textbf{Future work.}
This work only addresses the uni-variate Hawkes process.
However, drawing from the experiments in \cref{sec:real_world_experiments}, one could argue that tweets are not only exogenous sources, and they are influenced by the views also.
In other words, tweets and views should be jointly modeled using a bi-variate Hawkes, where one dimension is interval-censored.
Future work will address this gap by investigating partially censored multivariate MBPP and marked variants of the Hawkes process.
A second direction for future work is finding theoretical bounds for our approximation method for the MBPP compensator (\cref{eq:approx-compensator}).

\acks{This research was partially funded by the Commonwealth of Australia (represented by the Defence Science and Technology Group) through a Defence Science Partnerships Agreement, and Australian Research Council Project DP180101985.
We thanks the
NeCTAR Research Cloud for providing computational resources, an Australian research platform
supported by the National Collaborative Research Infrastructure Strategy.
Furthermore, this research was undertaken with the assistance of resources and services from the National Computational Infrastructure (NCI), which is supported by the Australian Government.
}

\newpage
\appendix

\etocdepthtag.toc{mtappendix}
\etocsettagdepth{mtchapter}{none}
\etocsettagdepth{mtappendix}{subsection}
\etoctocstyle{1}{Contents (Appendix)}
\tableofcontents

\section{Proofs and Derivations} \label{app:proofs}

Here, we present the proofs and derivations that we omitted in the main text for readability reasons.

\begin{proof}[of~\cref{eqn:intensity-renewal} --- which derives $\MBPintens( t )$, the expected intensity of the Hawkes process and the event intensity of MBPP.]
    We shall show that for a Hawkes process,
    \[
       \MBPintens( t ) = \mu( t ) + \int_0^t \phi( s ) \cdot \MBPintens( t - s ) \, \d s.
    \]
    Similar identities also exist~\citep[pg. 8]{Laub:2015}, \citep[Equation 9]{Helmstetter:2002}. 
    Invoking an idea from stochastic calculus (see for instance, Lipster and Shiryaev or Klebaner), let us define the martingale
    \begin{align*}
    \MBPCounts(\cdot) \defEq \HPCounts(\cdot)-\Lambda(\cdot),
    \end{align*}
    where \( \Lambda(\cdot)=\int^t_0\lambda(s)\, \d s \) is a compensator.
    We have:
    \begin{align*}
        \MBPintens( t ) &= \E{}{\lambdaSf( t )} \\
        &= \E{}{ \mu( t ) + \int_0^{t^-} \phi( t - s ) \, \d \HPCounts( s ) } \\
        &= \E{}{ \mu( t ) + \int_0^{t^-} \phi( t - s ) \, [\d \HPCounts( s )-\lambda(s)ds]+\int^{t-}_0\phi(t-s)\lambda(s)\,\d s } \\
        &= \mu( t ) + \E{}{\int_0^{t^-} { \phi( t - s ) \, (\d \HPCounts( s )-\lambda(s)\d s)}+\int^{t-}_0\phi(t-s)\lambda(s)\,\d s } \\
        &= \mu( t ) + \E{}{\int_0^{t^-} { \phi( t - s ) \, (\d \HPCounts( s )-\lambda(s)ds)}+\int^{t-}_0\phi(t-s)\lambda(s)\,\d s } \\
        &= \mu( t ) + \E{}{\int^{t-}_0\phi(t-s)\lambda(s)\,\d s}. \intertext{Since the term is \( \int^t_s\phi(t-s)(dN(s)-\lambda(s))\d s \) is a martingale and has expectation 0, }
        &= \mu( t ) + \int^{t-}_0\phi(t-s)\E{}{\lambda(s)}\d s\\
        &= \mu( t ) + \int^{t-}_0\phi(t-s)\MBPintens(s)\,\d s\\
        &= \mu( t ) + \int^{t-}_0\phi(s)\MBPintens(t-s)\,\d s.
    \end{align*}
\end{proof}

\begin{proof}[of~\cref{thm:laplacesolve} --- which derives the solution for $\MBPintens( t )$ using the Laplace transform.]
    The closed form MBPP equation is given by
    \begin{align*}
        \MBPintens( t ) = \immiIntens( t ) + \int_0^t \kernel( t - \tau ) \cdot \MBPintens( \tau ) \, \d \tau = \immiIntens( t ) + (\kernel * \MBPintens)(t).
    \end{align*}
    Applying the Laplace transform to both sides, we have,
\begin{align*}
        \mathcal{L}\{\MBPintens(t)\}(\omega)
        &= \mathcal{L}\{\immiIntens( t ) + (\kernel * \MBPintens)(t)\}(\omega) \\
        &= \mathcal{L}\{\immiIntens( t )\}(\omega) + \mathcal{L}\{(\kernel * \MBPintens)(t)\}(\omega) \\
        &= \mathcal{L}\{\immiIntens( t )\}(\omega) + \mathcal{L}\{\kernel(t)\}(\omega) \cdot \mathcal{L}\{\MBPintens(t)\}(\omega).
    \end{align*}
    Solving for \( \mathcal{L}\{\MBPintens(t)\}(\omega) \), we have
\begin{equation*}
        \mathcal{L}\{\MBPintens(t)\}(\omega)
        = \frac{\mathcal{L}\{\immiIntens( t )\}(\omega)}{1 - \mathcal{L}\{\kernel(t)\}(\omega)}.
    \end{equation*}
    Thus,
    \begin{equation*}
        \MBPintens(t)
        = \mathcal{L}^{-1}\left\{\frac{\mathcal{L}\{\immiIntens( t )\}(\omega)}{1 - \mathcal{L}\{\kernel(t)\}(\omega)}\right\}(t).
    \end{equation*}
\end{proof}

\begin{proof}[of~\cref{prop:bregman} --- which links the IC-LL loss function and the KL divergence.]
    Let \( \varphi(z) \defEq z \log z \) define a convex function.
    For clarity, further define \( \HPObserVolume_{i} \defEq \HPObserVolume(\obsTime_{i-1}, \obsTime_{i}] \) and \( \MBPcomp_{i} \defEq \MBPcomp(\obsTime_{i-1}, \obsTime_{i}] \).
    We may compute the loss composed of the sum of Bregman divergences with convex generator \( \varphi \):
\begin{align*}
        \LCal(\theta)
        &= \sum_{i=1}^{m} \bregdiv_{\varphi}(\HPObserVolume_{i}, \MBPcomp_{i}) \\
        &= \sum_{i=1}^{m} \left[ \varphi(\HPObserVolume_{i}) - \varphi(\MBPcomp_{i}) - \varphi{\prime}(\MBPcomp_{i}) \cdot (\HPObserVolume_{i} - \MBPcomp_{i}) \right] \\
        &= \sum_{i=1}^{m} \left[ \HPObserVolume_{i} \log \HPObserVolume_{i} - \MBPcomp_{i} \log \MBPcomp_{i} - (\log \MBPcomp_{i} + 1) \cdot (\HPObserVolume_{i} - \MBPcomp_{i}) \right] \\
        &= \sum_{i=1}^{m} \left[ \HPObserVolume_{i} \log \HPObserVolume_{i}  - \HPObserVolume_{i} + \MBPcomp_{i} - \HPObserVolume_{i} \log \MBPcomp_{i} \right] \\
        &= \sum_{i=1}^{m} \MBPcomp_{i} - \sum_{i=1}^{m} \HPObserVolume_{i} \log \MBPcomp_{i} + \sum_{i=1}^{m} \HPObserVolume_{i} (\log \HPObserVolume_{i} - 1).
    \end{align*}
As \( \sum_{i=1}^{m} \HPObserVolume_{i} (\log \HPObserVolume_{i} - 1) \) is just a constant with respect to \( \theta \), the minimisation of \( \LCal(\theta) \) with respect to \( \theta \), is equivalent to
    \[
        \LCal_{IC-LL}(\theta) 
        = \sum_{i=1}^{m} \MBPcomp_{i} - \sum_{i=1}^{m} \HPObserVolume_{i} \log \MBPcomp_{i}.
    \]
\end{proof}

\begin{proof}[of~\cref{eq:sse_loss} --- which shows the equivalence between IC-LL and SSE.]
    
    Let \( \hat\varphi(z) \defEq z^{2} \) define a convex function.
    For clarity, further define \( \HPObserVolume_{i} \defEq \HPObserVolume(\obsTime_{i-1}, \obsTime_{i}] \) and \( \MBPcomp_{i} \defEq \MBPcomp(\obsTime_{i-1}, \obsTime_{i}] \).

    \begin{align*}
        \LCal(\theta)
        = \sum_{i=1}^{m} \bregdiv_{\hat\varphi}(\HPObserVolume_{i}, \MBPcomp_{i})
        = \sum_{i=1}^{m} \left[ \HPObserVolume_{i}^{2} - \MBPcomp_{i}^{2} - 2\MBPcomp_{i} (\HPObserVolume_{i} - \MBPcomp_{i}) \right]
        = \sum_{i=1}^{m} \left[ \HPObserVolume_{i} - \MBPcomp_{i} \right]^{2}
    \end{align*}
    which is exactly the SSE between \( \HPObserVolume_{i} \) and \( \MBPcomp_{i} \).
\end{proof}

\begin{proof}[of~\cref{prop:approx_converge} --- both the lower-bound counting process and upper-bound counting process converge to the original counting process as the discretization points increase.]

To consider the convergence of our counting process approximation, we further assume that the approximation points are equidistant apart
    \[
        \approxpoint_{j}^{\approxupper} = \frac{T}{\approxupper}j.
    \]
    We explicitly add the superscript \( \approxupper \) is to denote the number of approximation points in the discretization sequence. 
    Further we let \( \approxset^{\approxupper} = \left\{ \approxpoint_{j}^{\approxupper} \right\}_{j=0}^{\approxupper} \) as the set of approximation points.

    We consider the convergence of the counting functions in \( L^{1} \)-space, as counting processes are simple functions by definition. In particular, for the lower-bound counting process we aim to show that
    \[
        \lim_{\approxupper \rightarrow \infty} \int_{0}^{T} \vert \MBPCounts(t) - \MBPCounts_{\approxupper}^{-}(t) \vert \, \d t = 0.
    \]
    However, as \( \MBPCounts^{-}_{\approxupper}(t) \) is a lower-bound to \( \MBPCounts(t) \) and by linearity of integration we can instead consider
    \[
        \int_{0}^{T} \vert \MBPCounts(t) - \MBPCounts_{\approxupper}^{-}(t) \vert \d t = \int_{0}^{T} \MBPCounts(t) \d t - \int_{0}^{T} \MBPCounts_{\approxupper}^{-}(t) \d t.
    \]

    For the original counting process \( \MBPCounts(t) \) we assume that we have an arbitrary realisation \( \mathcal{T} = \{ t_{i} \}_{i=1} \) over a time period \( t_{i} \in (0, T] \). Thus we consider the integral of the counting process \( \MBPCounts(t) \) as the following,
\begin{align*}
        \int_{0}^{T} \MBPCounts(t) \d t
        =& \int_{0}^{T} \left[ \sum_{s \in [0,T]} \MBPCounts(\{ s \}) \cdot \llbracket s \leq t \rrbracket \right] \d t \\
        =& \sum_{s \in [0,T]} \MBPCounts(\{ s \}) \cdot \int_{s}^{T} \d t \\
        =& \sum_{s \in [0,T]} \MBPCounts(\{ s \}) \cdot (T - s) \\
        =& \sum_{t_{i} \in \mathcal{T}} \delta(t_{i}) \cdot (T - t_{i}).
    \end{align*}

    For the lower-bound counting process, we first introduce indexing sets \( I_{j}^{\approxupper} = \{ i : t_{i} \in (\approxpoint^{\approxupper}_{j-1}, \approxpoint^{\approxupper}_{j}] \} \) which are the indices of the event times which occur in the half-open interval \( (\approxpoint^{\approxupper}_{j-1}, \approxpoint^{\approxupper}_{j}] \). Thus we have,
\begin{align*}
        \int_{0}^{T} \MBPCounts_{\approxupper}^{-}(t) \d t
        &= \int_{0}^{T} \left[ \sum^{2^{\approxupper}}_{j=1} \MBPCounts(\approxpoint_{j-1}^{\approxupper}, \approxpoint_{j}^{\approxupper}] \cdot \llbracket \approxpoint_{j}^{\approxupper} < t \leq \approxpoint_{\approxupper} \rrbracket \right] \d t \\
        &= \int_{0}^{T} \left[ \sum^{2^{\approxupper}}_{j=1} \MBPCounts( \approxpoint_{j-1}^{\approxupper}, \approxpoint_{j}^{\approxupper}] \cdot \llbracket \approxpoint_{j}^{D} < t \leq T \rrbracket \right] \d t \\
        &= \sum^{2^{\approxupper}}_{j=1} \MBPCounts( \approxpoint_{j-1}^{\approxupper}, \approxpoint_{j}^{\approxupper}] \cdot \int_{\approxpoint_{j}^{\approxupper}}^{T} \d t \\
        &= \sum^{2^{\approxupper}}_{j=1} \MBPCounts( \approxpoint_{j-1}^{\approxupper}, \approxpoint_{j}^{\approxupper}] \cdot ( T - \approxpoint_{j}^{\approxupper} ) \\
        &= \sum^{2^{\approxupper}}_{j=1} \left( \sum_{i \in I_{j}^{\approxupper}} \delta(t_{i}) \right) \cdot ( T - \approxpoint_{j}^{\approxupper} ).
    \end{align*}

    However, we know that \( ( \approxpoint_{j-1}^{\approxupper}, \approxpoint_{j}^{\approxupper}] \) will become a singleton set as \( \approxupper \rightarrow \infty \) from the uniformity of discretization points. Further we know that for any \( \approxupper \), the union \( \bigcup_{j=1}^{2^{\approxupper}} ( \approxpoint_{j-1}^{\approxupper}, \approxpoint_{j}^{\approxupper}] = (0, T] \). Thus in the limit of \( \approxupper \rightarrow \infty \), we have \( \{ ( \approxpoint_{j-1}^{\approxupper}, \approxpoint_{j}^{\approxupper}] : i \in \{ 1, 2, \ldots, 2^{D} \} \} \rightarrow \{ \{ t \} : t \in (0, T] \} \). Thus, we compute the limit as the following,
\begin{align*}
        \lim_{\approxupper \rightarrow \infty} \int_{0}^{T} \MBPCounts_{\approxupper}^{-}(t) \d t
        &= \lim_{\approxupper \rightarrow \infty} \left( \sum^{2^{\approxupper}}_{j=1} \sum_{i \in I_{j}^{\approxupper}} \delta(t_{i}) \cdot \left( T - \approxpoint_{j}^{\approxupper} \right) \right) \\
        &= \sum_{t \in (0,T]} \delta(t) \cdot \llbracket t \in \mathcal{T} \rrbracket \cdot \left( T - t \right) \\
        &= \sum_{t_{i} \in \mathcal{T}} \delta(t_{i}) \cdot \left( T - t_{i} \right) \\
        &= \int_{0}^{T} \MBPCounts(t) \d t.
    \end{align*}
Thus,
    \begin{align*}
        \lim_{\approxupper \rightarrow \infty} \int_{0}^{T} \vert \MBPCounts(t) - \MBPCounts_{\approxupper}^{-}(t) \vert \d t
        &= \lim_{\approxupper \rightarrow \infty} \left( \int_{0}^{T} \MBPCounts(t) \d t - \int_{0}^{T} \MBPCounts_{\approxupper}^{-}(t) \d t \right) \\
        &= \lim_{\approxupper \rightarrow \infty} \int_{0}^{T} \MBPCounts(t) \d t - \lim_{\approxupper \rightarrow \infty} \int_{0}^{T} \MBPCounts_{\approxupper}^{-}(t) \d t \\
        &= \int_{0}^{T} \MBPCounts(t) \d t - \lim_{\approxupper \rightarrow \infty} \int_{0}^{T} \MBPCounts_{\approxupper}^{-}(t) \d t \\
        &= \int_{0}^{T} \MBPCounts(t) \d t - \int_{0}^{T} \MBPCounts(t) \d t \\
        &= 0.
    \end{align*}
The proof is identical when the upper-bound approximation \( \MBPCounts_{\approxupper}^{+}(t) \) is chosen instead.
\end{proof}

\begin{proof}[of~\cref{eq:numapproxupperbound} -- derivation of the upper bound approximation of the MBPP compensator $\MBPcomp(t)$.]
    We consider the following compensator generated by the upper-bound counting process \( \MBPCounts^{+}_{\approxupper}(t) \):
\begin{align*}
        \MBPcomp^{+}(t) 
        &\defEq \immiCounts(t) + \int_{0}^{t} \mathbb{E}[\MBPCounts^{+}_{\approxupper}(y)] \phi(t-y) \d y \\
        &= \immiCounts(t) + \int_{0}^{t} \mathbb{E}\left[\sum_{j=1}^{\approxupper} \MBPCounts(\approxpoint_{j-1}, \approxpoint_{j}] \cdot \llbracket \approxpoint_{j-1} < y \leq \approxpoint_{\approxupper} \rrbracket\right] \phi(t-y) \d y \\
        &= \immiCounts(t) + \sum_{j=1}^{\approxupper} \mathbb{E}\left[\MBPCounts(\approxpoint_{j-1}, \approxpoint_{j}]\right] \int_{0}^{t}  \llbracket \approxpoint_{j-1} < y \leq \approxpoint_{\approxupper} \rrbracket \phi(t-y) \d y \\
        &= \immiCounts(t) + \sum_{j=1}^{\bar{\approxupper}(t)+1} \mathbb{E}\left[\MBPCounts(\approxpoint_{j-1}, \approxpoint_{j}]\right] \int_{\approxpoint_{j-1}}^{\min(t, \approxpoint_{\approxupper})} \phi(t-y) \d y,
    \end{align*}
    where \( \bar{\approxupper}(t) = \max \{ i \in \{ 1, \ldots, \approxupper \} : \approxpoint_{i} < t \} \). This is a similar equation to \cref{eq:numapproxlowerbound}.
\end{proof}

\begin{proof}[of~\cref{prop:numapprox} --- derivation of \( \MBPcomp^{=}_{\approxupper}(\obsTime_{i-1}, \obsTime_{i}] \) the approximation of the MBPP compensator over an interval]
We assume that the anti-derivative of the kernel function \( \kernel = \Phi^{\prime} \) exists. 
    Thus we have
\begin{align*}
        &\MBPcomp^{=}_{\approxupper}(\obsTime_{i-1}, \obsTime_{i}] 
        = \MBPcomp^{-}_{\approxupper}(\obsTime_{i}) - \MBPcomp^{-}_{\approxupper}(\obsTime_{i-1}) \\
        &= \immiCounts(\obsTime_{i}) - \immiCounts(\obsTime_{i-1})
        + \sum_{j=1}^{\bar{\approxupper}(\obsTime_{i})} \mathbb{E}\left[\MBPCounts(\approxpoint_{j-1}, \approxpoint_{j}]\right] \int_{\approxpoint_{j}}^{\min(\obsTime_{i}, \approxpoint_{\approxupper})} \phi(\obsTime_{i}-y) \d y \\
        &\hspace{4cm}- \sum_{j=1}^{\bar{\approxupper}(\obsTime_{i-1})} \mathbb{E}\left[\MBPCounts(\approxpoint_{j-1}, \approxpoint_{j}]\right] \int_{\approxpoint_{j}}^{\min(\obsTime_{i-1}, \approxpoint_{\approxupper})} \phi(\obsTime_{i-1}-y) \d y \\
        &= \immiCounts(\obsTime_{i}) - \immiCounts(\obsTime_{i-1})
        + \sum_{j=1}^{\bar{\approxupper}(\obsTime_{i})} \mathbb{E}\left[\MBPCounts(\approxpoint_{j-1}, \approxpoint_{j}]\right] \int_{\approxpoint_{j}}^{\obsTime_{i}} \phi(\obsTime_{i}-y) \d y \\
        &\hspace{4cm}- \sum_{j=1}^{\bar{\approxupper}(\obsTime_{i-1})} \mathbb{E}\left[\MBPCounts(\approxpoint_{j-1}, \approxpoint_{j}]\right] \int_{\approxpoint_{j}}^{\obsTime_{i-1}} \phi(\obsTime_{i-1}-y) \d y \\
        &= \immiCounts(\obsTime_{i}) - \immiCounts(\obsTime_{i-1})
        + \sum_{j=1}^{\bar{\approxupper}(\obsTime_{i-1})} \mathbb{E}\left[\MBPCounts(\approxpoint_{j-1}, \approxpoint_{j}]\right] \left[ \int_{\approxpoint_{j}}^{\obsTime_{i}} \phi(\obsTime_{i}-y) \d y - \int_{\approxpoint_{j}}^{\obsTime_{i-1}} \phi(\obsTime_{i-1}-y) \d y \right] \\
        &\hspace{4cm}+ \sum_{j=\bar{\approxupper}(\obsTime_{i-1}) + 1}^{\bar{\approxupper}(\obsTime_{i})} \mathbb{E}\left[\MBPCounts(\approxpoint_{j-1}, \approxpoint_{j}]\right] \int_{\approxpoint_{j}}^{\obsTime_{i}} \phi(\obsTime_{i}-y) \d y \\
        &= \immiCounts(\obsTime_{i}) - \immiCounts(\obsTime_{i-1}) 
        + \sum_{j=1}^{\bar{\approxupper}(\obsTime_{i-1})} \mathbb{E}\left[\MBPCounts(\approxpoint_{j-1}, \approxpoint_{j}]\right]\left[ -(\Phi(0) - \Phi(\obsTime_{i} - \approxpoint_{j})) + (\Phi(0) + \Phi(\obsTime_{i-1} - \approxpoint_{j})) \right] \\
        &\hspace{4cm}+ \sum_{j=\bar{\approxupper}(\obsTime_{i-1}) + 1}^{\bar{\approxupper}(\obsTime_{i})} \mathbb{E}\left[\MBPCounts(\approxpoint_{j-1}, \approxpoint_{j}]\right] \left[ -\Phi(0) + \Phi(\obsTime_{i} - \approxpoint_{j}) \right] \\
        &= \immiCounts(\obsTime_{i}) - \immiCounts(\obsTime_{i-1}) 
        + \sum_{j=1}^{\bar{\approxupper}(\obsTime_{i-1})} \mathbb{E}\left[\MBPCounts(\approxpoint_{j-1}, \approxpoint_{j}]\right]\left[ \Phi(\obsTime_{i} - \approxpoint_{j}) - \Phi(\obsTime_{i-1} - \approxpoint_{j}) \right] \\
        &\hspace{4cm}+ \sum_{j=\bar{\approxupper}(\obsTime_{i-1}) + 1}^{\bar{\approxupper}(\obsTime_{i})} \mathbb{E}\left[\MBPCounts(\approxpoint_{j-1}, \approxpoint_{j}]\right] \left[ \Phi(\obsTime_{i} - \approxpoint_{j}) - \Phi(0) \right] \\
        &= \immiCounts(\obsTime_{i}) - \immiCounts(\obsTime_{i-1}) + \sum_{j=1}^{\bar{\approxupper}(\obsTime_{i-1})} \mathbb{E}\left[\MBPCounts(\approxpoint_{j-1}, \approxpoint_{j}]\right] \int_{\obsTime_{i-1} - \approxpoint_{j}}^{\obsTime_{i} - \approxpoint_{j}} \phi(y) \d y \\
        &\hspace{4cm}+ \sum_{j=\bar{\approxupper}(\obsTime_{i-1}) + 1}^{\bar{\approxupper}(\obsTime_{i})} \mathbb{E}\left[\MBPCounts(\approxpoint_{j-1}, \approxpoint_{j}]\right] \int^{\obsTime_{i} - \approxpoint_{j}}_{0} \phi(y) \d y \\
        &= \immiCounts(\obsTime_{i}) - \immiCounts(\obsTime_{i-1}) + \sum_{j=1}^{\bar{\approxupper}(\obsTime_{i-1})} \mathbb{E}\left[\MBPCounts(\approxpoint_{j-1}, \approxpoint_{j}]\right] \int_{\obsTime_{i-1} - \approxpoint_{j}}^{\obsTime_{i} - \approxpoint_{j}} \phi(y) \d y \\
        &\hspace{4cm} + \sum_{j=\bar{\approxupper}(\obsTime_{i-1}) + 1}^{\bar{\approxupper}(\obsTime_{i})} \mathbb{E}\left[\MBPCounts(\approxpoint_{j-1}, \approxpoint_{j}]\right] \int_{\obsTime_{i-1} - \approxpoint_{j}}^{\obsTime_{i} - \approxpoint_{j}} \phi(y) \d y \\
        &\hspace{4cm}+ \sum_{j=\bar{\approxupper}(\obsTime_{i-1}) + 1}^{\bar{\approxupper}(\obsTime_{i})} \mathbb{E}\left[\MBPCounts(\approxpoint_{j-1}, \approxpoint_{j}]\right] \int^{\obsTime_{i-1} - \approxpoint_{j}}_{0} \phi(y) \d y \\
        &= \immiCounts(\obsTime_{i}) - \immiCounts(\obsTime_{i-1}) + \sum_{j=1}^{\bar{\approxupper}(\obsTime_{i})} \mathbb{E}\left[\MBPCounts(\approxpoint_{j-1}, \approxpoint_{j}]\right] \int_{\obsTime_{i-1} - \approxpoint_{j}}^{\obsTime_{i} - \approxpoint_{j}} \phi(y) \d y \\
        &\hspace{4cm}+ \sum_{j=\bar{\approxupper}(\obsTime_{i-1}) + 1}^{\bar{\approxupper}(\obsTime_{i})} \mathbb{E}\left[\MBPCounts(\approxpoint_{j-1}, \approxpoint_{j}]\right] \int^{\obsTime_{i-1} - \approxpoint_{j}}_{0} \phi(y) \d y.
    \end{align*}
\end{proof}
 
\begin{landscape}
    \begin{table*}[htbp]
        \centering
        \caption{
            Solutions for different immigrant intensities \( \immiIntens(t) \), and the exponentially time-decaying kernel in \cref{eq:exp-kernel} of the main text. \( \immiOccurTimeSet_{T} \) acts as a set of immigrant event times.
        }\label{tab:appendix_mbp_intensity}
        \begin{tabular}{c|l|p{14cm}} 
            \toprule
            & \( \immiIntens(t) \) & MBPP Intensity Function \( \MBPintens(t) \) \\
            \midrule
            I & \( \delta(t - a) \) & \( \delta(t - a) + \kappa \theta \exp( (k-1) \theta (t - a) ) \cdot \llbracket t > a \rrbracket \) \\[.3cm]
            \midrule
            II & \( \sum_{\immiOccurTime_{z} \in \immiOccurTimeSet_{T}} \delta(t - \immiOccurTime_{z}) \) & \( \sum_{\immiOccurTime_{z} \in \immiOccurTimeSet_{T}} \bigg[ \delta(t - \immiOccurTime_{z}) + \kappa \theta \exp( (k-1) \theta (t - \immiOccurTime_{z}) ) \cdot \llbracket t > \immiOccurTime_{z} \rrbracket \bigg] \) \\[.3cm]
            \midrule
            III & \( \llbracket a \leq t < b \rrbracket \) & \( \frac{\kappa}{1 - \kappa} \bigl[ \frac{1}{k} - \exp( (k - 1) \theta (t - a) ) \bigr] \cdot \llbracket a < t \leq b \rrbracket + \frac{\kappa}{1 - \kappa} \bigl[ \exp( (k - 1) \theta (t - b) ) - \exp( (k - 1) \theta (t - a) ) \bigr] \cdot \llbracket t > b \rrbracket \) \\[.3cm]
            \midrule
            IV & \( \sum_{i=1}^{m} \frac{\immiObserVolume(\immiTime_{i-1}, \immiTime_{i}]}{\immiTime_{i} - \immiTime_{i-1}} \cdot \llbracket \immiTime_{i-1} < t \leq \immiTime_{i} \rrbracket \) & \( \sum_{i=1}^{m} \frac{\immiObserVolume(\immiTime_{i-1}, \immiTime_{i}]}{\immiTime_{i} - \immiTime_{i-1}} \Bigl[ \frac{\kappa}{1 - \kappa} \bigl[ \frac{1}{k} - \exp( (k - 1) \theta (t - \immiTime_{i-1}) ) \bigr] \cdot \llbracket \immiTime_{i-1} < t \leq \immiTime_{i} \rrbracket + \frac{\kappa}{1 - \kappa} \bigl[ \exp( (k - 1) \theta (t - \immiTime_{i}) ) - \exp( (k - 1) \theta (t - \immiTime_{i-1}) ) \bigr] \cdot \llbracket t > \immiTime_{i} \rrbracket \Bigr] \) \\[.3cm]
            \midrule
            V & \( \kappa \theta + (u_0 - \kappa \theta)  \exp(-\theta t) \) & \( \frac{\kappa \theta}{1-\kappa} \left( 1 - \exp(-(1 - \kappa) \theta t) \right) + u_0 \exp(-(1 - \kappa) \theta t) \) \\[.3cm]
            \midrule
            VI & \( \sin(t) + \alpha \) & \( -\frac{\alpha}{\kappa - 1} + \frac{\kappa}{\kappa - 1} \frac{\alpha + \alpha \theta^{2} - 2 \alpha \kappa \theta^{2} + \alpha \kappa^{2} \theta^{2} + \theta \kappa - \theta}{1 + \theta^{2} - 2\kappa \theta^{2} + \kappa^{2} \theta^{2}} \exp((k-1)\theta t) + \frac{\sin(t) + \theta^{2}\sin(t) - \kappa \theta^{2} \sin(t) - \kappa \theta \cos(t)}{1 + \theta^{2} - 2\kappa \theta^{2} + \kappa^{2} \theta^{2}} \) \\
            \bottomrule
        \end{tabular}
    \end{table*}
\end{landscape} 
\section{MBPP Calculations} \label{app:mbp_cala}

\subsection{MBPP Intensity Functions} \label{app:mbp_intensity}

We consider the closed form solutions of an MBPP intensity function \( \MBPintens(t) \) with exponential kernel for various immigrant intensity functions. We use Corollary 3 where we have
\begin{equation*}
    h(t) = k\theta \exp(( k-1)\theta t) \cdot \llbracket t > 0 \rrbracket.
\end{equation*}
and
\begin{align*}
    E(t) &= \delta(t) + h(t) \\
    \MBPintens(t) &= (E * s)(t) = s(t) + (h * s)(t).
\end{align*}
Specifically, we calculate each row of~\cref{tab:appendix_mbp_intensity}.

\bheader{Row I. } Consider \( s(t) = \delta(t - a) \) for \( a \geq 0 \). Thus,
\begin{align*}
    (h * s)(t)
    &= (h(t) * \delta(t - a)) \\
    &= h(t - a) \cdot \llbracket t > a \rrbracket.
\end{align*}

Together we have,
\begin{equation*}
    \MBPintens(t) = \delta(t - a) + \kappa \theta \exp( (k-1) \theta (t - a) ) \cdot \llbracket t > a \rrbracket.
\end{equation*}

\bheader{Row II. } Consider \( s(t) = \sum_{\immiOccurTime_{z} \in \immiOccurTimeSet_{T}} \delta(t - \immiOccurTime_{z}) \) for some set of events \( \immiOccurTimeSet_{T} \). As the MBPP intensity is a LTI (\cite{Rizoiu:2017}[Corollary 2.2]), we can simply express the MBPP intensity as the sum of Row I.

Together we have,
\begin{equation*}
    \MBPintens(t) = \sum_{\immiOccurTime_{z} \in \immiOccurTimeSet_{T}} \bigg[ \delta(t - \immiOccurTime_{z}) + \kappa \theta \exp( (k-1) \theta (t - \immiOccurTime_{z}) ) \cdot \llbracket t > \immiOccurTime_{z} \rrbracket \bigg].
\end{equation*}

\bheader{Row III. } Consider \( s(t) = \llbracket a < t \leq b \rrbracket \) an interval. Then,
\begin{align*}
    (h * s)(t)
    &= (h(t) * \llbracket a < t \leq b \rrbracket).
\end{align*}

We break the calculation into three cases: (1) \( t \leq a \); (2) \( a < t \leq b \); and (3) \( t > b \).

\underline{Case (1)}: Suppose that \( t \leq a \), then simply \( (h(t) * \llbracket a < t \leq b \rrbracket ) = 0 \).

\underline{Case (2)}: Suppose that \( a < t \leq b \). Then we have,
\begin{align*}
    (h(t) * \llbracket a < t \leq b \rrbracket)
    &= \int_{0}^{\infty} \llbracket a < \tau \leq b \rrbracket \cdot h(t - \tau) \, \d \tau \\
    &= \int_{a}^{t} h(t - \tau) \, \d \tau \\
    &= \frac{\kappa \theta}{(1 - \kappa) \theta} \exp( (k - 1) \theta (t - \tau) ) \bigg]^{\tau = t}_{\tau = a} \\
    &= \frac{\kappa}{1 - \kappa} \left[ 1 - \exp( (k - 1) \theta (t - a) ) \right].
\end{align*}

\underline{Case (3)}: Suppose that \( t > b \). Then we have,
\begin{align*}
    (h(t) * \llbracket a < t \leq b \rrbracket)
    &= \int_{0}^{\infty} \llbracket a < \tau \leq b \rrbracket \cdot h(t - \tau) \, \d \tau \\
    &= \int_{a}^{b} h(t - \tau) \, \d \tau \\
    &= \frac{\kappa \theta}{(1 - \kappa) \theta} \cdot \exp( (k - 1) \theta (t - \tau) ) \bigg]^{\tau = b}_{\tau = a} \\
    &= \frac{\kappa}{1 - \kappa} \left[ \exp( (k - 1) \theta (t - b) ) - \exp( (k - 1) \theta (t - a) ) \right].
\end{align*}

Thus together we have that
\[
    (h(t) * \llbracket a < t \leq b \rrbracket) = \begin{cases}
        0 & t \leq a \\
        \frac{\kappa}{1 - \kappa} \left[ 1 - \exp( (k - 1) \theta (t - a) ) \right] & a < t \leq b \\
        \frac{\kappa}{1 - \kappa} \left[ \exp( (k - 1) \theta (t - b) ) - \exp( (k - 1) \theta (t - a) ) \right] & t > b \\
    \end{cases}.
\]

Giving,
\begin{align*}
    \MBPintens(t) 
    &= \llbracket a < t \leq b \rrbracket + \frac{\kappa}{1 - \kappa} \left[ 1 - \exp( (k - 1) \theta (t - a) ) \right] \cdot \llbracket a < t \leq b \rrbracket \\
    &\hspace{2.5cm}+ \frac{\kappa}{1 - \kappa} \left[ \exp( (k - 1) \theta (t - b) ) - \exp( (k - 1) \theta (t - a) ) \right] \cdot \llbracket t > b \rrbracket \\
    &= \frac{\kappa}{1 - \kappa} \left[ \frac{1}{k} - \exp( (k - 1) \theta (t - a) ) \right] \cdot \llbracket a < t \leq b \rrbracket \\
    &\hspace{2.5cm}+ \frac{\kappa}{1 - \kappa} \left[ \exp( (k - 1) \theta (t - b) ) - \exp( (k - 1) \theta (t - a) ) \right] \cdot \llbracket t > b \rrbracket.
\end{align*}

\bheader{Row IV. } Consider \( s(t) = \sum_{i=1}^{m} \frac{\immiObserVolume(\immiTime_{i-1}, \immiTime_{i}]}{\immiTime_{i} - \immiTime_{i-1}} \cdot \llbracket \immiTime_{i-1} < t \leq \immiTime_{i} \rrbracket \) for a set of intervals with non-negative immigrant volumes \( \immiObserVolume(\immiTime_{i-1}, \immiTime_{i}] \). As the MBPP intensity is an LTI (\cite{Rizoiu:2017}[Corollary 2.2]), we can simply express the MBPP intensity as the weighted sum of Row III.
\begin{align*}
    \MBPintens(t) &= 
    \sum_{i=1}^{m} \frac{\immiObserVolume(\immiTime_{i-1}, \immiTime_{i}]}{\immiTime_{i} - \immiTime_{i-1}} \left[ \frac{\kappa}{1 - \kappa} \left[ \frac{1}{k} - \exp( (k - 1) \theta (t - \immiTime_{i-1}) ) \right] \cdot \llbracket \immiTime_{i-1} < t \leq \immiTime_{i} \rrbracket \right.\\
    &\left. \hspace{3cm}+ \frac{\kappa}{1 - \kappa} \left[ \exp( (k - 1) \theta (t - \immiTime_{i}) ) - \exp( (k - 1) \theta (t - \immiTime_{i-1}) ) \right] \cdot \llbracket t > \immiTime_{i} \rrbracket \right].
\end{align*}

\bheader{Row V. } Proof shown in the main text.

\bheader{Row VI. } Consider \( s(t) = \sin(t) + \alpha \). Then
\begin{align*}
    (h * s)(t)
    &= (h * \alpha)(t) + (h * \sin)(t).
\end{align*}

For \( (h * \alpha)(t) \) we have,
\begin{align*}
    (h * \alpha)(t)
    &= \alpha \int_{0}^{t} h(\tau) \, \d \tau \\
    &= \frac{\alpha\kappa}{\kappa - 1} \left( \exp((k-1)\theta t) - 1 \right) \\
    &= \frac{\alpha\kappa}{\kappa - 1} \exp((k-1)\theta t) - \frac{\alpha\kappa}{\kappa - 1}.
\end{align*}

For \( (h * \sin)(t) \) first consider the indefinite integral
\begin{align*}
    \triangle = \int \sin(\tau) \exp(a(t - \tau)) \d \tau.
\end{align*}

By integration by parts,
\begin{align*}
    -\frac{1}{a}\sin(\tau) \exp(a(t-\tau)) + \frac{1}{a} \int \cos(\tau) \exp(a(t - \tau)) \d \tau.
\end{align*}

Then by integration by parts again,
\begin{align*}
    &-\frac{1}{a}\sin(\tau) \exp(a(t-\tau)) + \frac{1}{a} \left[ -\frac{1}{a} \cos(\tau)\exp(a(t - \tau)) - \frac{1}{a} \int \sin(\tau) \exp(a(t - \tau)) \d \tau \right] \\
    &= -\frac{1}{a}\sin(\tau) \exp(a(t-\tau)) -\frac{1}{a^{2}} \cos(\tau)\exp(a(t - \tau)) - \frac{1}{a^{2}} \int \sin(\tau) \exp(a(t - \tau)) \d \tau \\
    &= -\exp(a(t-\tau)) \left[ \frac{1}{a}\sin(\tau) + \frac{1}{a^{2}} \cos(\tau) \right] - \frac{1}{a^{2}} \int \sin(\tau) \exp(a(t - \tau)) \d \tau \\
    &= -\exp(a(t-\tau)) \left[ \frac{1}{a}\sin(\tau) + \frac{1}{a^{2}} \cos(\tau) \right] - \frac{1}{a^{2}} \triangle.
\end{align*}

Thus,
\begin{align*}
    \triangle &= -\exp(a(t-\tau)) \left[ \frac{1}{a}\sin(\tau) + \frac{1}{a^{2}} \cos(\tau) \right] - \frac{1}{a^{2}} \triangle \\
    (\iff) \hspace{2cm} \triangle &= \frac{-\exp(a(t-\tau)) \left[ a\sin(\tau) + \cos(\tau) \right]}{a^{2} + 1}.
\end{align*}

Therefore for \( (h * \sin)(t) \) we have
\begin{align*}
    (h * \sin)(t)
    &= \kappa \theta \frac{-\exp((k-1)\theta(t-\tau)) \left[ (k-1)\theta\sin(\tau) + \cos(\tau) \right]}{(k-1)^{2}\theta^{2} + 1} \Biggl]_{\tau=0}^{\tau=t} \\
    &= \kappa \theta \frac{-\exp((k-1)\theta(t-\tau)) \left[ (k-1)\theta\sin(\tau) + \cos(\tau) \right]}{1 + \theta^{2} - 2\kappa \theta^{2} + \kappa^{2} \theta^{2}} \Biggl]_{\tau=0}^{\tau=t} \\
    &= -\kappa \theta \frac{ \left[ (k-1)\theta\sin(t) + \cos(t) \right]}{1 + \theta^{2} - 2\kappa \theta^{2} + \kappa^{2} \theta^{2}} + \kappa \theta \frac{\exp((k-1)\theta t) }{1 + \theta^{2} - 2\kappa \theta^{2} + \kappa^{2} \theta^{2}}.
\end{align*}

Thus for the MBPP intensity,
\begin{align*}
    \MBPintens(t)
    &= \sin(t) + \alpha + (h * \sin)(t) \\
    &= \sin(t) + \alpha + \frac{\alpha\kappa}{\kappa - 1} \exp((k-1)\theta t) - \frac{\alpha\kappa}{\kappa - 1} \\
    &\hspace{2cm}-\kappa \theta \frac{ \left[ (k-1)\theta\sin(t) + \cos(t) \right]}{1 + \theta^{2} - 2\kappa \theta^{2} + \kappa^{2} \theta^{2}} + \kappa \theta \frac{\exp((k-1)\theta t) }{1 + \theta^{2} - 2\kappa \theta^{2} + \kappa^{2} \theta^{2}}.
\end{align*}

Now we collect the trigonometric terms, the exponential terms, and the other constants. The trigonometric terms:
\begin{align*}
    &\sin(t) -\kappa \theta \frac{ \left[ (k-1)\theta\sin(t) + \cos(t) \right]}{1 + \theta^{2} - 2\kappa \theta^{2} + \kappa^{2} \theta^{2}} \\
    &= \sin(t) - \frac{\kappa^{2}\theta^{2}\sin(t) - \kappa \theta^{2}\sin(t) + \kappa \theta \cos(t)}{1 + \theta^{2} - 2\kappa \theta^{2} + \kappa^{2} \theta^{2}} \\
    &= \frac{\sin(t) + \theta^{2}\sin(t) - 2 \kappa \theta^{2} \sin(t) + \kappa^{2}\theta^{2} \sin(t)- \kappa^{2}\theta^{2}\sin(t) + \kappa \theta^{2}\sin(t) - \kappa \theta \cos(t)}{1 + \theta^{2} - 2\kappa \theta^{2} + \kappa^{2} \theta^{2}} \\
    &= \frac{\sin(t) + \theta^{2}\sin(t) - \kappa \theta^{2} \sin(t) - \kappa \theta \cos(t)}{1 + \theta^{2} - 2\kappa \theta^{2} + \kappa^{2} \theta^{2}}.
\end{align*}
The exponential terms:
\begin{align*}
    &\frac{\alpha\kappa}{\kappa - 1} \exp((k-1)\theta t) + \kappa \theta \frac{\exp((k-1)\theta t) }{1 + \theta^{2} - 2\kappa \theta^{2} + \kappa^{2} \theta^{2}} \\
    &= \frac{\kappa}{\kappa - 1} \frac{\alpha + \alpha \theta^{2} - 2 \alpha \kappa \theta^{2} + \alpha \kappa^{2} \theta^{2} + \theta \kappa - \theta}{1 + \theta^{2} - 2\kappa \theta^{2} + \kappa^{2} \theta^{2}} \exp((k-1)\theta t).
\end{align*}
The constant terms:
\begin{align*}
    \alpha - \frac{\alpha\kappa}{\kappa - 1} = -\frac{\alpha}{\kappa - 1}.
\end{align*}

Thus together,
\begin{align*}
    \MBPintens(t) 
    &= -\frac{\alpha}{\kappa - 1} 
    + \frac{\kappa}{\kappa - 1} \frac{\alpha + \alpha \theta^{2} - 2 \alpha \kappa \theta^{2} + \alpha \kappa^{2} \theta^{2} + \theta \kappa - \theta}{1 + \theta^{2} - 2\kappa \theta^{2} + \kappa^{2} \theta^{2}} \exp((k-1)\theta t) \\
    & \hspace{2cm} + \frac{\sin(t) + \theta^{2}\sin(t) - \kappa \theta^{2} \sin(t) - \kappa \theta \cos(t)}{1 + \theta^{2} - 2\kappa \theta^{2} + \kappa^{2} \theta^{2}}.
\end{align*}

\newpage

\begin{landscape}
    \begin{table*}[htbp]
        \centering
        \caption{
            Solutions for different immigrant intensities \( \immiIntens(t) \), and the exponentially time-decaying kernel in \cref{eq:exp-kernel} of the main text. \( \immiOccurTimeSet_{T} \) acts as a set of immigrant event times.
        }\label{tab:appendix_mbp_compensator}
        \begin{tabular}{c|l|p{14cm}} 
            \toprule
            & \( \immiIntens(t) \) & MBPP Compensator Function \( \MBPcomp(t) \) \\
            \midrule
            I & \( \delta(t - a) \) & \( \llbracket t > a \rrbracket \left[ 1 + \frac{\kappa}{\kappa - 1} \bigl(\exp( (\kappa-1) \theta (t - a) ) - 1 \bigr)\right] \) \\[.3cm]
            \midrule
            II & \( \sum_{\immiOccurTime_{z} \in \immiOccurTimeSet_{T}} \delta(t - \immiOccurTime_{z}) \) & \( \sum_{\immiOccurTime_{z} \in \immiOccurTimeSet_{T}} \llbracket t > \immiOccurTime_{z} \rrbracket \cdot \left[ 1 + \frac{\kappa}{\kappa - 1} \bigl(\exp( \kappa -1) \theta (t - \immiOccurTime_{z}) ) - 1 \bigr) \right] \) \\[.3cm]
            \midrule
            III & \( \llbracket a \leq t < b \rrbracket \) & \( (t-a) \cdot \frac{\kappa}{1 - \kappa} \Bigl[ \frac{1}{k} - \exp( (k - 1) \theta (t - a) ) \Bigr] \cdot \llbracket a < t \leq b \rrbracket + \frac{\kappa t}{1 - \kappa} \Bigl[ \exp( (k - 1) \theta (t - b) ) - \exp( (k - 1) \theta (t - a) ) \Bigr] \cdot \llbracket t > b \rrbracket + \frac{\kappa b}{1 - \kappa} \Bigl[ \frac{1}{k} - \exp( (k - 1) \theta (t - b) ) \Bigr] \cdot \llbracket  t > b \rrbracket + \frac{\kappa a}{1 - \kappa} \Bigl[ \exp( (k - 1) \theta (t - a) ) - \frac{1}{k} \Bigr] \cdot \llbracket  t > b \rrbracket
            \) \\[.3cm]
            \midrule
            IV & \( \sum_{i=1}^{m} \frac{\immiObserVolume(\immiTime_{i-1}, \immiTime_{i}]}{\immiTime_{i} - \immiTime_{i-1}} \cdot \llbracket \immiTime_{i-1} < t \leq \immiTime_{i} \rrbracket \) & \( \frac{(t - \immiTime_{m^{\star}(t)}) \immiObserVolume(\immiTime_{m^{\star}(t)+ 1} - \immiTime_{m^{\star}(t)})}{\immiTime_{m^{\star}(t) + 1} - \immiTime_{m^{\star}(t)}} + \sum_{i=1}^{m^{\star}(t)} \immiObserVolume( \immiTime_{i-1}, \immiTime_{i}] + \sum_{i=1}^{m^{\star}(t)} \frac{\immiObserVolume(\immiTime_{i-1}, \immiTime_{i}]}{\immiTime_{i} - \immiTime_{i-1}} \frac{\kappa}{1 - \kappa} \Bigl[ \exp((\kappa - 1)\theta(t - \immiTime_{i})) - 1 - \exp((\kappa - 1)\theta(t - \immiTime_{i-1})) + \exp((\kappa - 1)\theta(\immiTime_{i} - \immiTime_{i-1})) \Bigr] + \frac{\kappa}{1 - \kappa} \Bigl[ t - \immiTime_{m^{\star}(t)} + \frac{1}{\theta (\kappa - 1)} - \frac{1}{\theta (\kappa - 1)} \exp((k-1) \theta (t - \immiTime_{m^{\star}(t)})) \, \d \tau \Bigr], \textrm{ where } m^{\star}(t) = 1 + \max \{ i \mid \immiTime_{i} > t\} \) \\[.3cm]
            \midrule
            V & \( \kappa \theta + (u_0 - \kappa \theta)  \exp(-\theta t) \) & \( \frac{\kappa \theta t}{(1 - \kappa)} + \frac{u_{0} - 1}{(1 - \kappa)\theta}(1 -  \exp(-(1 - \kappa) \theta t)) \) \\[.3cm]
            \midrule
            VI & \( \sin(t) + \alpha \) & \( -\frac{\alpha t}{\kappa - 1} + \frac{\kappa}{(\kappa - 1)^{2}\theta} \frac{\alpha + \alpha \theta^{2} - 2 \alpha \kappa \theta^{2} + \alpha \kappa^{2} \theta^{2} + \theta \kappa - \theta}{1 + \theta^{2} - 2\kappa \theta^{2} + \kappa^{2} \theta^{2}} (\exp((k-1)\theta \tau) - 1) + \frac{-\cos(t) - \theta^{2}\cos(t) + \kappa \theta^{2} \cos(t) - \kappa \theta \sin(t) + 1 + \theta^{2} - \kappa \theta^{2}}{1 + \theta^{2} - 2\kappa \theta^{2} + \kappa^{2} \theta^{2}} \) \\
            \bottomrule
        \end{tabular}
    \end{table*}
\end{landscape} 
\subsection{MBPP Compensator Functions} \label{app:mbp_compensator}

We similarly calculate the corresponding MBPP compensator function of the MBPP intensity function we looked at above. In particular we similarly calculate each row of~\cref{tab:appendix_mbp_compensator}.

\bheader{Row I. } Consider \( s(t) = \delta(t - a) \) for \( a \geq 0 \). We simply integrate the result of Row I in~\cref{tab:appendix_mbp_intensity},
\begin{align*}
    \MBPcomp(t)
    &= \int_{0}^{t} \MBPintens(\tau) \d \tau \\
    &= \int_{0}^{t} \left( \delta(\tau - a) + \kappa \theta \exp( (\kappa-1) \theta (\tau - a) ) \cdot \llbracket \tau > a \rrbracket  \right) \d \tau \\
    &= \int_{0}^{t} \delta(\tau - a) \d \tau + \int_{0}^{t} \kappa \theta \exp( (\kappa-1) \theta (\tau - a) ) \cdot \llbracket \tau > a \rrbracket \d \tau \\
    &= \llbracket t > a \rrbracket + \llbracket t > a \rrbracket \cdot \int_{a}^{t} \kappa \theta \exp( (\kappa-1) \theta (\tau - a) ) \d \tau \\
    &= \llbracket t > a \rrbracket + \llbracket t > a \rrbracket \cdot \frac{\kappa}{\kappa - 1} \exp( (\kappa-1) \theta (\tau - a) ) \Biggr]_{\tau = a}^{\tau = t} \\
    &= \llbracket t > a \rrbracket + \llbracket t > a \rrbracket \cdot \frac{\kappa}{\kappa - 1} \bigl(\exp( (\kappa-1) \theta (t - a) ) - 1 \bigr) \\
    &= \llbracket t > a \rrbracket  \cdot \left[1 + \frac{\kappa}{\kappa - 1} \bigl(\exp( (\kappa-1) \theta (t - a) ) - 1 \bigr) \right].
\end{align*}

\bheader{Row II. } Consider \( s(t) = \sum_{\immiOccurTime_{z} \in \immiOccurTimeSet_{T}} \delta(t - \immiOccurTime_{z}) \) for some set of events \( \immiOccurTimeSet_{T} \). As MBPP compensator is LTI, we compute the weighted sum of Row II.
\begin{align*}
    \MBPcomp(t)
    &= \sum_{\immiOccurTime_{z} \in \immiOccurTimeSet_{T}} \llbracket t > \immiOccurTime_{z} \rrbracket \cdot 
    \left[ 1 + \frac{\kappa}{\kappa - 1} \bigl(\exp( (\kappa-1) \theta (t - \immiOccurTime_{z}) ) - 1 \bigr) \right].
\end{align*}

\bheader{Row III. } Consider \( s(t) = \llbracket a < t \leq b \rrbracket \) an interval. First we consider the integral of \( s(t) \).
\begin{align*}
    \immiCounts(t)
    &= \int_{0}^{t} s(\tau) \d \tau = (t-a) \cdot \llbracket a < t \leq b \rrbracket + (b - a) \cdot \llbracket t > b \rrbracket,
\end{align*}
which can be simply computed by considering different values of \( t \).

As the MBPP compensator is a LTI, the result can be calculated as the weighted sum of of Row III in~\cref{tab:appendix_mbp_intensity}.

The first term,
\begin{align*}
    &(t-a) \cdot \frac{\kappa}{1 - \kappa} \Bigl[ \frac{1}{k} - \exp( (k - 1) \theta (t - a) ) \Bigr] \cdot \llbracket a < t \leq b \rrbracket \\
    &\hspace{1cm}+ (t-a) \cdot \frac{\kappa}{1 - \kappa} \Bigl[ \exp( (k - 1) \theta (t - b) ) - \exp( (k - 1) \theta (t - a) ) \Bigr] \cdot \llbracket t > b \rrbracket.
\end{align*}

The second term,
\begin{align*}
    (b-a) \cdot \frac{\kappa}{1 - \kappa} \Bigl[ \frac{1}{k} - \exp( (k - 1) \theta (t - b) ) \Bigr] \cdot \llbracket  t > b \rrbracket.
\end{align*}

Together we have
\begin{align*}
    \MBPcomp(t)
    &= (t-a) \cdot \frac{\kappa}{1 - \kappa} \Bigl[ \frac{1}{k} - \exp( (k - 1) \theta (t - a) ) \Bigr] \cdot \llbracket a < t \leq b \rrbracket \\
    &\hspace{1cm}+ (t-a) \cdot \frac{\kappa}{1 - \kappa} \Bigl[ \exp( (k - 1) \theta (t - b) ) - \exp( (k - 1) \theta (t - a) ) \Bigr] \cdot \llbracket t > b \rrbracket \\
    &\hspace{1cm}+ (b-a) \cdot \frac{\kappa}{1 - \kappa} \Bigl[ \frac{1}{k} - \exp( (k - 1) \theta (t - b) ) \Bigr] \cdot \llbracket  t > b \rrbracket \\
    &= (t-a) \cdot \frac{\kappa}{1 - \kappa} \Bigl[ \frac{1}{k} - \exp( (k - 1) \theta (t - a) ) \Bigr] \cdot \llbracket a < t \leq b \rrbracket \\
    &\hspace{1cm}+ \frac{\kappa t}{1 - \kappa} \Bigl[ \exp( (k - 1) \theta (t - b) ) - \exp( (k - 1) \theta (t - a) ) \Bigr] \cdot \llbracket t > b \rrbracket \\
    &\hspace{1cm}+ \frac{\kappa b}{1 - \kappa} \Bigl[ \frac{1}{k} - \exp( (k - 1) \theta (t - b) ) \Bigr] \cdot \llbracket  t > b \rrbracket \\
    &\hspace{1cm}+ \frac{\kappa a}{1 - \kappa} \Bigl[ \exp( (k - 1) \theta (t - a) ) - \frac{1}{k} \Bigr] \cdot \llbracket  t > b \rrbracket \\
\end{align*}

\bheader{Row IV. } To calculate this row, one could utilise the LTI with Row III. However, we derive this differently, which provides an equation which is easier to implement.

First, let us calculate the exogenous component.
\begin{align*}
    \immiComp(t)
    &= \int_{0}^{t} \immiIntens(\tau) \, \d \tau \\
    &= \int_{0}^{t} \left( \sum_{i=1}^{m} \frac{\immiObserVolume(\immiTime_{i-1}, \immiTime_{i}]}{\immiTime_{i} - \immiTime_{i-1}} \cdot \llbracket \immiTime_{i-1} < \tau \leq \immiTime_{i} \rrbracket \right) \, \d \tau \\
    &= \sum_{i=1}^{m} \frac{\immiObserVolume(\immiTime_{i-1}, \immiTime_{i}]}{\immiTime_{i} - \immiTime_{i-1}} \cdot \int_{0}^{t} \llbracket \immiTime_{i-1} < \tau \leq \immiTime_{i} \rrbracket  \, \d \tau
\end{align*}

The integral within the summation is what we need to calculate now \( \int_{0}^{t} \llbracket \immiTime_{i-1} < \tau \leq \immiTime_{i} \rrbracket  \, \d \tau \). We split up the calculation into three cases: (1) \( t \leq \immiTime_{i-1} \); (2) \( \immiTime_{i-1} < t \leq \immiTime_{i} \); and (3) \( t > \immiTime_{i} \).

\underline{Case (1)}: Suppose that \( t \leq \immiTime_{i-1} \), then simply \( \int_{0}^{t} \llbracket \immiTime_{i-1} < \tau \leq \immiTime_{i} \rrbracket  \, \d \tau = 0 \).

\underline{Case (2)}: Suppose that \( \immiTime_{i-1} < t \leq \immiTime_{i} \),
\[
    \int_{0}^{t} \llbracket \immiTime_{i-1} < \tau \leq \immiTime_{i} \rrbracket  \, \d \tau
    = \int_{\immiTime_{i-1}}^{t} 1 \, \d \tau 
    = t - \immiTime_{i-1}.
\]

\underline{Case (3)}: Suppose that \( t > \immiTime_{i} \),
\[
    \int_{0}^{t} \llbracket \immiTime_{i-1} < \tau \leq \immiTime_{i} \rrbracket  \, \d \tau
    = \int_{\immiTime_{i-1}}^{\immiTime_{i}} 1 \, \d \tau 
    = \immiTime_{i} - \immiTime_{i-1}.
\]

For arbitrary fixed \( t \), let \( m^{\star}(t) = 1 + \max \{ i \mid \immiTime_{i} > t \} \). Then, by combining the three cases we have that,
\[
    \immiComp(t) = \frac{(t - \immiTime_{m^{\star}(t)}) \immiObserVolume(\immiTime_{m^{\star}(t)+ 1} - \immiTime_{m^{\star}(t)})}{\immiTime_{m^{\star}(t) + 1} - \immiTime_{m^{\star}(t)}} + \sum_{i=1}^{m^{\star}(t)} \immiObserVolume( \immiTime_{i-1}, \immiTime_{i}].
\]

Similarly to the exogenous function, we consider the endogenous response in the same three cases. We directly use the piecewise derivations found in Row III of~\cref{tab:appendix_mbp_intensity}. Thus, we only need to compute \( \int_{0}^{t} (h * \immiIntens) \, \d \tau \). First we consider \( \int_{0}^{t} (h(\tau) * \llbracket \immiTime_{i-1} < \tau \leq \immiTime_{i} \rrbracket) \, \d \tau \)

\textbf{Case (1)}: Suppose that \( t \leq \immiTime_{i-1} \), then simply \( \int_{0}^{t} (h(\tau) * \llbracket \immiTime_{i-1} < \tau \leq \immiTime_{i} \rrbracket) \llbracket t \leq \immiTime_{i-1} \rrbracket \, \d \tau = 0 \).

\textbf{Case (2)}: Suppose that \( \immiTime_{i-1} < t \leq \immiTime_{i} \),
\begin{align*}
    &\int_{0}^{t} (h(\tau) * \llbracket \immiTime_{i-1} < \tau \leq \immiTime_{i} \rrbracket) \llbracket \immiTime_{i-1} < \tau \leq \immiTime_{i} \rrbracket \, \d \tau \\
    &= \int_{\immiTime_{i-1}}^{t} (h * \immiIntens)(\tau) \, \d \tau \\
    &= \int_{\immiTime_{i-1}}^{t} \frac{\kappa}{1 - \kappa} \left[ 1 - \exp((k-1) \theta (\tau - \immiTime_{i-1})) \right] \, \d \tau \\
    &= \frac{\kappa}{1 - \kappa} \left[ t - \immiTime_{i-1} - \int_{\immiTime_{i-1}}^{t} \exp((k-1) \theta (\tau - \immiTime_{i-1})) \, \d \tau \right] \\
    &= \frac{\kappa}{1 - \kappa} \left[ t - \immiTime_{i-1} + \frac{1}{\theta (\kappa - 1)} - \frac{1}{\theta (\kappa - 1)} \exp((k-1) \theta (t - \immiTime_{i-1})) \, \d \tau \right].
\end{align*}

\textbf{Case (3)}: Suppose that \( t > \immiTime_{i} \),
\begin{align*}
    &\int_{0}^{t} (h(\tau) * \llbracket \immiTime_{i-1} < \tau \leq \immiTime_{i} \rrbracket) \llbracket \tau > \immiTime_{i} \rrbracket \, \d \tau \\
    &= \int_{\immiTime_{i}}^{t} (h * \immiIntens)(\tau) \, \d \tau \\
    &= \int_{\immiTime_{i}}^{t} \frac{\kappa}{1 - \kappa} \left[ \exp((\kappa - 1)\theta(\tau - \immiTime_{i})) - \exp((\kappa - 1)\theta(\tau - \immiTime_{i-1})) \right] \, \d \tau \\
    &= \frac{\kappa}{1 - \kappa} \left[ \exp((\kappa - 1)\theta(t - \immiTime_{i})) - 1 - \exp((\kappa - 1)\theta(t - \immiTime_{i-1})) \right. \\ &\left.\quad \quad \quad \quad + \exp((\kappa - 1)\theta(\immiTime_{i} - \immiTime_{i-1})) \right].
\end{align*}

Thus, combining the three cases, we have that,
\begin{align*}
    &\int_{0}^{t} (h * \immiIntens)(\tau) \, \d \tau \\
    &= \sum_{i=1}^{m} \frac{\immiObserVolume(\immiTime_{i-1}, \immiTime_{i}]}{\immiTime_{i} - \immiTime_{i-1}} \int_{0}^{t} (h(t) * \llbracket \immiTime_{i-1} < t \leq \immiTime_{i} \rrbracket) \, \d \tau \\
    &= \sum_{i=1}^{m^{\star}(t)} \frac{\immiObserVolume(\immiTime_{i-1}, \immiTime_{i}]}{\immiTime_{i} - \immiTime_{i-1}} \frac{\kappa}{1 - \kappa} \left[ \exp((\kappa - 1)\theta(t - \immiTime_{i})) - 1 - \exp((\kappa - 1)\theta(t - \immiTime_{i-1})) \right. \\ &\left.\quad \quad \quad \quad + \exp((\kappa - 1)\theta(\immiTime_{i} - \immiTime_{i-1})) \right] + \\ &\quad \quad \quad \quad \frac{\kappa}{1 - \kappa} \left[ t - \immiTime_{m^{\star}(t)} + \frac{1}{\theta (\kappa - 1)} - \frac{1}{\theta (\kappa - 1)} \exp((k-1) \theta (t - \immiTime_{m^{\star}(t)})) \, \d \tau \right].
\end{align*}

Together, we can now compute the compensator function.
\begin{align*}
    \MBPcomp(t) &= \frac{(t - \immiTime_{m^{\star}(t)}) \immiObserVolume(\immiTime_{m^{\star}(t)+ 1} - \immiTime_{m^{\star}(t)})}{\immiTime_{m^{\star}(t) + 1} - \immiTime_{m^{\star}(t)}} + \sum_{i=1}^{m^{\star}(t)} \immiObserVolume( \immiTime_{i-1}, \immiTime_{i}] \\
    & \hspace{1.5cm} + \sum_{i=1}^{m^{\star}(t)} \frac{\immiObserVolume(\immiTime_{i-1}, \immiTime_{i}]}{\immiTime_{i} - \immiTime_{i-1}} \frac{\kappa}{1 - \kappa} \left[ \exp((\kappa - 1)\theta(t - \immiTime_{i})) - 1 - \exp((\kappa - 1)\theta(t - \immiTime_{i-1})) \right. \\
    &\left. \hspace{1.5cm} + \exp((\kappa - 1)\theta(\immiTime_{i} - \immiTime_{i-1})) \right] \\
    & \hspace{1.5cm} + \frac{\kappa}{1 - \kappa} \left[ t - \immiTime_{m^{\star}(t)} + \frac{1}{\theta (\kappa - 1)} - \frac{1}{\theta (\kappa - 1)} \exp((k-1) \theta (t - \immiTime_{m^{\star}(t)})) \, \d \tau \right].
\end{align*}

\bheader{Row V. } To calculate the MBPP compensator, we integrate Row V of Table~\ref{tab:appendix_mbp_intensity}.
\begin{align*}
    \MBPcomp(t)
    &= \int_{0}^{t} \MBPintens(\tau) \d \tau \\
    &= \int_{0}^{t} \frac{\kappa \theta}{1-\kappa} \left( 1 - \exp(-(1 - \kappa) \theta \tau) \right) + u_0 \exp(-(1 - \kappa) \theta \tau) \d \tau \\
    &= \frac{\kappa \theta}{(1 - \kappa)}\left[ \tau + \frac{1}{(1 - \kappa)\theta}\exp(-(1 - \kappa)\theta \tau) \right] - \frac{u_{0}}{(1 - \kappa)\theta} \exp(-(1 - \kappa) \theta \tau) \Bigg]_{\tau = 0}^{\tau = t} \\
    &= \frac{\kappa \theta t}{(1 - \kappa)} + \frac{u_{0} - 1}{(1 - \kappa)\theta}(1 -  \exp(-(1 - \kappa) \theta t)).
\end{align*}

\bheader{Row VI. } To calculate the MBPP compensator, we integrate Row VI of Table~\ref{tab:appendix_mbp_intensity}.
\begin{align*}
    \MBPcomp(t)
    = \int_{0}^{t} \MBPintens(\tau) \d \tau 
    &= \int_{0}^{t} \Bigl[ -\frac{\alpha}{\kappa - 1} 
    + \frac{\kappa}{\kappa - 1} \frac{\alpha + \alpha \theta^{2} - 2 \alpha \kappa \theta^{2} + \alpha \kappa^{2} \theta^{2} + \theta \kappa - \theta}{1 + \theta^{2} - 2\kappa \theta^{2} + \kappa^{2} \theta^{2}} \exp((k-1)\theta \tau) \\
    & \hspace{1.2cm} + \frac{\sin(\tau) + \theta^{2}\sin(\tau) - \kappa \theta^{2} \sin(\tau) - \kappa \theta \cos(\tau)}{1 + \theta^{2} - 2\kappa \theta^{2} + \kappa^{2} \theta^{2}} \Bigr] \d \tau.
\end{align*}

We calculate the integral term by term. For the first term,
\begin{align*}
    \int_{0}^{t} -\frac{\alpha}{\kappa - 1} \d \tau = -\frac{\alpha t}{\kappa - 1}.
\end{align*}

For the second term,
\begin{align*}
    &\int_{0}^{t} \frac{\kappa}{\kappa - 1} \frac{\alpha + \alpha \theta^{2} - 2 \alpha \kappa \theta^{2} + \alpha \kappa^{2} \theta^{2} + \theta \kappa - \theta}{1 + \theta^{2} - 2\kappa \theta^{2} + \kappa^{2} \theta^{2}} \exp((k-1)\theta \tau) \d \tau \\
    &= \frac{\kappa}{\kappa - 1} \frac{\alpha + \alpha \theta^{2} - 2 \alpha \kappa \theta^{2} + \alpha \kappa^{2} \theta^{2} + \theta \kappa - \theta}{1 + \theta^{2} - 2\kappa \theta^{2} + \kappa^{2} \theta^{2}} \int_{0}^{t} \exp((k-1)\theta \tau) \d \tau \\
    &= \frac{\kappa}{(\kappa - 1)^{2}\theta} \frac{\alpha + \alpha \theta^{2} - 2 \alpha \kappa \theta^{2} + \alpha \kappa^{2} \theta^{2} + \theta \kappa - \theta}{1 + \theta^{2} - 2\kappa \theta^{2} + \kappa^{2} \theta^{2}} (\exp((k-1)\theta \tau) - 1).
\end{align*}

For the third term,
\begin{align*}
    &\int_{0}^{t} \frac{\sin(\tau) + \theta^{2}\sin(\tau) - \kappa \theta^{2} \sin(\tau) - \kappa \theta \cos(\tau)}{1 + \theta^{2} - 2\kappa \theta^{2} + \kappa^{2} \theta^{2}} \d \tau \\
    &=\frac{-\cos(\tau) - \theta^{2}\cos(\tau) + \kappa \theta^{2} \cos(\tau) - \kappa \theta \sin(\tau)}{1 + \theta^{2} - 2\kappa \theta^{2} + \kappa^{2} \theta^{2}} \Bigg]_{\tau = 0}^{\tau = t} \\
    &= \frac{-\cos(t) - \theta^{2}\cos(t) + \kappa \theta^{2} \cos(t) - \kappa \theta \sin(t) + 1 + \theta^{2} - \kappa \theta^{2}}{1 + \theta^{2} - 2\kappa \theta^{2} + \kappa^{2} \theta^{2}}.
\end{align*}

Thus together,
\begin{align*}
    \MBPcomp(t)
    &= -\frac{\alpha t}{\kappa - 1}
    + \frac{\kappa}{(\kappa - 1)^{2}\theta} \frac{\alpha + \alpha \theta^{2} - 2 \alpha \kappa \theta^{2} + \alpha \kappa^{2} \theta^{2} + \theta \kappa - \theta}{1 + \theta^{2} - 2\kappa \theta^{2} + \kappa^{2} \theta^{2}} (\exp((k-1)\theta \tau) - 1) \\
    & \hspace{1.2cm} + \frac{-\cos(t) - \theta^{2}\cos(t) + \kappa \theta^{2} \cos(t) - \kappa \theta \sin(t) + 1 + \theta^{2} - \kappa \theta^{2}}{1 + \theta^{2} - 2\kappa \theta^{2} + \kappa^{2} \theta^{2}}.
\end{align*} 
\section{Latent Homogeneous Poisson Process Exogenous Function - Probabilistic Approach}
\label{app:lhpp_prob}

The Latent Homogenous Poisson Process (LHPP) exogenous function presented in~\cref{subsec:exogenous_interval_censored} can be considered in a probabilistic approach. Instead of assigning a point process to dictate the interval-censored exogenous events, we consider a probability distribtion \( p(\immiOccurTime) \) corresponding to the probability an exogenous event occurs from \( (0, T] \). Furthermore, we have the corresponding immigrant event times which are sampled from the probability distributions \( \immiOccurTime \in \immiOccurTimeSet_{T} \).

We first consider the impulse response MBPP with the position of the impulse as a parameter,
\begin{equation}
    \MBPintens_{\delta}(t; \immiOccurTime) \defEq \delta(t - \immiOccurTime) + \int_{0}^{t} \phi(s) \cdot \MBPintens_{\delta}(t - s; \immiOccurTime) \d s.
\end{equation}

Then, we can consider the MBPP corresponding to the expectation of the intensity function with respect to the position of the impulse:
\begin{align*}
    \E{\immiOccurTime \sim p(\immiOccurTime)} {\MBPintens_{\delta}(t; \immiOccurTime) }
    &= \E{\immiOccurTime \sim p(\immiOccurTime)} {\delta(t - \immiOccurTime) + \int_{0}^{t} \phi(s) \cdot \MBPintens_{\delta}(t - s; \immiOccurTime) \d s} \\
    &= \E{\immiOccurTime \sim p(\immiOccurTime)} {\delta(t - \immiOccurTime)} + \E{\immiOccurTime \sim p(\immiOccurTime)} {\int_{0}^{t} \phi(s) \cdot \MBPintens_{\delta}(t - s; \immiOccurTime) \d s} \\
    &= \E{\immiOccurTime \sim p(\immiOccurTime)} {\delta(t - \immiOccurTime)} + \int_{0}^{t} \phi(s) \cdot \E{\immiOccurTime \sim p(\immiOccurTime)} {\MBPintens_{\delta}(t - s; \immiOccurTime)} \d s \\
    &=  \int_{0}^{T} {\delta(t - \immiOccurTime) \cdot p(\immiOccurTime)} \d \immiOccurTime + \int_{0}^{t} \phi(s) \cdot \E{\immiOccurTime \sim p(\immiOccurTime)} {\MBPintens_{\delta}(t - s; \immiOccurTime)} \d s \\
    &= p(t) + \int_{0}^{t} \phi(s) \cdot \E{\immiOccurTime \sim p(\immiOccurTime)} {\MBPintens_{\delta}(t - s; \immiOccurTime)} \d s.
\end{align*}

We now consider probability distributions \( p_{i}(\immiOccurTime) \) defined as a uniform distribution over the exogenous event observation period \( ( \immiTime_{i-1}, \immiTime_{i} ] \), for \( i \in \{\immiTime_{1} ,\ldots, \immiTime_{m} \} \):
\begin{align*}
    p_{i}(\immiOccurTime)
    &=
    \begin{cases}
    \frac{1}{\immiTime_{i} - \immiTime_{i-1}} & \immiOccurTime \in (\immiTime_{i-1}, \immiTime_{i}] \\
    0 & \textrm{otherwise}
    \end{cases} \\
    &= \frac{1}{\immiTime_{i} - \immiTime_{i-1}} \cdot \llbracket \immiTime_{i-1} < \immiOccurTime \leq \immiTime_{i} \rrbracket.
\end{align*}

The probabilistic interpretation of the LHPP exogenous function can now be expressed as the weighted sum of MBPP impulse expectations, where the weights are given by exogenous event volumes \( \immiObserVolume(\immiTime_{i-1}, \immiTime_{i}] \) and the summation is over the probability distribution \( p_{i}(\immiOccurTime) \). That is, the intensity function is given by:
\begin{align*}
    \MBPintens_{\rectangle}(t)
    &\defEq \sum_{i=1}^{m} \immiObserVolume(\immiTime_{i-1}, \immiTime_{i}] \cdot \E{\immiOccurTime \sim p_{i}(\immiOccurTime)} {\MBPintens_{\delta}(t; \immiOccurTime) } \\
    &= \sum_{i=1}^{m} \immiObserVolume(\immiTime_{i-1}, \immiTime_{i}] \cdot \left[ p_{i}(t) + \int_{0}^{t} \phi(s) \cdot \E{\immiOccurTime \sim p_{i}(\immiOccurTime)} {\MBPintens_{\delta}(t - s; \immiOccurTime)} \d s \right] \\
    &= \sum_{i=1}^{m} \immiObserVolume(\immiTime_{i-1}, \immiTime_{i}] \cdot p_{i}(t) + \sum_{i=1}^{m} \immiObserVolume(\immiTime_{i-1}, \immiTime_{i}] \cdot \left[\int_{0}^{t} \phi(s) \cdot \E{\immiOccurTime \sim p_{i}(\immiOccurTime)} {\MBPintens_{\delta}(t - s; \immiOccurTime)} \d s \right] \\
    &= \sum_{i=1}^{m} \immiObserVolume(\immiTime_{i-1}, \immiTime_{i}] \cdot p_{i}(t) + \int_{0}^{t} \phi(s) \cdot  \left[\sum_{i=1}^{m} \immiObserVolume(\immiTime_{i-1}, \immiTime_{i}] \cdot \E{\immiOccurTime \sim p_{i}(\immiOccurTime)} {\MBPintens_{\delta}(t - s; \immiOccurTime)} \right] \d s \\
    &= \sum_{i=1}^{m} \immiObserVolume(\immiTime_{i-1}, \immiTime_{i}] \cdot p_{i}(t) + \int_{0}^{t} \phi(s) \cdot \MBPintens_{\rectangle}(t - s) \d s \\
    &= \sum_{i=1}^{m} \frac{\immiObserVolume(\immiTime_{i-1}, \immiTime_{i}]}{\immiTime_{i} - \immiTime_{i-1}} \cdot \llbracket \immiTime_{i-1} < t \leq \immiTime_{i} \rrbracket + \int_{0}^{t} \phi(s) \cdot \MBPintens_{\rectangle}(t - s) \d s,
\end{align*}
where the exogenous term of the intensity function is the LHPP exogenous function \( \immiIntensStep(t) \), \cref{eq:latentexogfunc}.

\section{Convexity analysis}
\label{app:convexity analysis}

This section analyzes the loss functions of the Hawkes process and MBPP using the standard point process log-likelihood. 
We find that with the exponential triggering kernel and multi-impulse response in Equation~\eqref{eq:multi-impulse exogenous function}, both the Hawkes process and MBPP are convex in $\kappa$ for fixed $\theta$.

To account for exogenous functions, the exogenous events \( \immiOccurTimeSet_T \) and the endogenous events \( \offspringOccurTimeSet_T \) are considered to be separable. 

\subsection{Hawkes process}

It is trivial to verify that the negative log-likelihood functions corresponding to the exogenous functions below are convex in $\kappa$, as the terms consist of a negative-log function and linear functions which are convex functions of $\kappa$, but not $\theta$.

\begin{itemize}
\item Multiple impulse response: Equation~\eqref{eq:multi-impulse exogenous function}
\begin{align}
    \mathcal{L}(\theta;T)&\doteq -\sum_{t_n\in \Omega_T}\log\left(\sum_{\immiOccurTime_j\in\Psi_T} \delta(t_n-\immiOccurTime_j)+\sum_{t_i < t_n} \kappa\theta e^{-\theta(t_n-t_i)}\right) \nonumber \\
    &\quad +\vert \Psi_T\vert + \kappa \sum_{t_i\in\Omega_T} (1-e^{-\theta(T-t_i)}).
\end{align}
	\item Step response: $s(t) = \llbracket t \geq 0 \rrbracket$
\begin{align}
	\mathcal{L}(\theta;T)&\doteq -\sum_{t_n\in \Omega_T}\log\left(\llbracket t_n \geq 0 \rrbracket+\sum_{t_i < t_n} \kappa\theta e^{-\theta(t_n-t_i)}\right) \nonumber \\
	&\quad   +T+\kappa\sum_{t_i\in\Omega_T}(1-e^{-\theta(T-t_i)}).
\end{align}    	
	\item Dassios response: Equation~\eqref{eq:dassio response}
\begin{align}
	\mathcal{L}(\theta;T)&\doteq -\sum_{t_n\in \Omega_T}\log\left(\kappa\theta+(u_0-\kappa\theta)e^{-\theta t_n}+\sum_{t_i < t_n} \kappa\theta e^{-\theta(t_n-t_i)}\right) \nonumber \\
	&+\kappa\theta T+\frac{u_0-\kappa\theta}{\theta}(1-e^{-\theta T})+\kappa\sum_{t_i\in\Omega_T}(1-e^{-\theta(T-t_i)})
\end{align}
\end{itemize}

\subsection{Mean Behaviour Poisson}

The negative log likelihood for the multiple impulse response is given by
\begin{align}\label{mbpmimp}
	\mathcal{L}(\theta;T)&\doteq -\sum_{t_n\in \Omega_T}\log\left(\sum_{\immiOccurTime_j \in \Psi_T} \delta(t_n - \immiOccurTime_j) + \kappa\theta e^{(\kappa-1)\theta (t_n - \immiOccurTime_j)} \cdot \llbracket t_n > \immiOccurTime_j \rrbracket\right)\notag\\
    &+ \vert \Psi_T \vert + \sum_{\immiOccurTime_j \in \Psi_T} \left[ \frac{\kappa}{\kappa-1}\left(e^{(\kappa-1)\theta (T-\immiOccurTime)} -1\right)\right]
\end{align}

We will prove that for fixed $\theta$, $\mathcal{L}(\theta, T)$ is convex with respect to $\kappa$ over $\kappa \in [0,1)$.

Our first step is to prove convexity in $\kappa$ of the NLL for the single impulse response. Suppose that $s(t) = \delta(t-\immiOccurTime)$ for $\immiOccurTime \geq 0$. The corresponding NLL is given by

\begin{align*}
	\mathcal{L}(\theta;T)&\doteq \overbrace{-\sum_{t_n\in \Omega_T}\log\left( \delta(t_n - \immiOccurTime) + \kappa\theta e^{(\kappa-1)\theta (t_n - \immiOccurTime)} \cdot \llbracket t_n > \immiOccurTime \rrbracket\right)}^{l(\xi(\cdot; \kappa))}\notag\\&+ \underbrace{1 + \frac{\kappa}{\kappa-1}\left(e^{(\kappa-1)\theta (T-\immiOccurTime)} -1\right)}_{\Xi(T; \kappa)}.
\end{align*}

As we can see, the single impulse NLL can be written as a sum of two terms, a function of the conditional intensity $l(\xi(\cdot; \kappa))$, where $l: x \mapsto -\sum \log(x)$, and the compensator at the final time $\Xi(T; \kappa)$. To prove convexity of $\mathcal{L}(\theta, T)$, we show that both components are convex in $\kappa$.

Given that the function $l$ is non-decreasing and convex (since a sum of convex functions is still convex), it suffices to show that 
\begin{equation*}
    \xi(t_n; \kappa) = \delta(t_n - \immiOccurTime) + \kappa\theta e^{(\kappa-1)\theta (t_n - \immiOccurTime)} \cdot \llbracket t_n > \immiOccurTime \rrbracket
\end{equation*}
is convex in $\kappa$ to prove that $l(\xi(\cdot; \kappa))$ is convex in $\kappa$. This follows from the property that if $f$ and $g$ are convex and $f$ is non-decreasing, then $f \circ g$ is convex.

It can be shown that
\begin{equation*}
\frac{\partial^2 \xi(t_n; \kappa)}{\partial \kappa^2} = \theta^2 (t_n-\immiOccurTime) e^{\theta (t_n - \immiOccurTime) (\kappa-1)} (\kappa\theta (t_n - \immiOccurTime) + 2) \cdot \llbracket t_n > \immiOccurTime \rrbracket \geq 0,
\end{equation*}
which proves convexity of $\xi(t_n; \kappa)$ for $\kappa$. Thus, $l(\xi(\cdot; \kappa))$ is a convex function.

We now show convexity of $\Xi(T; \kappa)$ in $\kappa$. It can be shown that
\begin{equation*}
\frac{\partial^2 \Xi(T; \kappa)}{\partial \kappa^2} = \frac{2 - e^{-\theta(T-\immiOccurTime)(1-\kappa)}(2 + 2\theta(T-\immiOccurTime)(1-\kappa)+(\theta(T-\immiOccurTime))^2\kappa(1-\kappa)^2)}{(1-\kappa)^3}
\end{equation*}

So that  $\frac{\partial^2 \Xi(T; \kappa)}{\partial \kappa^2} \geq 0$, we must show that
\begin{equation*}
p(k) = e^{-\theta(T-\immiOccurTime)(1-\kappa)}(2 + 2\theta(T-\immiOccurTime)(1-\kappa)+(\theta(T-\immiOccurTime))^2\kappa(1-\kappa)^2){(1-\kappa)^3} \leq 2
\end{equation*}
whenever $0 \leq \kappa < 1$. Since
\begin{equation*}
    \frac{\d p}{\d k} = \theta^2(T-\immiOccurTime)^2 (1-\kappa)^2 e^{-\theta(T-\immiOccurTime)(1-(\theta(T-\immiOccurTime)))} (\theta(T-\immiOccurTime)\kappa + 3) > 0,
\end{equation*}
for $0 \leq \kappa < 1$, which means that $p(k)$ is non-decreasing over this interval, and $p(1)=2$, we have 
\begin{equation*}
\frac{\partial^2 \Xi(T; \kappa)}{\partial \kappa^2} \geq 0,
\end{equation*}
for $0 \leq \kappa < 1$. Thus, $\Xi(T; \kappa)$ is convex in $\kappa$ over this range.

Since $l(\xi(\cdot; \kappa))$ and $\Xi(T; \kappa)$ are both convex, their sum $\mathcal{L}(\theta, T)$, the NLL for the single impulse response, is convex.

Convexity of the NLL for the multi impulse response follows from the following observation: Observe that the conditional intensity $\xi$ and the compensator $\Xi$ for the multi impulse response can be expressed, respectively, as sums of $\xi$ and $\Xi$ of the single impulse response. Given that a nonnegative-weighted sum of convex functions is still convex, it follows that the NLL for the multi impulse response is convex as well.

For $\theta$, observe that

\begin{equation*}
    \theta\to -\log\left( \delta(t_n- \immiOccurTime) + \kappa\theta e^{(\kappa-1)\theta (t_n- \immiOccurTime)} \cdot \llbracket t_n > \immiOccurTime \rrbracket\right)
\end{equation*}
is convex in $\theta$ but 
\begin{align*}
    \frac{\partial^2}{\partial\theta^2}\left(\frac{\kappa}{\kappa-1}e^{(\kappa-1)\theta (T - \immiOccurTime)}\right)=\kappa(\kappa-1) (T-\immiOccurTime)^2 e^{(\kappa-1)\theta (T-\immiOccurTime)}<0
\end{align*}
for $0<\kappa<1$. Thus $\frac{\partial^2\mathcal{L}}{\partial\theta^2}\ge 0$ does not hold, implying that the NLL is not convex in $\theta$ for fixed $\kappa$.
 
\section{Experiments}

\subsection{Number of Approximation Points}
\label{app:experiments_num_approx}

\begin{figure}[h]
    \centering
    \includegraphics[width=0.45\textwidth]{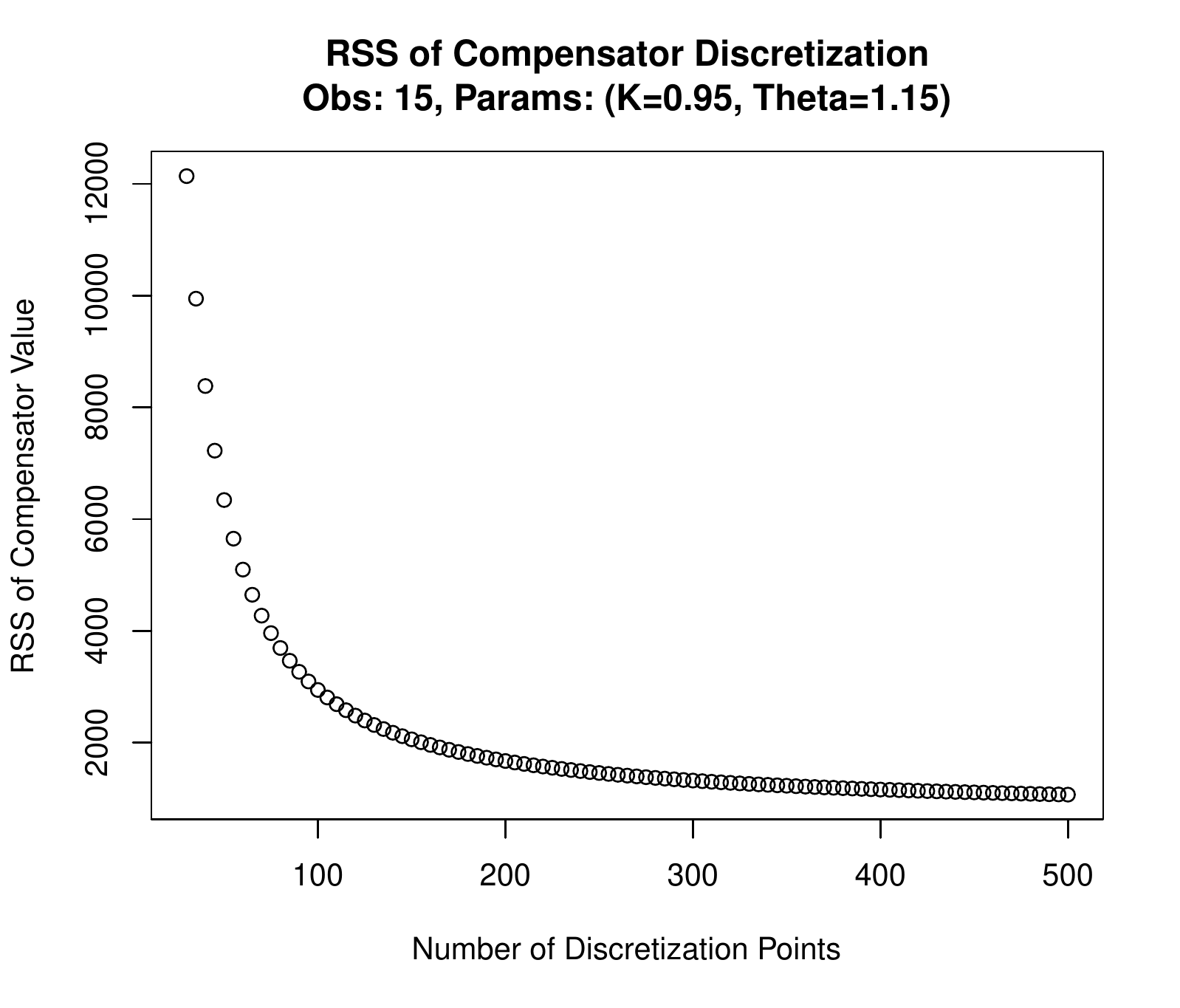}
    \caption{Compensator approximation error with respect to the number of discretization points.}
    \label{fig:comp_multi_disc}
\end{figure}

\cref{fig:comp_multi_disc} shows the RSS of the compensator values of the discrete compensator function versus the closed-form compensator function.

\subsection{Other Non-endogenous Experiments} \label{app:experiments_non_endogenous}

\begin{figure}
    \centering
    \subfloat[]{\includegraphics[width=0.45\textwidth]{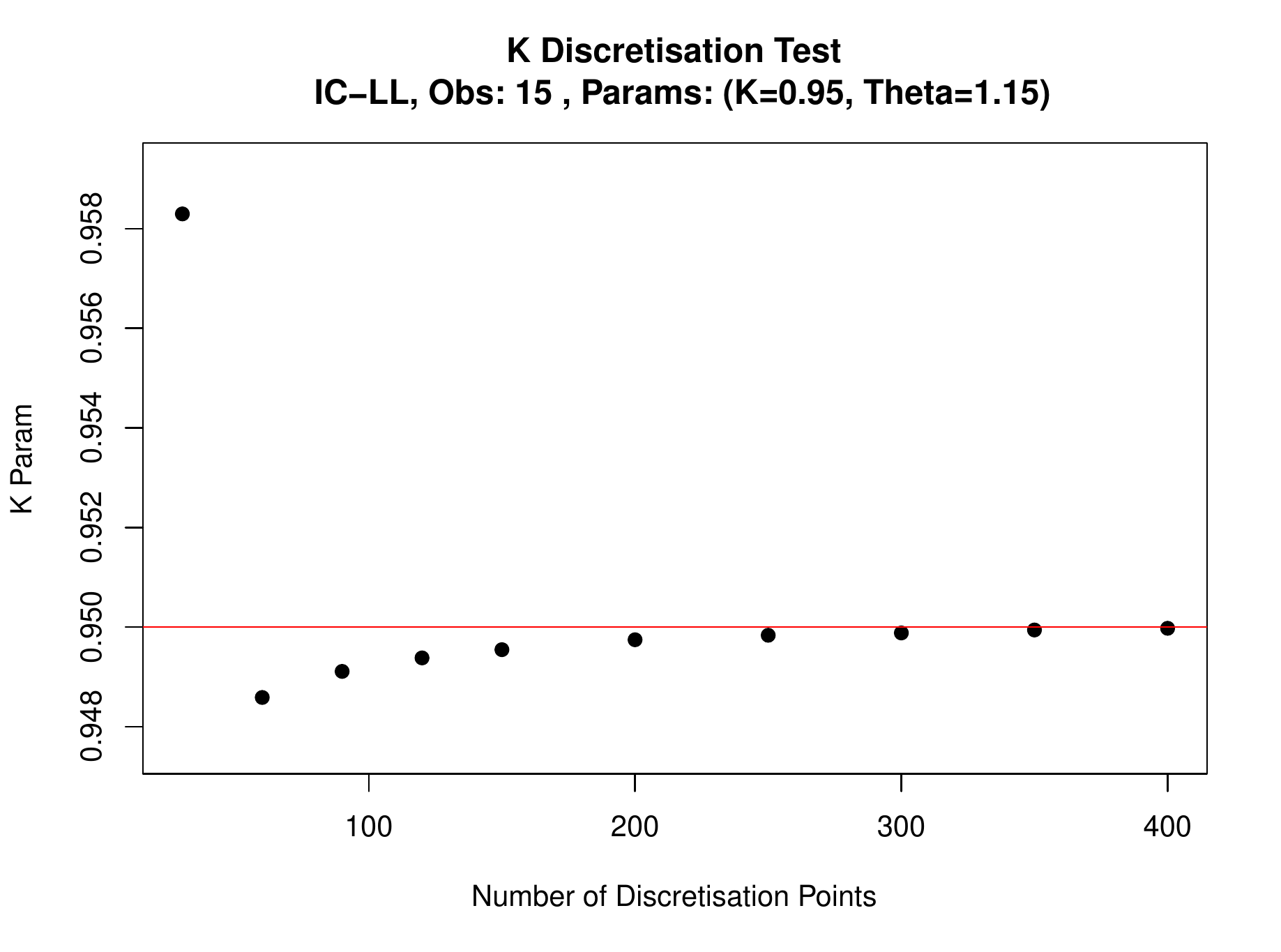}}
    \subfloat[]{\includegraphics[width=0.45\textwidth]{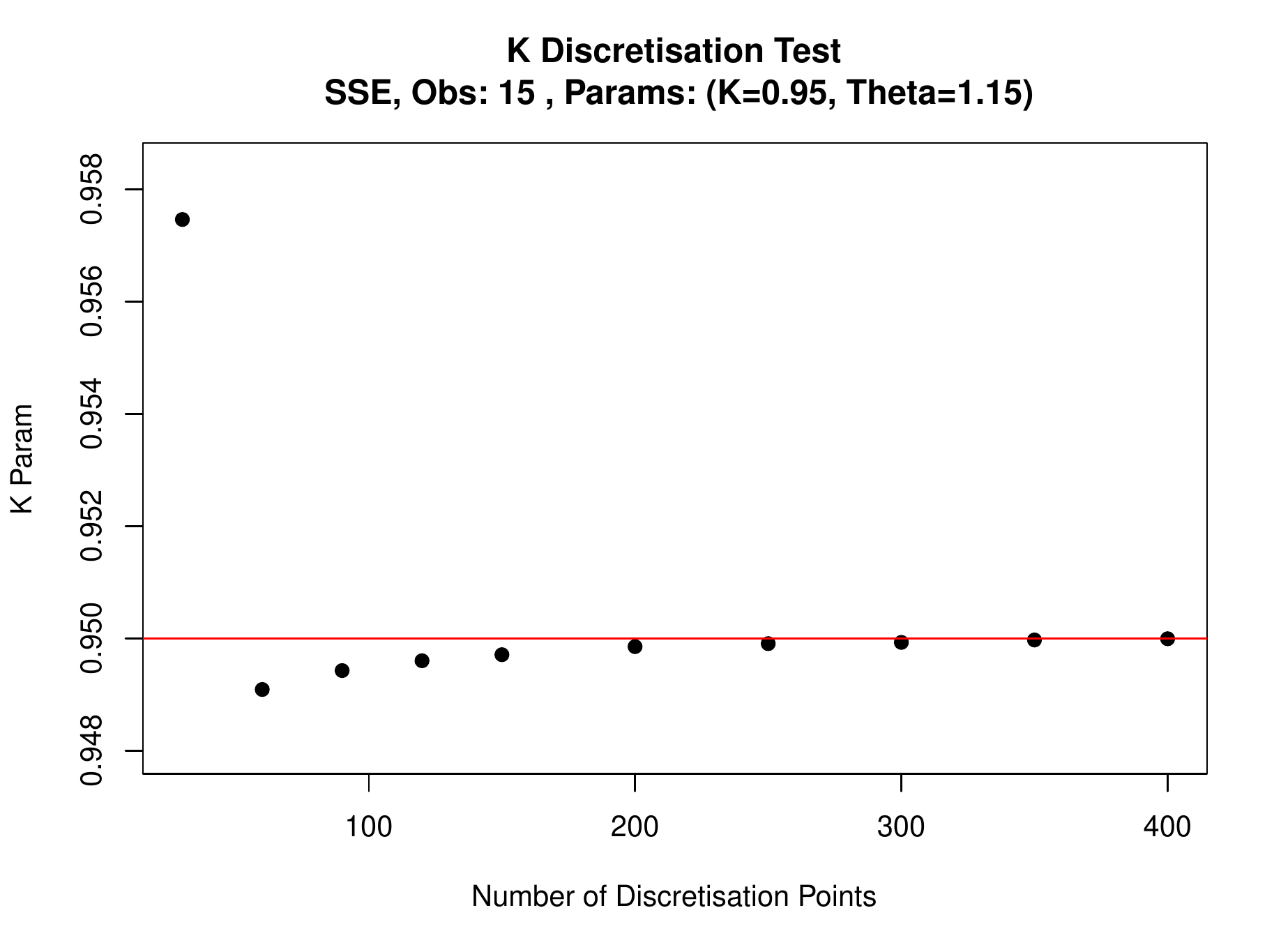}}\\
    \subfloat[]{\includegraphics[width=0.45\textwidth]{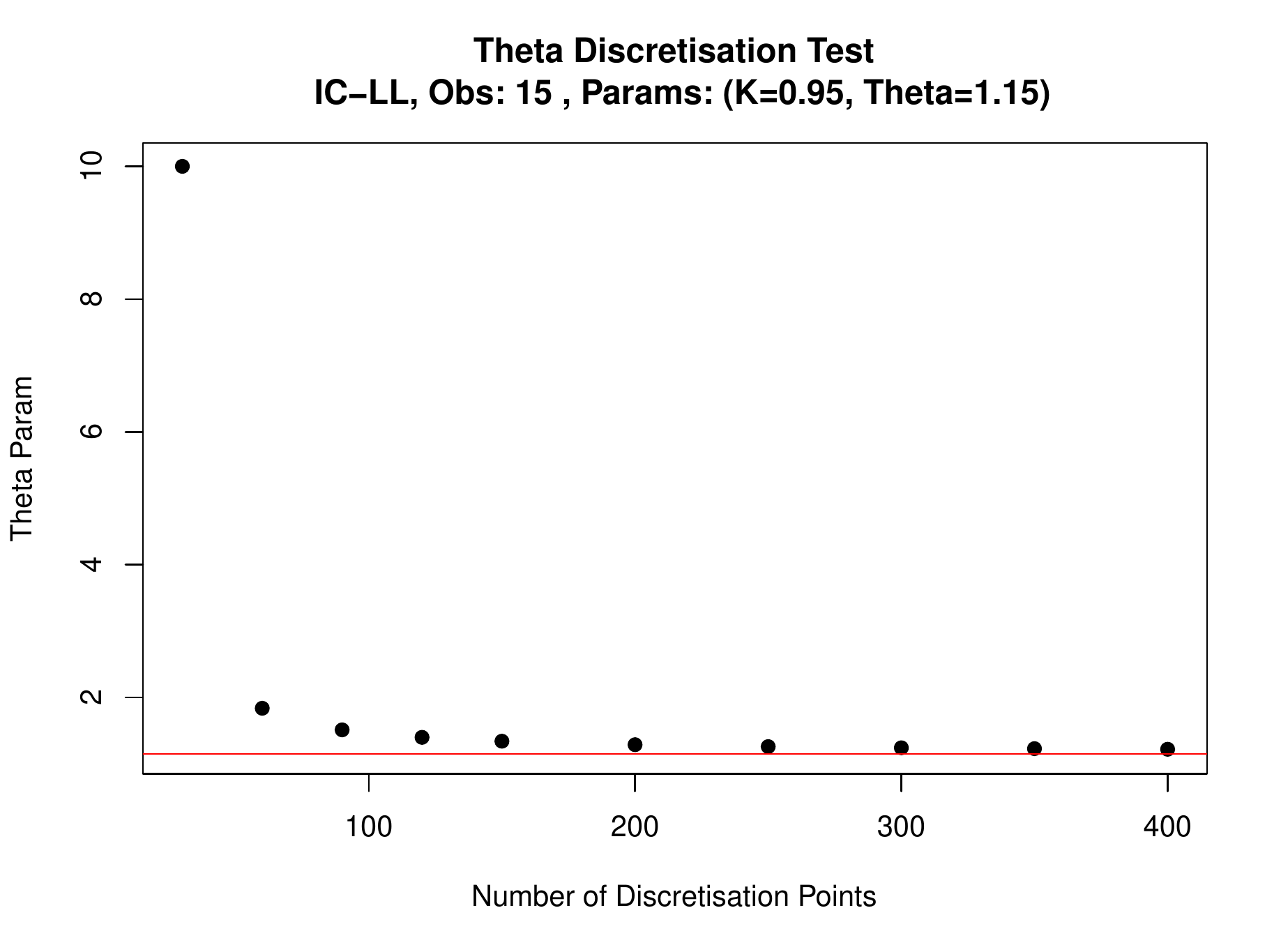}}
    \subfloat[]{\includegraphics[width=0.45\textwidth]{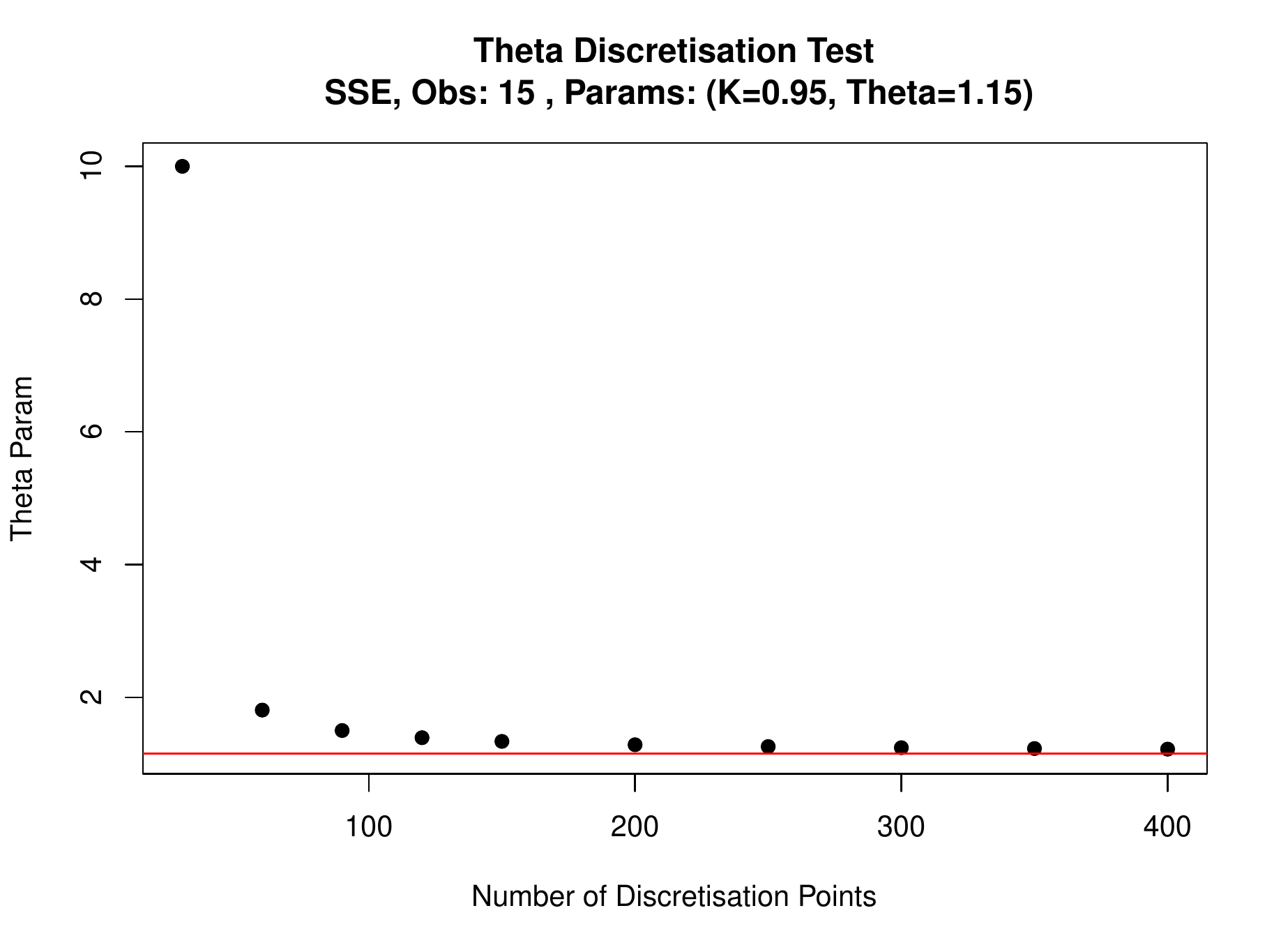}}\\
    \caption{Plots detailing the parameter fits of synthetic data in setting D with varying numbers of discretization points. The plots are taken over 15 observation intervals, using \( \kappa = 0.95 \) and \( \theta = 1.15 \) parameters. The red line indicates these true parameters of the synthetic data.}
    \label{fig:varying_discretisation}
\end{figure}

\cref{fig:varying_discretisation} shows the parameter fittings resulting from a different number of discretization points.

\begin{table}[htbp]
    \centering
    \caption{Non-endogenous synthetic experiments for Scenario F.}
    \label{tab:app_experiment_g}
    \resizebox{\columnwidth}{!}{\begin{tabular}{lclcllcccccccccccc}
            trule
            \multirow{2}{*}{True Params.} &  & \multirow{2}{*}{Loss Func.} &  & \multirow{2}{*}{Model}  & \multirow{2}{*}{} &  & \multicolumn{10}{c}{Number of Observation Intervals}                                                                                                                                        \\
            \cmidrule{8-18}
                                          &  &                             &  &                         &                   &  & 5                                                    &  & 10                    &  & 15                    &  & 30                    &  & 60                    &  & 100                   \\
            \cmidrule{1-1}
            \cmidrule{3-3}
            \cmidrule{5-6}
            \cmidrule{8-8}
            \cmidrule{10-10}
            \cmidrule{12-12}
            \cmidrule{14-14}
            \cmidrule{16-16}
            \cmidrule{18-18}
            \multirow{8}{*}{\shortstack[1]{\( \kappa = 0.6 \)                                                                                                                                                                                                                                                                \\ \( \theta = 0.8 \)}} && \multirow{4}{*}{SSE} && \multirow{2}{*}{Closed} & \( \kappa \) && \( 0.726 \pm 0.004 \) && \( 0.682 \pm 0.005 \) && \( 0.668 \pm 0.005 \) && \( 0.642 \pm 0.005 \) && \( 0.62 \pm 0.005 \) && \( 0.612 \pm 0.005 \) \\
                                          &  &                             &  &                         & \( \theta \)      &  & \( 0.647 \pm 0.021 \)                                &  & \( 0.641 \pm 0.024 \) &  & \( 0.801 \pm 0.031 \) &  & \( 0.995 \pm 0.052 \) &  & \( 0.915 \pm 0.052 \) &  & \( 0.865 \pm 0.045 \) \\
            \cmidrule{5-18}
                                          &  &                             &  & \multirow{2}{*}{Approx} & \( \kappa \)      &  & \( 10 \pm 0 \)                                       &  & \( 0.621 \pm 0.006 \) &  & \( 0.605 \pm 0.006 \) &  & \( 0.601 \pm 0.006 \) &  & \( 0.601 \pm 0.006 \) &  & \( 0.601 \pm 0.006 \) \\
                                          &  &                             &  &                         & \( \theta \)      &  & \( 0.005 \pm 0 \)                                    &  & \( 8.823 \pm 0.866 \) &  & \( 9.546 \pm 1.589 \) &  & \( 1.09 \pm 0.091 \)  &  & \( 0.909 \pm 0.052 \) &  & \( 0.856 \pm 0.043 \) \\
            \cmidrule{3-18}
                                          &  & \multirow{4}{*}{IC-LL}      &  & \multirow{2}{*}{Closed} & \( \kappa \)      &  & \( 0.728 \pm 0.004 \)                                &  & \( 0.69 \pm 0.005 \)  &  & \( 0.673 \pm 0.005 \) &  & \( 0.644 \pm 0.005 \) &  & \( 0.622 \pm 0.005 \) &  & \( 0.613 \pm 0.005 \) \\
                                          &  &                             &  &                         & \( \theta \)      &  & \( 0.644 \pm 0.019 \)                                &  & \( 0.67 \pm 0.023 \)  &  & \( 0.828 \pm 0.031 \) &  & \( 1.014 \pm 0.054 \) &  & \( 0.931 \pm 0.05 \)  &  & \( 0.876 \pm 0.045 \) \\
            \cmidrule{5-18}
                                          &  &                             &  & \multirow{2}{*}{Approx} & \( \kappa \)      &  & \( 10 \pm 0 \)                                       &  & \( 0.618 \pm 0.007 \) &  & \( 0.599 \pm 0.006 \) &  & \( 0.594 \pm 0.006 \) &  & \( 0.595 \pm 0.006 \) &  & \( 0.596 \pm 0.006 \) \\
                                          &  &                             &  &                         & \( \theta \)      &  & \( 0.005 \pm 0 \)                                    &  & \( 8.355 \pm 1.02 \)  &  & \( 9.607 \pm 1.603 \) &  & \( 1.091 \pm 0.088 \) &  & \( 0.905 \pm 0.051 \) &  & \( 0.85 \pm 0.043 \)  \\
            \midrule
            \multirow{8}{*}{\shortstack[1]{\( \kappa = 0.95 \)                                                                                                                                                                                                                                                               \\ \( \theta = 1.15 \)}} && \multirow{4}{*}{SSE} && \multirow{2}{*}{Closed} & \( \kappa \) && \( 0.958 \pm 0.002 \) && \( 1.496 \pm 2.17 \) && \( 1.494 \pm 2.171 \) && \( 0.95 \pm 0.004 \) && \( 0.95 \pm 0.004 \) && \( 0.95 \pm 0.004 \) \\
                                          &  &                             &  &                         & \( \theta \)      &  & \( 1.765 \pm 0.098 \)                                &  & \( 1.362 \pm 0.358 \) &  & \( 1.301 \pm 0.344 \) &  & \( 1.283 \pm 0.089 \) &  & \( 1.216 \pm 0.08 \)  &  & \( 1.19 \pm 0.076 \)  \\
            \cmidrule{5-18}
                                          &  &                             &  & \multirow{2}{*}{Approx} & \( \kappa \)      &  & \( 10 \pm 0 \)                                       &  & \( 1.139 \pm 0.007 \) &  & \( 1.033 \pm 0.005 \) &  & \( 0.958 \pm 0.003 \) &  & \( 0.95 \pm 0.004 \)  &  & \( 0.95 \pm 0.004 \)  \\
                                          &  &                             &  &                         & \( \theta \)      &  & \( 0.014 \pm 0 \)                                    &  & \( 9.409 \pm 0.764 \) &  & \( 10 \pm 0 \)        &  & \( 10 \pm 0 \)        &  & \( 1.765 \pm 0.177 \) &  & \( 1.433 \pm 0.111 \) \\
            \cmidrule{3-18}
                                          &  & \multirow{4}{*}{IC-LL}      &  & \multirow{2}{*}{Closed} & \( \kappa \)      &  & \( 0.959 \pm 0.002 \)                                &  & \( 0.953 \pm 0.003 \) &  & \( 0.95 \pm 0.003 \)  &  & \( 0.949 \pm 0.003 \) &  & \( 0.95 \pm 0.004 \)  &  & \( 0.95 \pm 0.004 \)  \\
                                          &  &                             &  &                         & \( \theta \)      &  & \( 1.703 \pm 0.086 \)                                &  & \( 1.52 \pm 0.091 \)  &  & \( 1.451 \pm 0.096 \) &  & \( 1.314 \pm 0.083 \) &  & \( 1.233 \pm 0.075 \) &  & \( 1.202 \pm 0.071 \) \\
            \cmidrule{5-18}
                                          &  &                             &  & \multirow{2}{*}{Approx} & \( \kappa \)      &  & \( 10 \pm 0 \)                                       &  & \( 1.161 \pm 0.007 \) &  & \( 1.041 \pm 0.005 \) &  & \( 0.958 \pm 0.003 \) &  & \( 0.95 \pm 0.004 \)  &  & \( 0.95 \pm 0.004 \)  \\
                                          &  &                             &  &                         & \( \theta \)      &  & \( 0.017 \pm 0 \)                                    &  & \( 9.767 \pm 0.403 \) &  & \( 10 \pm 0 \)        &  & \( 10 \pm 0 \)        &  & \( 1.776 \pm 0.163 \) &  & \( 1.433 \pm 0.102 \) \\
            \bottomrule
        \end{tabular}}
\end{table}

\subsection{Examine the MBPP fitting bias} \label{app:mbpp_bias}
\begin{figure}[htb]
    \centering
    \subfloat[5 observational intervals]{
        \begin{tabular}{cc}
            \includegraphics[width=0.45\columnwidth]{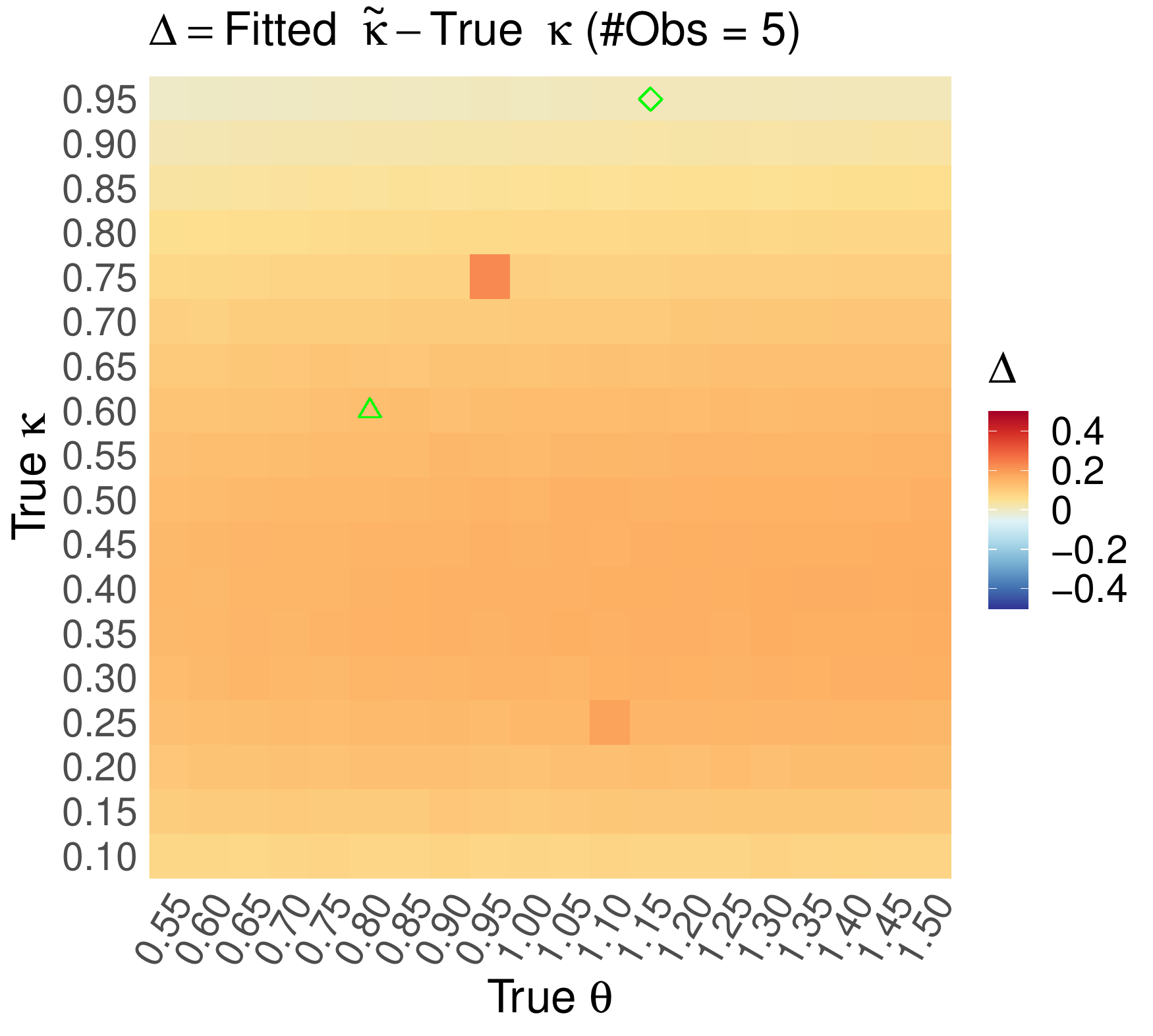}
             &
            \includegraphics[width=0.45\columnwidth]{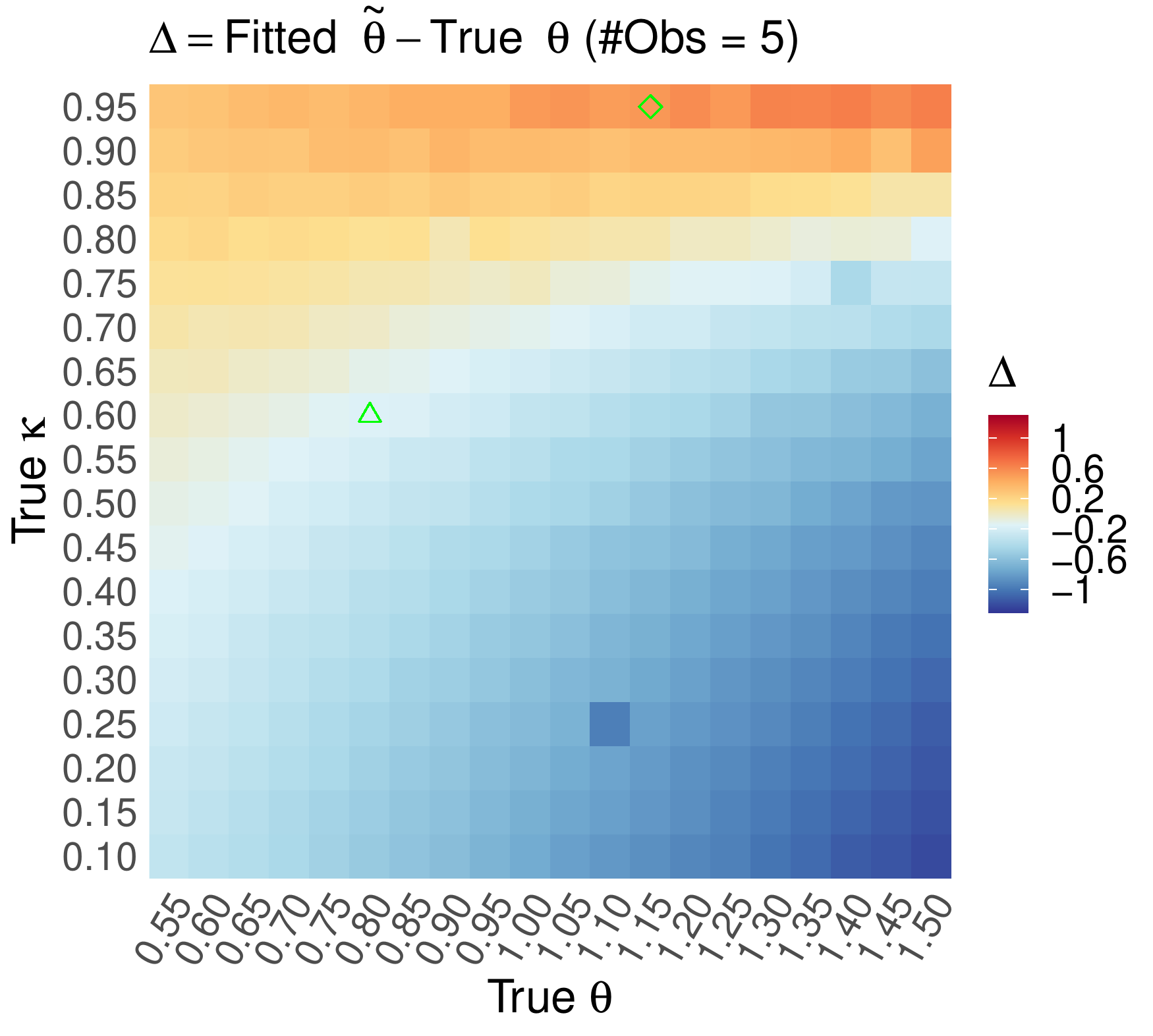}
        \end{tabular}
        \label{fig:heatmap_mean_5_obs}
    }\\\subfloat[10 observational intervals]{
        \begin{tabular}{cc}
            \includegraphics[width=0.45\columnwidth]{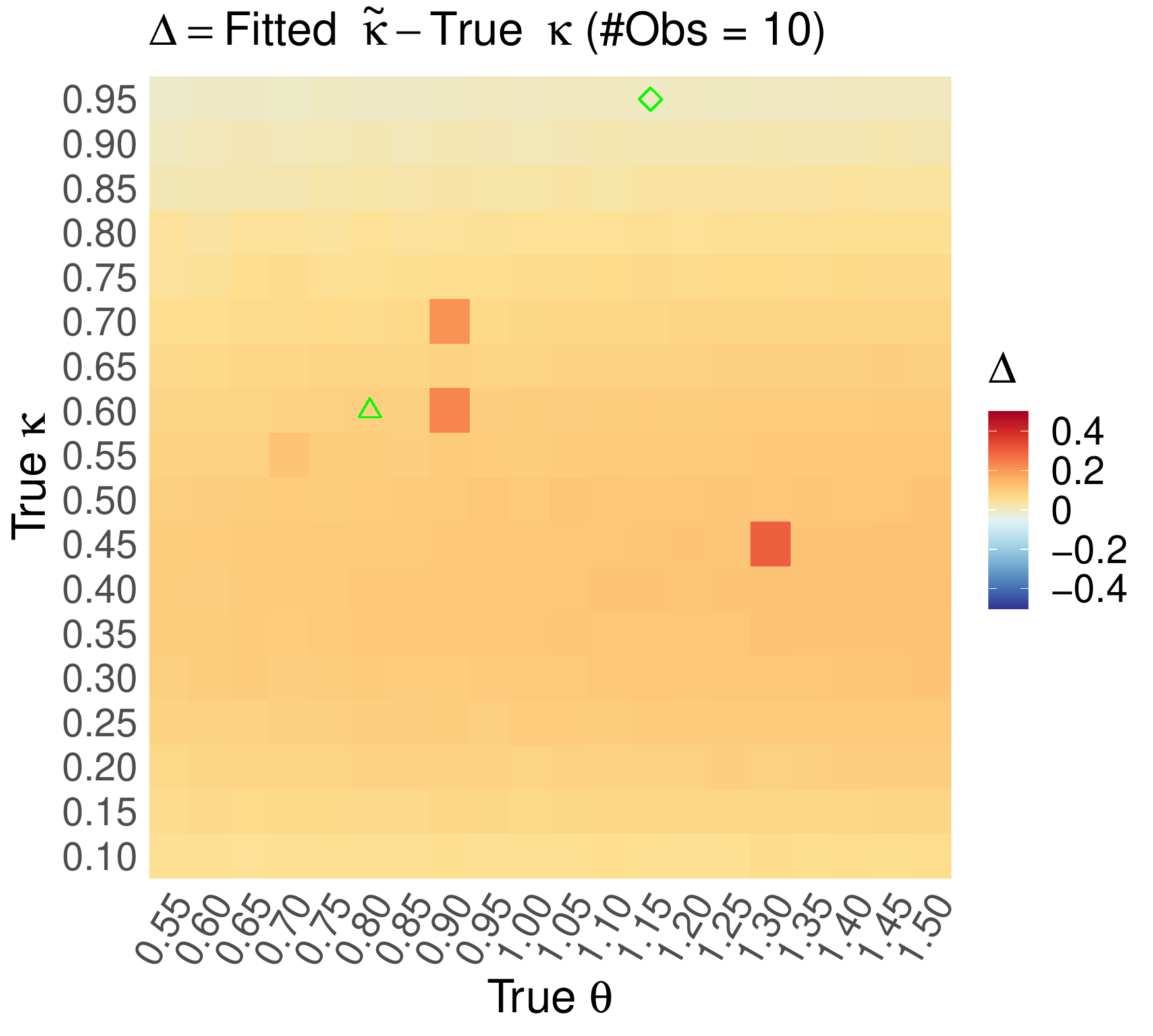}
             &
            \includegraphics[width=0.45\columnwidth]{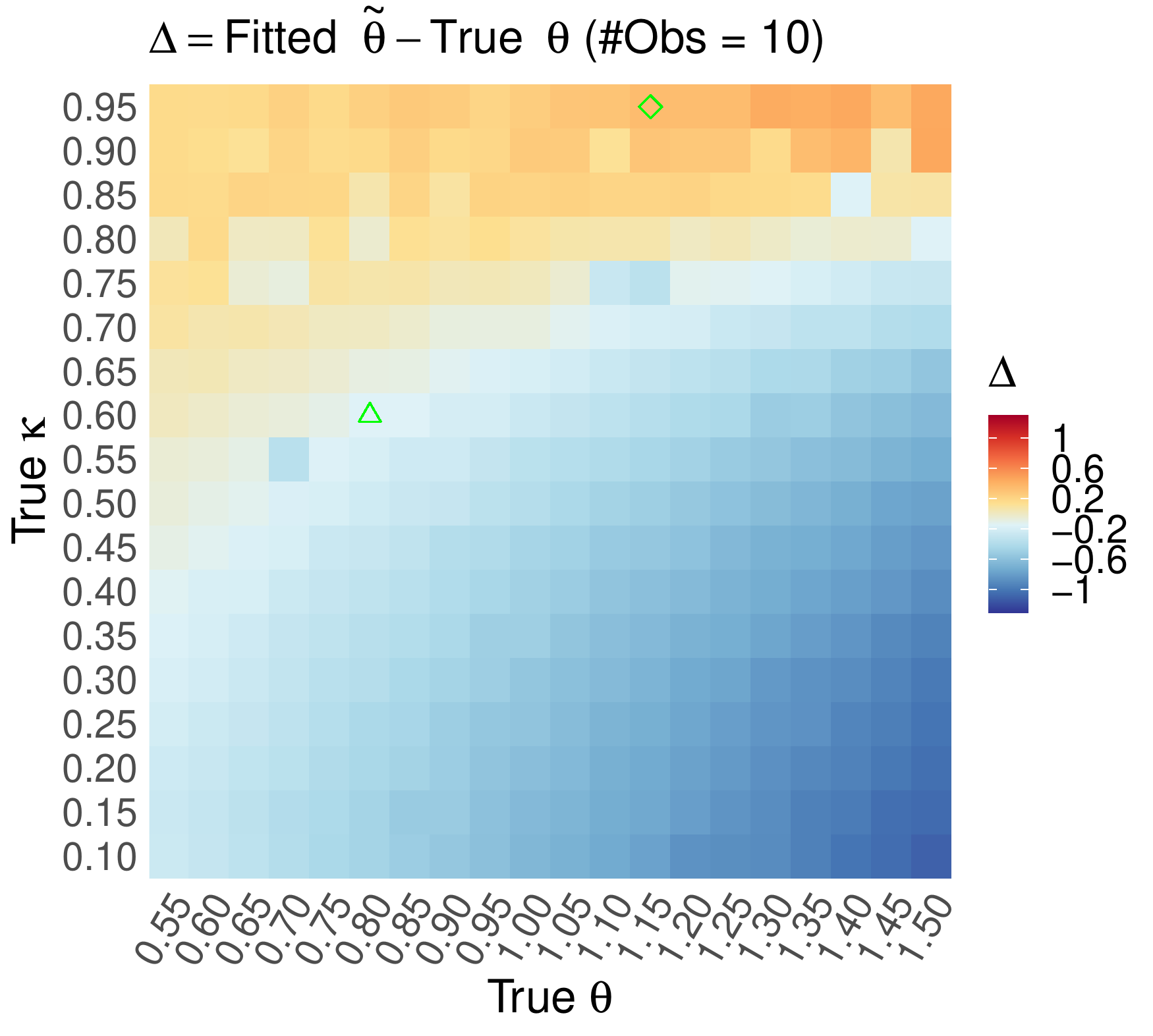}
        \end{tabular}
        \label{fig:heatmap_mean_10_obs}
    }\\\subfloat[15 observational intervals]{
        \begin{tabular}{cc}
            \includegraphics[width=0.45\columnwidth]{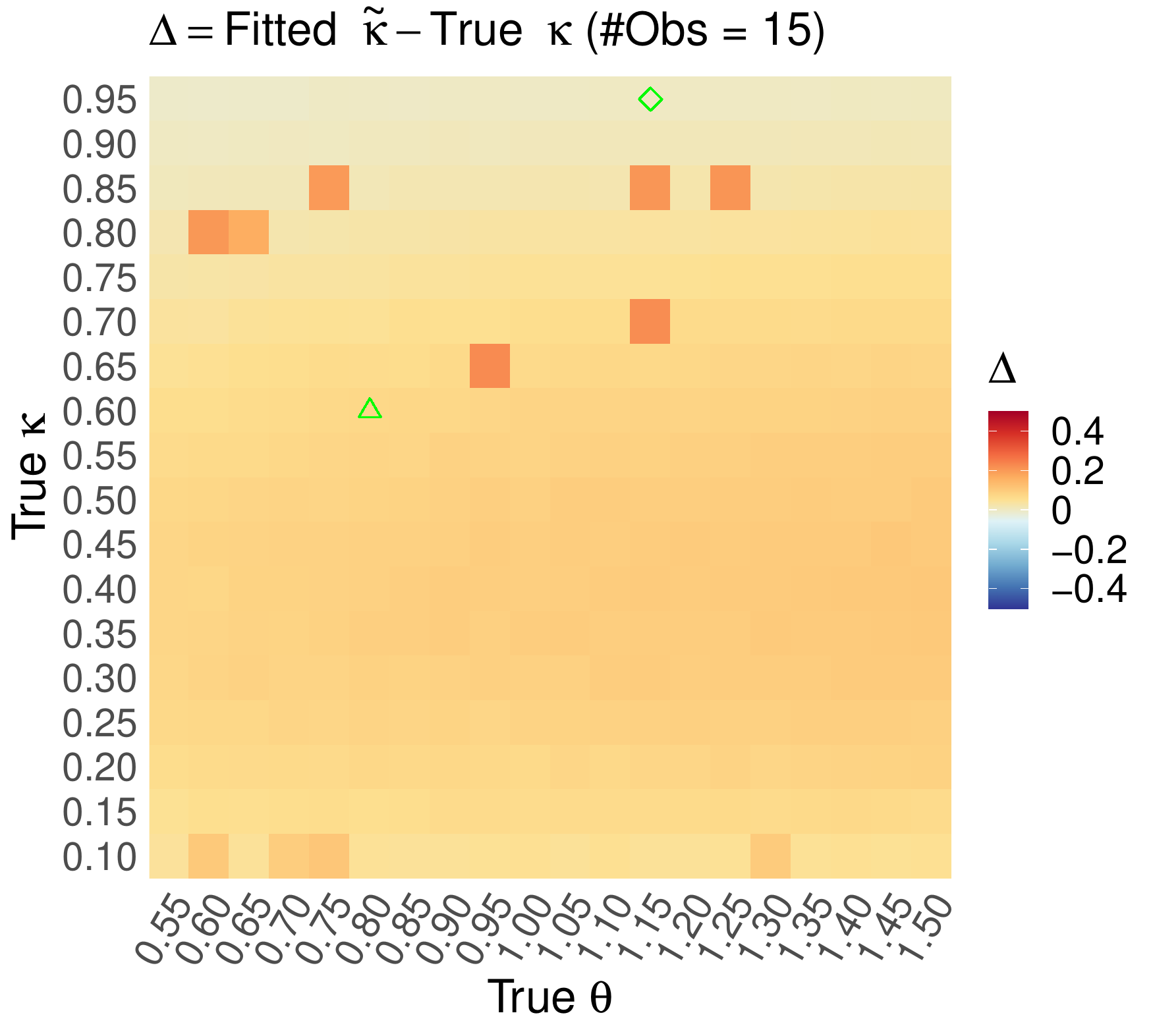}
             &
            \includegraphics[width=0.45\columnwidth]{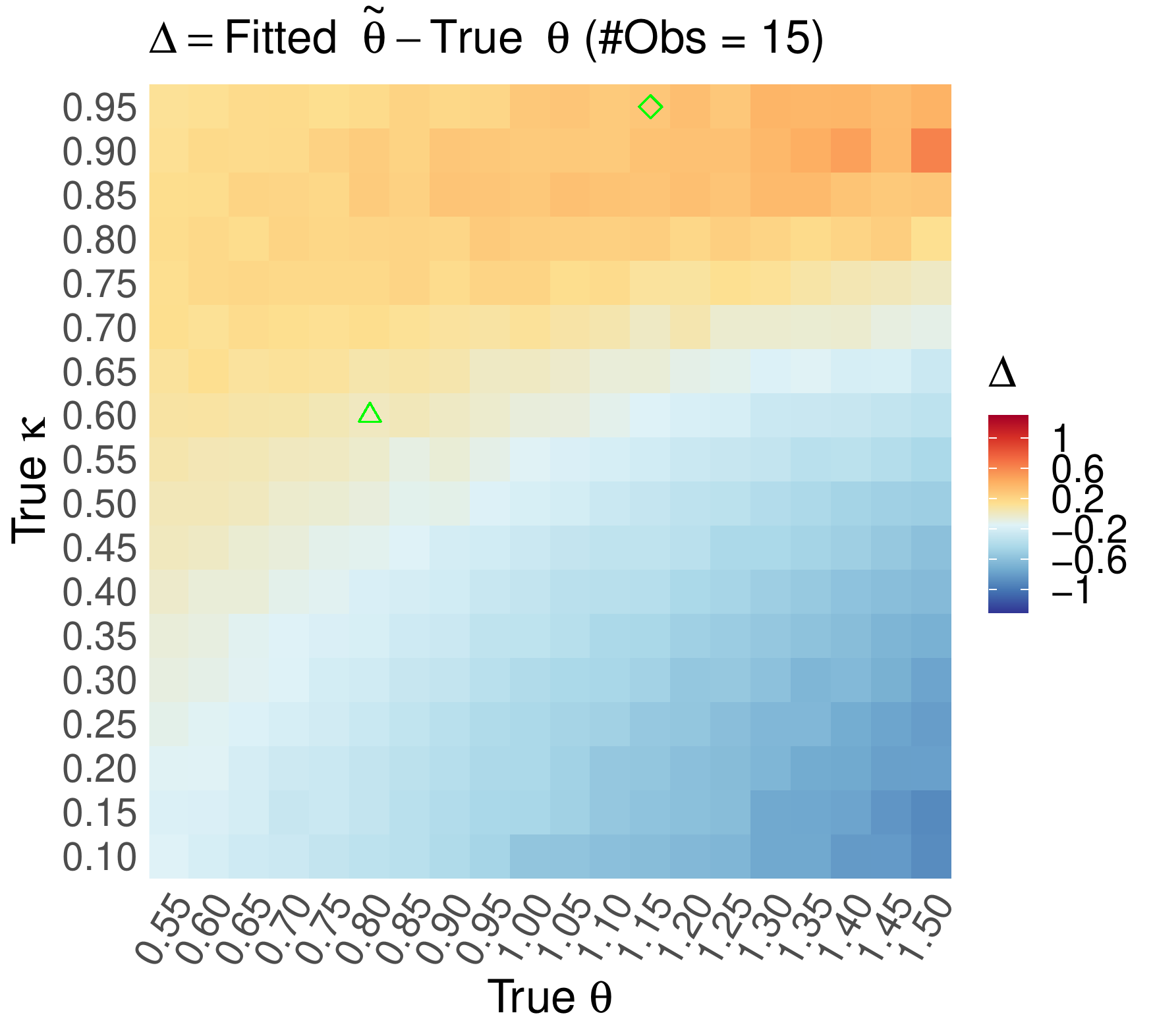}
        \end{tabular}
        \label{fig:heatmap_mean_15_obs}
    }\caption{MBPP parameter error heatmap for \( \kappa \) \textbf{(left)} and \( \theta \) \textbf{(right)}.}
    \label{fig:heatmap_mean_5-15_obs}
\end{figure}

\begin{figure}[htb]
    \centering
    \subfloat[30 observational intervals]{
        \begin{tabular}{cc}
            \includegraphics[width=0.45\columnwidth]{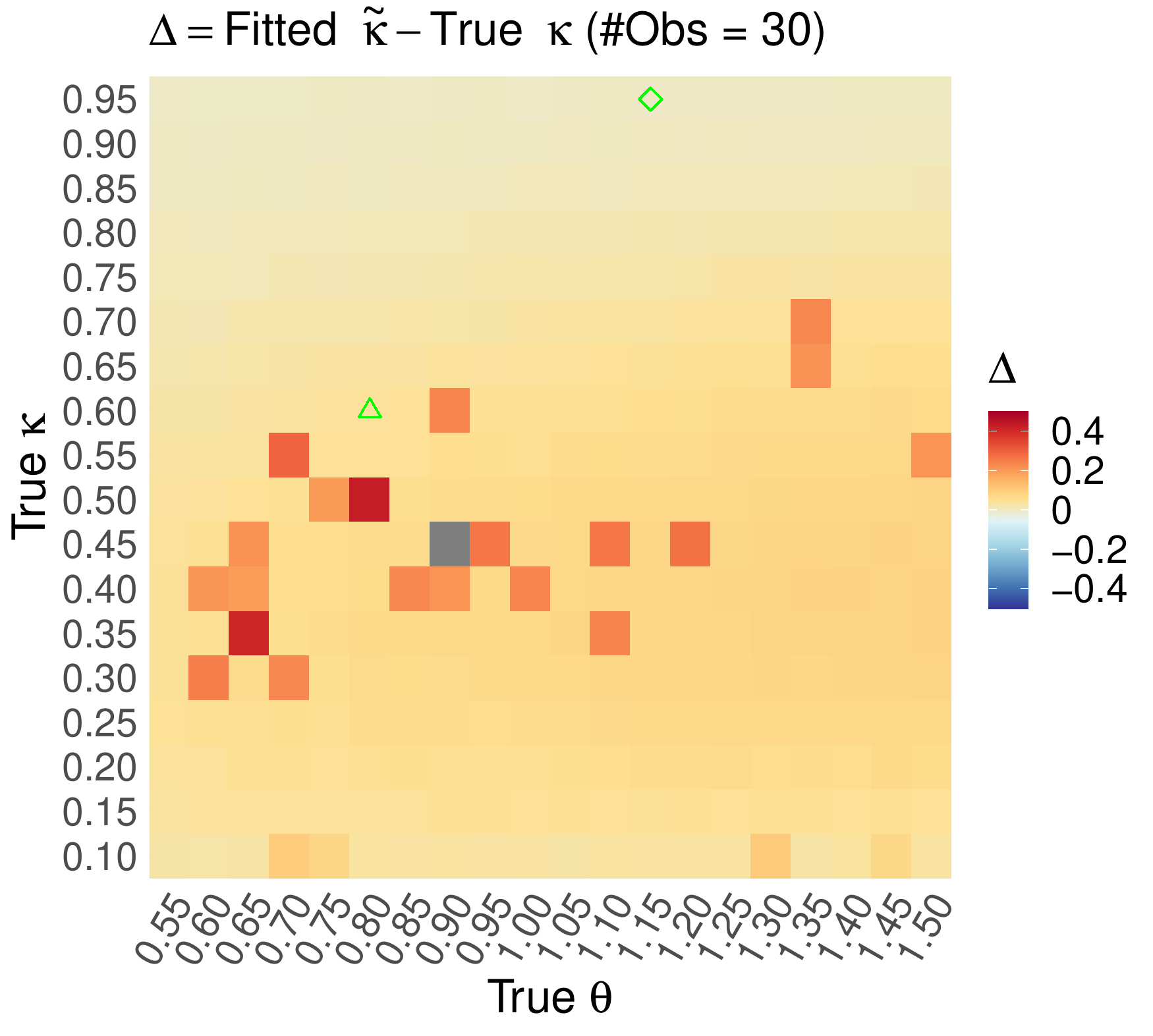}
             &
            \includegraphics[width=0.45\columnwidth]{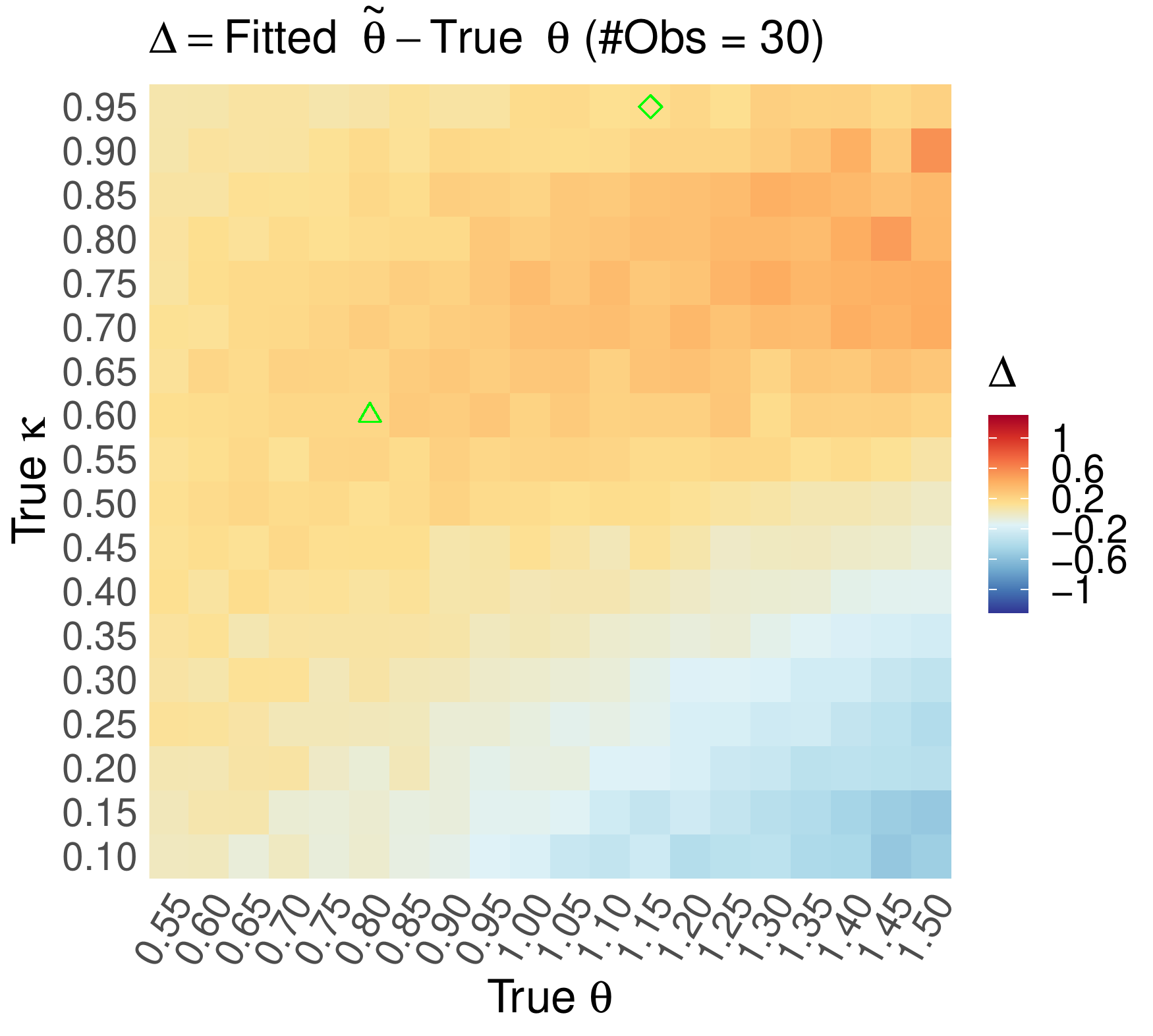}
        \end{tabular}
        \label{fig:heatmap_mean_30_obs}
    }\\\subfloat[60 observational intervals]{
        \begin{tabular}{cc}
            \includegraphics[width=0.45\columnwidth]{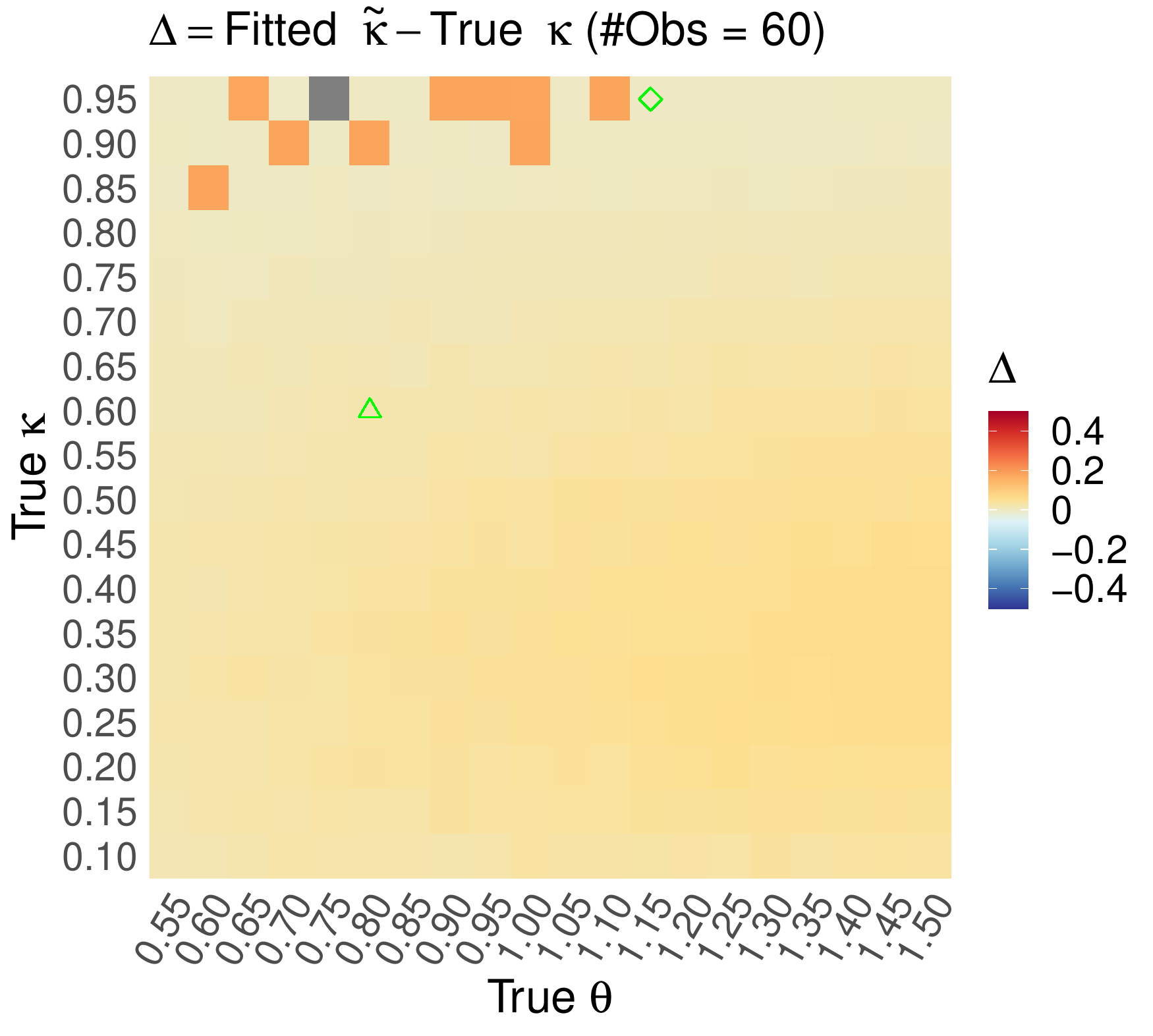}
             &
            \includegraphics[width=0.45\columnwidth]{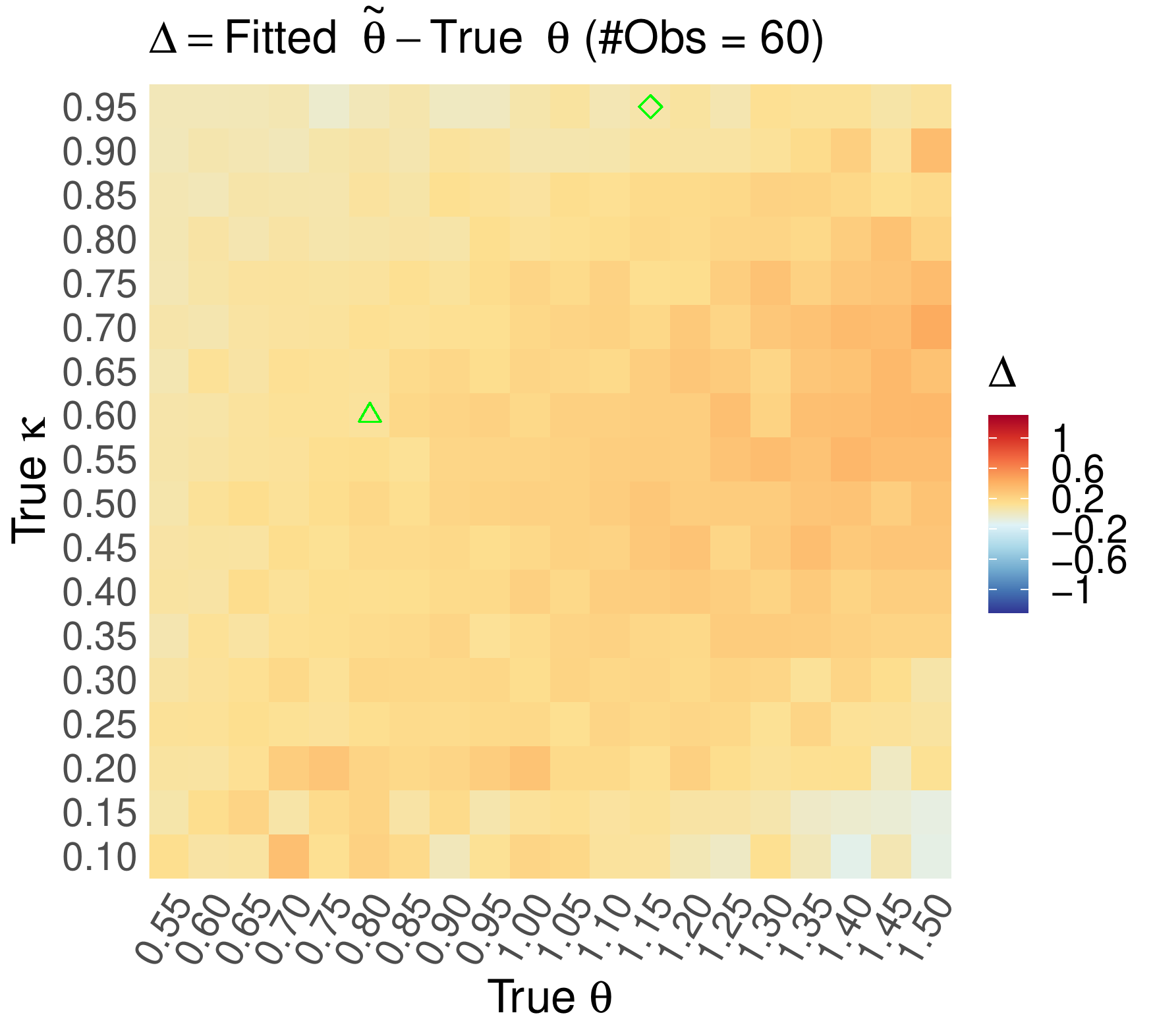}
        \end{tabular}
        \label{fig:heatmap_mean_60_obs}
    }\\\subfloat[100 observational intervals]{
        \begin{tabular}{cc}
            \includegraphics[width=0.45\columnwidth]{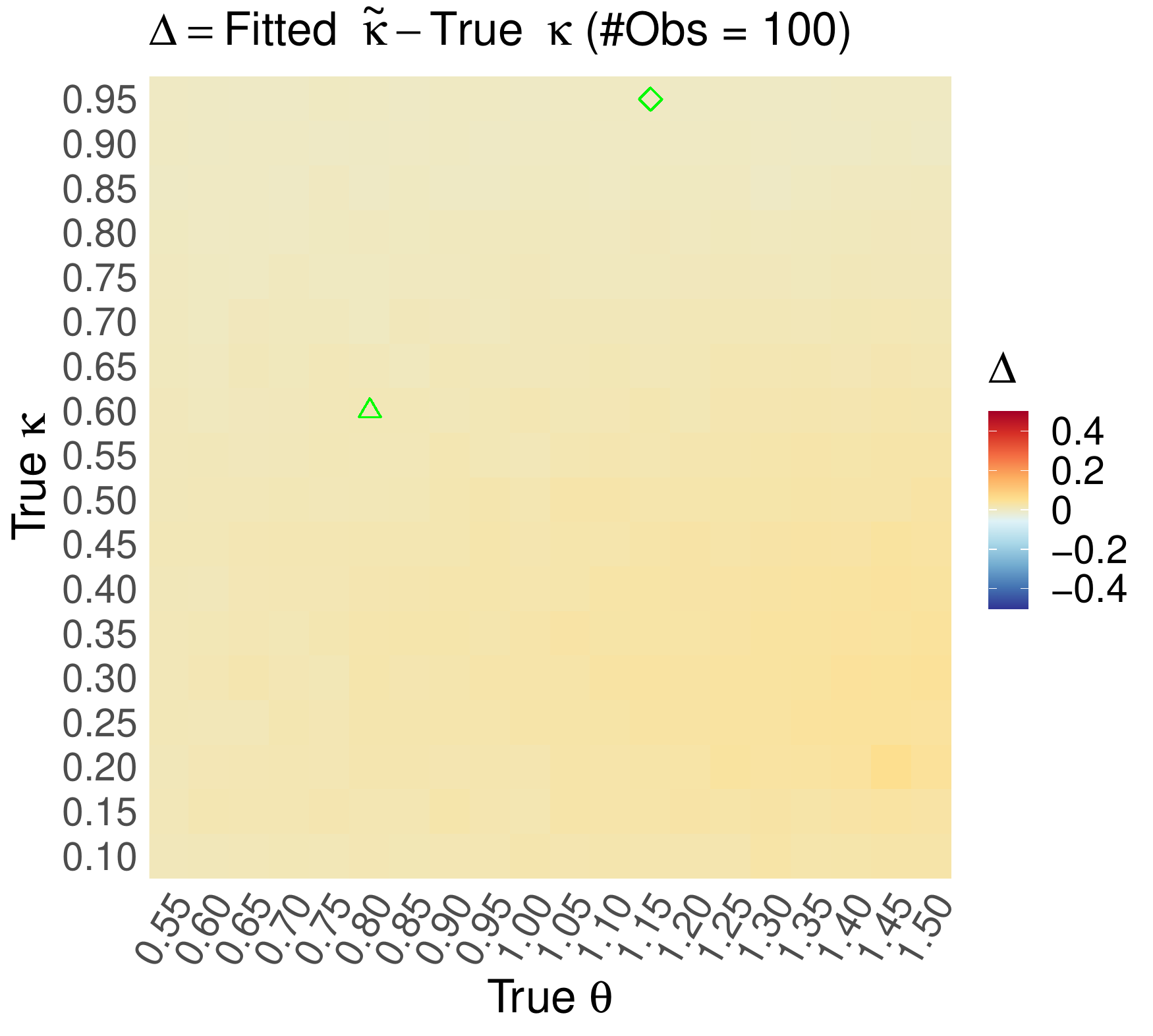}
             &
            \includegraphics[width=0.45\columnwidth]{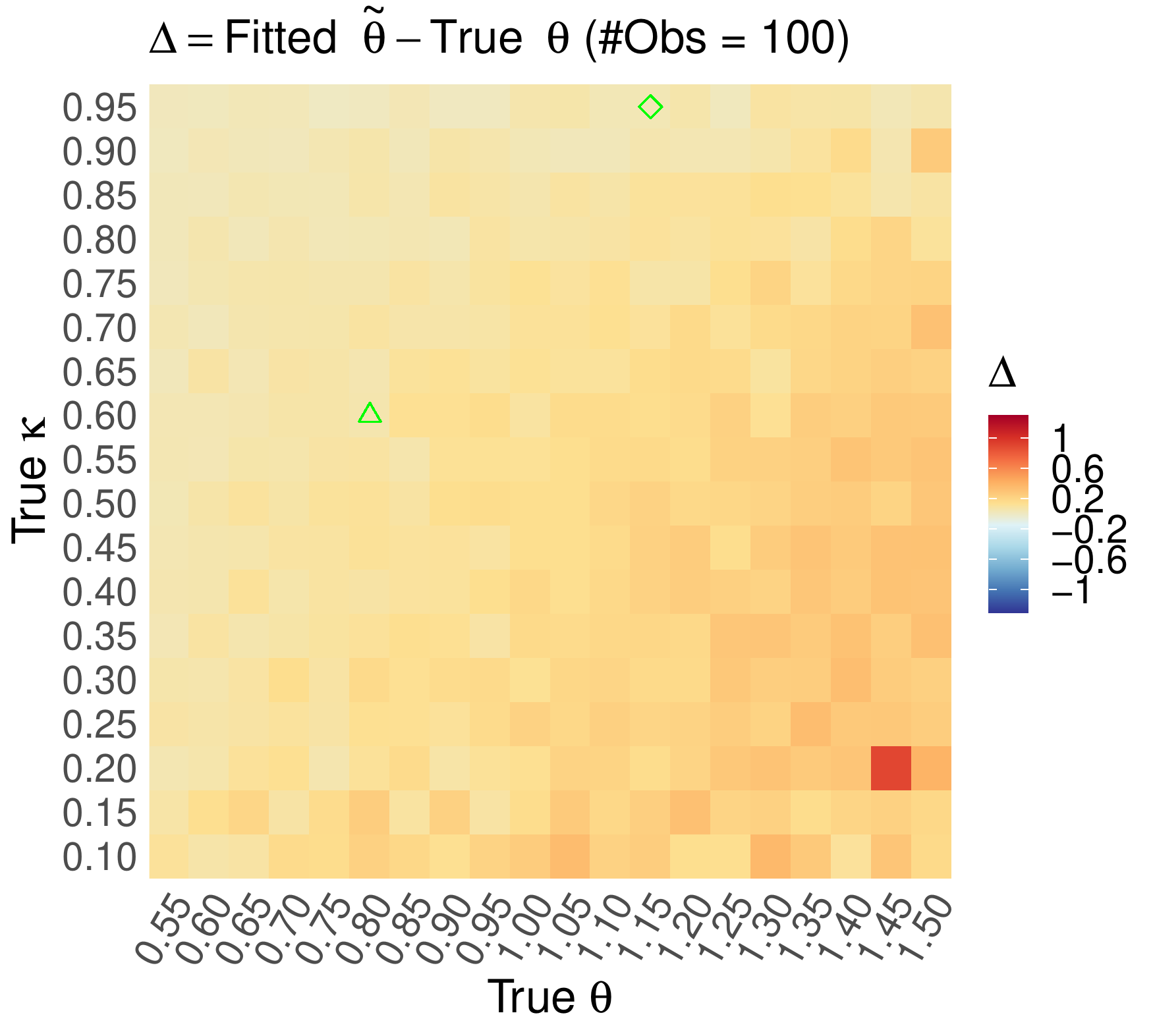}
        \end{tabular}
        \label{fig:heatmap_mean_100_obs}
    }\caption{MBPP parameter error heatmap for \( \kappa \) \textbf{(left)} and \( \theta \) \textbf{(right)}.}
    \label{fig:heatmap_mean_30-100_obs}
\end{figure}

\paragraph{Setup and protocol.}
In this section, we examine the parameter fitting bias for the MBPP process (shown in
\cref{fig:heatmap_mean_5-15_obs,fig:heatmap_mean_30-100_obs}).
We achieve this by sampling the Hawkes process over a wide range of parameters, fitting the MBPP process for each parameter combination in Scenario F---as per \cref{tab:settings}---, and examining the difference between the true parameter (used for sampling) and the fitted parameter.
We sample synthetic Hawkes realizations using \hpsin (see \cref{eq:hp_sine}) with a parameter grid.
We vary \( \kappa \) between \( 0.1 \) and \( 0.95 \) (step $0.05$) and \( \theta \) between \( 0.5 \) to \( 1.55 \) (step \( 0.05 \)).
For each parameter combination, we jointly fit 50 groups of realizations, and we obtain 50 MBPP parameter estimates (see the main-text experiments section).
We use six values for the number of observation intervals, controlling the width of the observation interval:
$5$ intervals (largest width, \cref{fig:heatmap_mean_5_obs}),
$10$ intervals (\cref{fig:heatmap_mean_10_obs}),
$15$ intervals (\cref{fig:heatmap_mean_15_obs}),
$30$ intervals (\cref{fig:heatmap_mean_30_obs}),
$60$ intervals (\cref{fig:heatmap_mean_60_obs}), and
$100$ intervals (smallest width, \cref{fig:heatmap_mean_100_obs}).
We report the mean difference between fitted and true values for $\kappa$ (left panel of each figure) and $\theta$ (right panel). 
Furthermore we highlight the specific parameters used in the main-text by the green shapes, with the triangle signifying \( \kappa = 0.6 \) and \( \theta = 0.8 \); and diamond signifying  \( \kappa = 0.95 \) and \( \theta = 1.15 \).

\paragraph*{Results and interpretation}
There are several conclusions that emerge from \cref{fig:heatmap_mean_5-15_obs,fig:heatmap_mean_30-100_obs}.
First, we observe that we retrieve parameter \( \kappa \) with only minor errors over every parameter configuration.
The occasional more consistent over-estimation is likely due to convergence failures of the optimization algorithm.
Second, we observe that fitting the \( \theta \) parameter is significantly more challenging. 
For high values of \( \theta \) the fitting under-estimates the true $\theta$ for low values of $\kappa$ (bottom-right corner) and over-estimates it for high values of $\kappa$ (top-right corner).
Note that high $\theta$ corresponds to a faster decay rate for the Hawkes kernel $\phi(t)$, which typically translates to offspring events being tightly clustered close to their parent event.
As $\theta$ decreases (slower kernel decay), the parameter is increasingly well recovered.
Third observation, there seems to be a breakpoint between regions where parameters are underestimated (blue) and overestimated (red) for the \( \theta \) plots. 
Forth and as expected, for higher data granularity --- \ie, the higher the number of intervals --- the \( \theta \) are recovered closer to their true values.
Fifth and finally, for high data granularity (100 intervals in \cref{fig:heatmap_mean_100_obs}) we observe a slight over-estimation for both $\kappa$ and $\theta$.
We posit that this consistent over-estimation is due to the model mismatch between the generating model (Hawkes process) and the fitting model (MBPP).

\end{document}